\documentclass[lettersize,journal]{IEEEtran}
\usepackage{amsmath,amsfonts}
\usepackage{algorithmic}
\usepackage{algorithm}
\usepackage{array}
\usepackage[caption=false,font=normalsize,labelfont=sf,textfont=sf]{subfig}
\usepackage{textcomp}
\usepackage{stfloats}
\usepackage{url}
\usepackage{verbatim}
\usepackage{graphicx}
\usepackage{cite}
\usepackage{hyperref}
\usepackage{etoolbox}
\usepackage{booktabs}
\usepackage{xcolor}
\usepackage{multirow}

\usepackage{cleveref}
\crefname{figure}{Figure}{Figures} 
\Crefname{figure}{Figure}{Figures} 
\crefname{table}{Table}{Tables}
\Crefname{table}{Table}{Tables}
\crefname{section}{Section}{Sections}
\Crefname{section}{Section}{Sections}
\crefname{appendix}{Appendix}{Appendix}
\crefname{equation}{Equation}{Equations}
\Crefname{equation}{Equation}{Equations}

\usepackage{amsmath}
\usepackage{graphicx}
\usepackage{xcolor}
\usepackage{tikz}
\usepackage{forest}
\usepackage{adjustbox}
\usepackage[numbers]{natbib} 
\usetikzlibrary{trees} 
\definecolor{mycolor}{RGB}{134,150,167}
\definecolor{hidden-draw}{RGB}{0,0,0}

\makeatletter
\patchcmd{\@makecaption}
  {\scshape}
  {}
  {}
  {}
\makeatother

\begin{document}

\title{Towards Explainable and Interpretable Multimodal Large Language Models: A Comprehensive Survey}


\author{
\bf Yunkai Dang\textsuperscript{1,*}\quad Kaichen Huang\textsuperscript{1,*}\quad Jiahao Huo\textsuperscript{1,*}\quad Yibo Yan\textsuperscript{1,2}\quad
\bf Sirui Huang\textsuperscript{1}\quad \\Dongrui Liu\textsuperscript{3} 
\quad Mengxi Gao\textsuperscript{1}\quad Jie Zhang\textsuperscript{3}\quad
\bf Chen Qian\textsuperscript{3}\quad Kun Wang\textsuperscript{4}\quad \\ Yong Liu\textsuperscript{5}\quad Jing Shao\textsuperscript{3}\quad
\bf Hui Xiong\textsuperscript{1,2}\quad Xuming Hu\textsuperscript{1,2}\thanks{Corresponding author: Xuming HU. Email: xuminghu@hkust-gz.edu.cn}\\
\textsuperscript{1}The Hong Kong University of Science and Technology (Guangzhou)\\
\textsuperscript{2}The Hong Kong University of Science and Technology
\textsuperscript{3}Shanghai AI Laboratory\\
\textsuperscript{4}Nanyang Technological University
\textsuperscript{5}Renmin University of China \\
}

\markboth{Journal of \LaTeX\ Class Files,~Vol.~14, No.~8, October~2024}%
{Shell \MakeLowercase{\textit{et al.}}: A Sample Article Using IEEEtran.cls for IEEE Journals}


\maketitle

\begin{abstract}
The rapid development of Artificial Intelligence (AI) has revolutionized numerous fields, with large language models (LLMs) and computer vision (CV) systems driving advancements in natural language understanding and visual processing, respectively. 
The convergence of these technologies has catalyzed the rise of multimodal AI, enabling richer, cross-modal understanding that spans text, vision, audio, and video modalities. 
Multimodal large language models (MLLMs), in particular, have emerged as a powerful framework, demonstrating impressive capabilities in tasks like image-text generation, visual question answering, and cross-modal retrieval. 
Despite these advancements, the complexity and scale of MLLMs introduce significant challenges in interpretability and explainability, essential for establishing transparency, trustworthiness, and reliability in high-stakes applications.
This paper provides a comprehensive survey on the interpretability and explainability of MLLMs, proposing a novel framework that categorizes existing research across three perspectives: (I) Data, (II) Model, (III) Training \& Inference. 
We systematically analyze interpretability from token-level to embedding-level representations, assess approaches related to both architecture analysis and design, and explore training and inference strategies that enhance transparency. By comparing various methodologies, we identify their strengths and limitations and propose future research directions to address unresolved challenges in multimodal explainability. This survey offers a foundational resource for advancing interpretability and transparency in MLLMs, guiding researchers and practitioners toward developing more accountable and robust multimodal AI systems.
\end{abstract}

\begin{IEEEkeywords}
Multimodal Large Language Models, Explainability, Interpretability, Survey.
\end{IEEEkeywords}

\section{Introduction}
\IEEEPARstart{T}{he} rapid advancement of Artificial Intelligence (AI) has significantly transformed a wide array of fields. 
Recently one of the most influential advancements in AI is the development of large language models (LLMs), which exhibit remarkable language understanding and generating capabilities in a wide range of natural language tasks like text generation, translation, and conversational AI~\cite{zhao2023survey}. 
Similarly, advancements in computer vision (CV) have enabled systems to effectively process and interpret complex visual data, powering tasks like object detection, action recognition, and semantic segmentation with high precision~\cite{voulodimos2018deep}.
More recently, the convergence of these technologies has spurred interest in multimodal AI, which seeks to integrate text, vision, audio, and video for a richer, more comprehensive understanding of multiple modalities~\cite{cui2024survey,jin2024efficient,caffagni2024r,zhang2024mm,wu2024visual, xie2024large,yan2024errorradar,yan2024urbanclip,zheng2024reefknot}.
Multimodal large language models (MLLMs) have experienced rapid advancements, driven by significant improvements in deep learning techniques~\cite{yin2023survey,wu2023multimodal,zou2025deep,song2023bridge,zou2024look,zhou2024mitigating}.
By integrating diverse data sources, MLLMs demonstrate advanced understanding, reasoning, and generative capabilities across a wide range of multimodal tasks, including image-text generation~\cite{huang2021unifying,koh2024generating,hu2024instruct}, visual question answering~\cite{fu2023mme,liu2025mmbench,yue2024mmmu,lu2023mathvista,zhang2024unveiling,li2024seed2plus,li2023seed2,li2023seed}, cross-modal retrieval~\cite{wang2016comprehensive,zhen2019deep,chun2021probabilistic}, video understanding~\cite{abdu2021multimodal,khalid2021fakeavceleb,seo2022end,fan2025videoagent,dzabraev2021mdmmt,botach2022end,luo2020univl}.
Consequently, MLLMs have found diverse applications across various domains~\cite{chen2024evolution,wang2024comprehensive,yan2024georeasoner}, including natural language processing (NLP)~\cite{huang2023chatgpt,ye2023mplug}, CV~\cite{bayoudh2022survey,bayoudh2023survey}, video~\cite{ xu2023multimodal,song2023bridge,tang2023video}, autonomous driving~\cite{nie2025reason2drive,li2024cog,cui2024survey}, medicine \cite{wang2024interpretable,liu2024mcan,xiao2024comprehensive}, and robotics~\cite{li2024mmro,li2021toward,xue2020progress,duan2022multimodal,wake2024gpt,sermanet2024robovqa}. 
However, as the complexity and scale of MLLMs grow, a critical challenge arises: \textit{deciphering the decision-making processes of MLLMs.}~\cite{zhang2024mm,zwilling2019enhanced,zhao2024deep}.

\begin{figure}[tbp]
    \centering
    \includegraphics[width=1\linewidth]{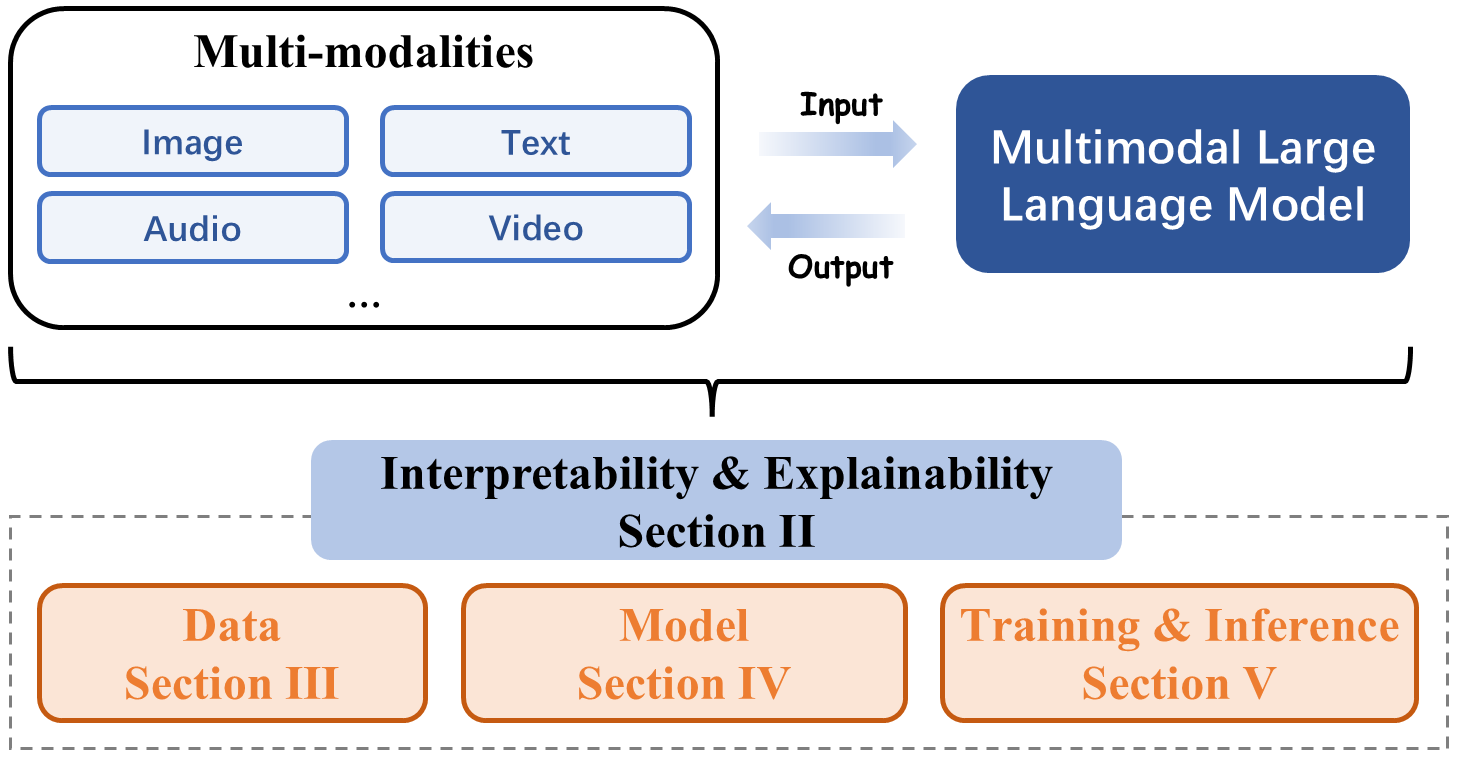}
    \caption{The conceptual framework of this survey. MLLMs handle inputs and outputs that span multiple modalities, such as images, text, video, and audio. We explore interpretability and explainability along three major dimensions: the data, the model, and the training \& inference. }
    \label{fig:skeleton}
\end{figure}

The field of explainable artificial intelligence (XAI) has become pivotal in making the decision-making processes of complex AI systems more transparent and accessible~\cite{arrieta2020explainable,das2020opportunities,adadi2018peeking}.
Interpretability and explainability are defined as the ability to explain or to present in human-understandable terms~\cite{doshi2017towards,du2019techniques}.
Although significant progress has been made in unimodal explainability and interpretability, such as in convolutional neural networks (CNN)~\cite{zeiler2014visualizing,zhang2018interpretable} or transformers~\cite{chefer2021transformer} for images and LLMs~\cite{zhao2024explainability} for text, the multimodal domain presents unique challenges, such as the alignment and decomposition of different modalities. 
Moreover, the interpretability and explainability of MLLMs are essential for ensuring transparency and trustworthiness, especially in high-stakes applications where artificial intelligence decisions have a significant human impact, which addresses how disparate data types are combined within a model and how their interplay affects outputs. 
Following recent research~\cite {doshi2017towards,nauta2023anecdotal,lipton2018mythos}, in this paper, we define \textit{interpretability} in MLLMs refers to the internal structures that are inherently understandable, allowing for straightforward comprehension of how inputs are transformed into outputs. 
\textit{Explainability} in MLLMs, on the other hand, involves post-hoc techniques that provide external analyses of model behavior behind a model's decisions. 

In this paper, we present a new insight for categorizing the interpretability and explainability of MLLMs by integrating perspectives from data, model, training and inference. 
As illustrated in Figure~\ref{fig:skeleton}, we examine the interpretability and explainability of MLLMs from three perspectives: data (\cref{section-data-overall}), model (\cref{section-model}), training and inference (\cref{section-training-inference}).
Following the data-driven explainability~\cite{kaur2020interpreting,zhou2008low,guillaume2001designing,frye2020shapley} research, we examine the data perspective (\cref{section-data-overall}), exploring how both input and output data can be attributed to the model's decisions. We also analyze benchmarks and applications to evaluate the trustworthiness and reliability across various tasks, thereby ensuring the robustness and applicability in real-world scenarios~\cite{hooker2019benchmark,ismail2020benchmarking}.
As for the model interpretability and explainability~\cite{chakraborty2017interpretability,zhang2021survey,fan2021interpretability,linardatos2020explainable,marcinkevivcs2020interpretability,dovsilovic2018explainable,gilpin2018explaining}, from the model perspective (\cref{section-model}), we conduct an in-depth analysis at the token level, embedding level, neuron level, layer level, and architectural level. 
At the token level~\cite{madsen2022post,bibal2016interpretability,chen2016infogan, zou2310representation, qian2024towards}, we investigate the impact of individual tokens on model outputs and explore methodologies to enhance interpretability. 
At the embedding level~\cite{zhang2018visual}, we assess how multimodal embeddings influence the performance and interpretability of MLLMs, providing a deeper understanding of the underlying representation mechanisms.
For neuron-level~\cite{sajjad2022neuron,bai2024improving, qian2024dean}, we analyze individual units and specialized groups of neurons to comprehend their contributions to overall model behavior. 
At the layer level~\cite{chakraborty2017interpretability,zhang2018interpretable,ren2024identifying}, we investigate how different layers affect decision-making processes within the model. 
Regarding architecture, we differentiate between architecture-analysis and architecture-design~\cite{ribeiro2016model,danesh2023hybridization,rumbelow2024model,early2022model} approaches to interpretability, highlighting strategies that promote transparency and facilitate a better understanding of model operations.
Furthermore, we explore training and inference strategies that enhance model transparency and explainability (\cref{section-training-inference}). 
In the training phase~\cite{zhang2021survey}, we summarize how various training mechanisms and weight adjustments impact the interpretability of MLLMs. 
We discuss techniques aimed at improving alignment, reducing hallucinations, and promoting the acquisition of core knowledge and generalization capabilities in MLLMs. During inference, we investigate methods to mitigate issues such as hallucination without the need for retraining, including over-trust penalty mechanisms and chain-of-thought reasoning techniques. 

By integrating these perspectives~\cite{li2024survey,cui2024survey,han2022survey}, our survey offers a holistic understanding of the challenges and advancements in the interpretability and explainability of MLLMs. We believe this comprehensive analysis will serve as a valuable resource for researchers and practitioners dedicated to developing more transparent, reliable, and trustworthy multimodal models.
The major contributions of this work are summarized as follows:
\begin{itemize}
    \item We are the first to offer an in-depth and comprehensive review of existing research on the explainability and interpretability of MLLMs.
    \item We present a structured and comparative analysis of current methods for MLLMs explainability and interpretability, introducing a novel categorization that organizes these methods into data, model, training \& inference perspectives.
    \item We highlight potential research directions that could advance the field, offering valuable guidance for researchers aiming to further develop explainability and interpretability approaches for MLLMs.
\end{itemize}

\section{Survey Landscape}

\subsection{Survey Scope}
Both multimodal models and XAI have achieved significant advancements in recent years, with a substantial body of research exploring methods for making these complex models more transparent and interpretable~\cite{kaur2020interpreting,zhou2008low,guillaume2001designing}. To narrow the scope of this survey to a manageable range, we focus on the explainability and interpretability of MLLMs. Interpretability of MLLMs refers to internal structures that are inherently understandable, allowing for straightforward insights into how inputs are processed and transformed into outputs~\cite{chakraborty2017interpretability,zhang2021survey}. Interpretable MLLMs enable researchers and practitioners to gain deeper insights into these cross-modal dynamics, providing clarity on how each modality influences and shapes the model’s decision-making processes~\cite{zhang2018visual}. Explainability involves using external techniques to clarify the reasons behind a model's decisions, which is essential in MLLMs for understanding the intricate interactions among multiple modalities~\cite{ribeiro2016model}. This focus not only enhances our understanding of multimodal integration but also addresses the growing demand for transparency in complex AI systems~\cite{zhang2021survey}.

In this survey, we concentrate on four main dimensions of explainability and interpretability within MLLMs: Data Explainability – How input data from different modalities is preprocessed, aligned, and represented to support interpretability across modalities, and how causal attribution methods are applied to outputs to enhance understanding of model decisions~\cite{kaur2020interpreting,frye2020shapley}. Model Explainability – Techniques that elucidate the structure and functioning of the multimodal model itself, offering insights into how neurons, layers, and architectures contribute to interpretability~\cite{chakraborty2017interpretability,zhang2021survey,fan2021interpretability,ribeiro2016model,sajjad2022neuron,zhang2018interpretable,zhang2018visual,bibal2016interpretability,chen2016infogan,madsen2022post}. Training and Inference Explainability – Understanding how MLLMs' training and inference processes affect interpretability, which is vital for refining transparency during both the learning phase and real-world application.
To maintain focus, we exclude single-modality explainability methods from the main scope of this survey, such as Transformer interpretability, CNN interpretability or LLMs interpretability, except for brief background information. 
Similarly, general approaches to explainability that do not address the unique challenges of multimodal interactions are outside the primary scope of this review. 
Instead, our emphasis remains on methods and models explicitly designed to interpret and explain the interactions between multiple modalities.

\subsection{Survey Methodology}
To provide a comprehensive overview of explainability and interpretability in MLLMs, we conducted an extensive review of research papers spanning the fields of machine learning, NLP, CV, and multimodal systems. 
We examined papers published over the past decade (2010–2024), focusing on the growing body of work that explores interpretability and explainability in these areas. 
Our methodology consisted of several key steps. 
First, we searched for papers using keywords such as ``multimodal large models," ``interpretability," and ``explainability" in databases like Google Scholar, detailed is shown in \cref{keywords-MLLMs}.
To further ensure the completeness of our survey, we also reviewed reference lists of key papers and included early influential works that shaped this domain. 
After collecting the candidate papers, we followed a multi-step filtering process. 
Titles were first reviewed to identify potentially relevant papers, followed by abstract screening to confirm relevance. 
In cases where the title and abstract were insufficient for a decision, we reviewed the full text. 
As is shown in Figure~\ref{fig:fusion tree}, the final selection covers a variety of interpretability and explainability techniques applied to MLLMs, including input-output analyses, model components, and training dynamics.

\begin{table}[t]
\centering
\renewcommand{\arraystretch}{1.2}
\caption{Keywords for our paper search query.}
\begin{tabular}{c|>{\centering\arraybackslash}p{6.5cm}} 
\hline
 & \textbf{Explainable and Interpretable MLLMs}  \\
\hline
\textbf{Keywords} & XAI, explainable AI, explanation, explainable, explanatory, interpretable, intelligible, black-box, white-box, explainability, interpretability, intelligibility, text-to-image, image-to-text, diffusion, GAN, CLIP, MLLMs, VLMs, VQA \\
\hline
\end{tabular}
\label{keywords-MLLMs}
\end{table}

\section{Data}
\label{section-data-overall}
\begin{figure*}[t!]
    \centering
    \begin{adjustbox}{width=0.95\textwidth}
        \begin{forest}
            for tree={
                grow=east,
                reversed=true,
                anchor=base west,
                parent anchor=east,
                child anchor=west,
                base=center,
                font=\large,
                rectangle,
                draw=hidden-draw,
                rounded corners,
                align=left,
                text centered,
                minimum width=5em,
                edge+={darkgray, line width=1pt},
                s sep=3pt,
                inner xsep=2pt,
                inner ysep=3pt,
                line width=0.8pt,
                ver/.style={rotate=90, child anchor=north, parent anchor=south, anchor=center},
            },
            where level=1{text width=10em,font=\normalsize,}{},
            where level=2{text width=12em,font=\normalsize,}{},
            where level=3{text width=10em,font=\normalsize,}{},
            where level=4{text width=7em,font=\normalsize,}{},
            where level=5{text width=7em,font=\normalsize,}{},
            [MLLMs [Data 
                [Input and Output
                    [\cite{park2018multimodal}{, }\cite{leng2024mitigating}{, } \cite{walker2023causal}{, }\cite{klaassen2024doublemldeep}{, }\cite{liang2022high}, leaf, text width=19em]]
                [Benchmark 
                    [\cite{li2024survey}{, }\cite{huang2024survey}{, }\cite{tasdizen2024vista}{, }\cite{cai2023benchlmm}{, }\cite{dang2024exploring}{, }\cite{zhang2024benchmarking}{, } \cite{hu2023tifa}{, }\cite{verma2024cross}{, }\\ \cite{tiong2024we}{, } \cite{alipour2020study}{, }\cite{liu2020deep}{, }\cite{lu2022learn}, leaf, text width=19em]]
                [Application 
                    [\cite{xie2024large}{, }\cite{shaham2024multimodal}{, }\cite{cuadra2024digital}{, }\cite{liu2022make}{, }\cite{wang2024interpretable}{, }\cite{kanehira2019multimodal}{, } \cite{zang2023discovering}{, }\cite{ko2023large}{, }\cite{zhang2024holmes}{, }\\\cite{sanders2024tv}{, }\cite{cui2024survey}{, }\cite{hu2019multi}{, }\cite{xu2024drivegpt4}{, }\cite{li2024cog}{, }\cite{liu2022group}{, }\cite{zadeh2018multimodal}{, }\cite{sotirou2024musiclime}{, }\cite{amara2024enhancing}, leaf, text width=19em]]
            ]
            [Model
                [Token 
                    [Visual Token
                        [\cite{gandelsman2023interpreting}{, }\cite{neo2024towards}{, }\cite{zhang2024redundancy}{, }\cite{chen2024image}{, }\cite{wang2023visual}{, }\cite{yao2024deco}, leaf, text width=14em]]
                    [Visual-Text Token                       
                        [\cite{bi2023vl}{, }\cite{zhao2024first}{, }\cite{li2024unified}{, }\cite{dai2022plausible_hjh3}{, }\cite{huang2024opera}, leaf, text width=14em]]
                ]
                [Embedding
                    [Visual Embedding
                        [\cite{verma2024cross}{, }\cite{shi2024eagle}, leaf, text width=14em]]
                    [Textual Embedding
                        [\cite{hendricks2021probing}{, }\cite{moayeri2023text}, leaf, text width=14em]]
                    [Cross-Modal Embedding
                        [\cite{derby2018using}{, }\cite{chen2023stair}{, }\cite{dominici2023sharcs}{, }\cite{bhalla2024interpreting}{, }\cite{gandelsman2023interpreting}{, }\cite{frank2021vision}{, }\\ \cite{wangfreebind2024}{, }\cite{parekh2024concept}{, }\cite{evirgen2024text}{, }\cite{salin2022vision}{, }\cite{ramesh2022investigation}, leaf, text width=14em]]
                ]
                [Neuron 
                    [Indvidual Units
                         [\cite{goh2021multimodal}{, }\cite{gandelsman2024interpreting}, leaf, text width=14em]]
                    [Specialization Group
                        [\cite{pan2023finding}{, }\cite{huo2024mmneuron}{, }\cite{Huang2024MINERMT}, leaf, text width=14em]]
                ]
                [Layer 
                    [Individual Components
                        [\cite{cao2020behind}{, }\cite{gandelsman2023interpreting}{, }\cite{quantmeyer2024and}{, }\cite{xu2023bridging}, leaf, text width=14em]]
                    [Decision-Making \\ Workflow 
                        [\cite{xu2023bridging}{, }\cite{pan2023finding}{, }\cite{huo2024mmneuron}{, }\cite{zhang2024redundancy}{, }\cite{tao2024probing}{, }\cite{nguyen2019multi}, leaf, text width=14em]]
                ]
                [Architecture
                    [Architecture Analysis
                        [\cite{lyu2022dime}{, }\cite{goyal2016towards}{, }\cite{wu2018faithful}{, }\cite{natarajan2024vale}{, }\cite{shaham2024multimodal}{, }\cite{stan2024lvlm}{, }\\ \cite{lee2023diffusion}{, }\cite{yang2024law}, leaf, text width=14em]]
                    [Architecture Design
                        [\cite{wan2020nbdt}{, }\cite{yang2023language}{, }\cite{guo2024trace}{, }\cite{li2024multimodal}{, }\cite{swamy2024multimodn}{, }\cite{shen2023scaling}{, }\\ \cite{wen2023imkga}, leaf, text width=14em]]
                ]
            ]
            [Training and Inference
                [Training                   
                    [\cite{cao2020behind}{, }\cite{zang2024pre}{, }\cite{zhang2024disttrain}{, }\cite{neo2024towards}{, }\cite{sun2023aligning}{, }\cite{yan2024vigor}{, }\cite{yu2024rlhf}{, }\\\cite{tsai2020multimodal}{, }\cite{mallick2024ifi}{, }\cite{zhao2023beyond}{, }\cite{yu2024rlaif}{, }\cite{zhou2024aligning}{, }\cite{dai2022plausible_hjh3}{, }\cite{huang2024opera}, leaf, text width=16em]]
                [Inference 
                    [\cite{liu2024survey}{, }\cite{bai2024hallucination}{, }\cite{huang2024opera}{, }\cite{leng2024mitigating}{, }\cite{zhang2023multimodal}{, }\cite{ge2023chain}{, }\cite{yao2023thinking}, leaf, text width=16em]]
            ]
            ]  
        \end{forest}
    \end{adjustbox}
    \vspace{-2mm}
    \caption{We classify MLLM explainability into three main categories: Data, Model, and Training \& Inference. This structure facilitates a comprehensive overview of the various techniques used to explain MLLMs, along with a discussion of the methods for evaluating these explanations across different paradigms.}
    \label{fig:fusion tree}
\end{figure*}
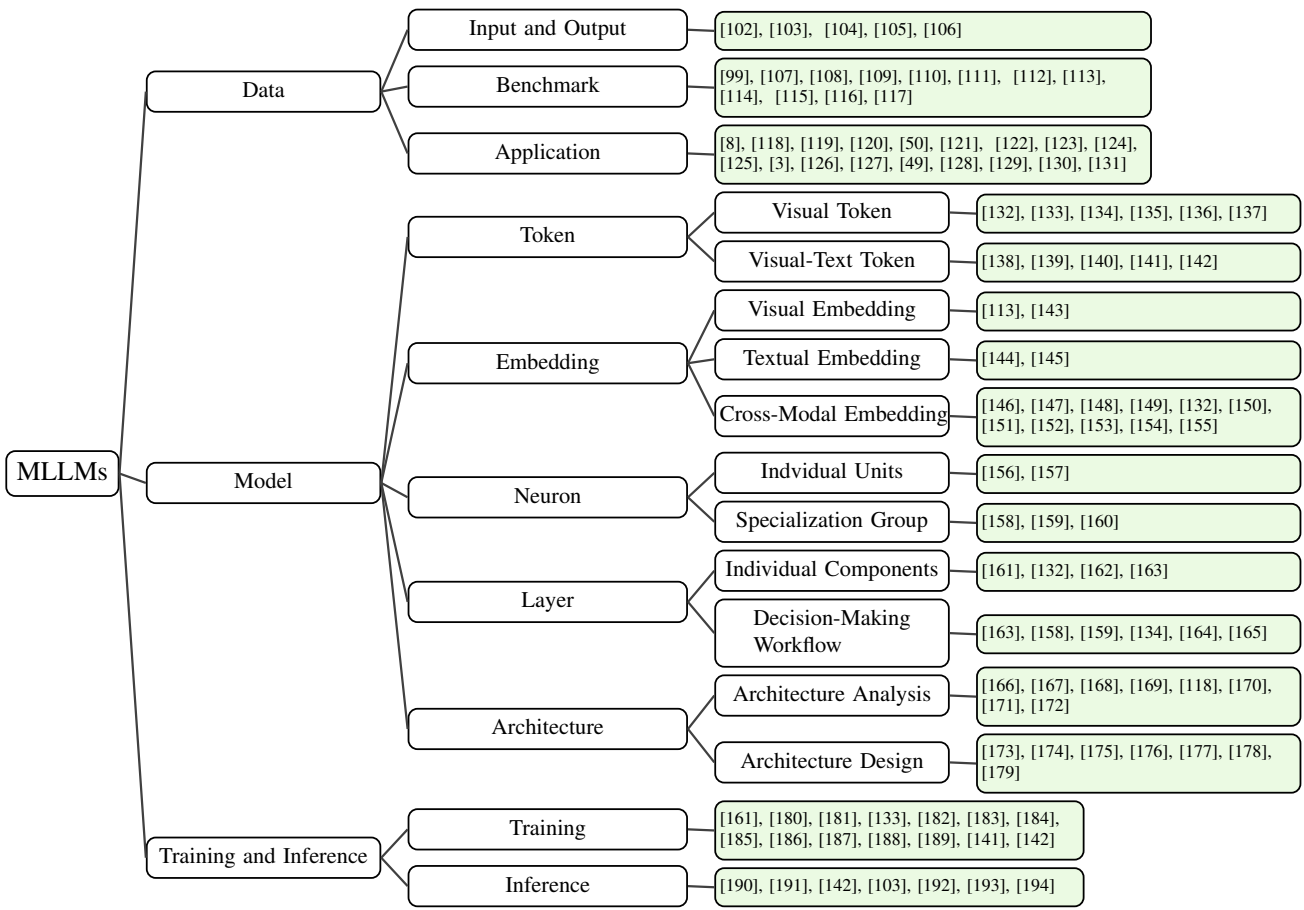

LLMs primarily focus on processing text inputs at the levels of words, phrases, or sentences~\cite{zhao2024explainability}. 
Explainability in LLMs involves understanding how these models interpret input text data and generate interpretable text data~\cite{zhao2024explainability}. 
In contrast, explainability in computer vision typically relies on models like CNNs~\cite{zhang2021survey} or Vision Transformers (ViTs)~\cite{kashefi2023explainability,han2022survey} to analyze and interpret visual image data. 
MLLMs extend these capabilities by integrating visual, audio and language information, enabling the generation and understanding of multimodal data. 
In this section, we mainly explore the role of data in enhancing the interpretability of MLLMs. 
As is shown in Figure~\ref{fig:fusion tree}, we categorize these works into three groups:
\begin{itemize} 
    \item \textbf{Input and Output} (\cref{section-data-in/outputs}): Focuses on methods to improve interpretability by analyzing how models process inputs and outputs, including techniques like perturbation, saliency maps, and causal inference.
    \item \textbf{Benchmarks} (\cref{section-data-benchmark}): Highlights benchmarks, datasets, and metrics for evaluating interpretability and robustness in multimodal models.
    \item \textbf{Applications} (\cref{section-more-modalities}): Explores interpretability techniques applied to domains beyond vision and language, such as audio, video, autonomous driving, and medicine.
\end{itemize}

\definecolor{mycolor}{RGB}{215, 245, 200}

\tikzstyle{my-box}=[
    rectangle,
    draw=hidden-draw,
    rounded corners,
    text opacity=1,
    minimum height=1.5em,
    minimum width=5em,
    inner sep=2pt,
    align=center,
    fill opacity=.5,
    line width=0.8pt,
]
\tikzset{
leaf/.style={
my-box,
minimum height=1.5em,
fill=mycolor, 
text=black,
align=left,
font=\footnotesize,
inner xsep=2pt,
inner ysep=4pt,
line width=0.8pt
}
}

\subsection{Input and Output}
\label{section-data-in/outputs}
The robustness and explainability of MLLMs are critically dependent on how these models handle input data and produce outputs, as well as the transparency of their decision-making processes. 
Early works highlighted explainability in model input processing, zintgraf et al.~\cite{zintgraf2017visualizing} investigated the influence of specific regions within input images on model classification and revealed how different image regions contribute to predictions to make the decision-making processes of deep networks more interpretable.
Park et al.~\cite{park2018multimodal} developed a multimodal explainability framework combining visual attention maps with textual justifications, enhancing model transparency across visual and textual inputs.
Szegedy~\cite{szegedy2013intriguing} showed that deep neural networks could be highly sensitive to small, almost imperceptible changes in input data. 

Furthering the exploration of input explainability, methods such as TISE~\cite{petsiuk2018rise} and Extremal Perturbations~\cite{fong2019understanding} developed perturbation-based approaches to create saliency maps. 
These maps highlight critical areas in input images that significantly affect model predictions, thus providing interpretable explanations by revealing which input features are most influential in model decision-making. 
Complementing these methods, Kanehira~\cite{kanehira2019learning} proposed a novel framework for generating visual explanations that combine linguistic and visual information. By maximizing the interaction between modalities, highlighting how complementary information from different modalities influences model decisions. 
More recently, Fel et al.~\cite{fel2024holistic} introduced a unified theoretical framework for concept-based explainability, formalizing concept extraction as a process of dictionary learning and concept importance estimation as an attribution method.

Beyond these methods, causal inference has emerged as a crucial approach for uncovering meaningful relationships in multimodal data. Morioka~\cite{morioka2023connectivity} introduced connectivity-contrastive learning (CCL), a framework for causal discovery in multimodal contexts. CCL disentangles mixed observations into independent latent components and identifies their causal structures, thus enhancing interpretability by providing insight into the underlying causal relationships in multimodal data. 
Within a similar context, CausalPIMA~\cite{walker2023causal} presented a causal representation learning algorithm that integrates multimodal data and physics-based constraints. CausalPIMA employs a differentiable directed acyclic graph (DAG) learning structure with a variational autoencoder to discover essential causal relationships in an unsupervised manner, enabling interpretable causal patterns without predefined causal assumptions.
Further advancing causal learning, Klaassen et al.~\cite{klaassen2024doublemldeep} proposed a neural network architecture within the double machine learning (DML) framework, aimed at causal inference with unstructured data, such as text and images.

Recent works~\cite{yuan2024diffusion,luo2022understanding,yang2023diffusion,jeanneret2022diffusion} in diffusion models have provided new methods for interpretability, particularly through pixel-level attribution and information-theoretic approaches. DAAM~\cite{tang2022daam} improved the interpretability of large-scale diffusion models by generating attribution maps that use cross-attention scores, offering insight into how specific words influence image regions. This method reveals complex syntactic and semantic relationships in image generation, such as feature entanglement between objects and descriptions. 
Liang et al.~\cite{liang2022high} proposed an efficient high-modality learning method that uses information-theoretic metrics to measure modality and interaction heterogeneity, improving multimodal model explainability in complex tasks.
Building on this, Kong et al.~\cite{kong2023interpretable} introduced an information-theoretic approach to enhance explainability in denoising diffusion models. 



\begin{figure*}[!t]
\centering
\includegraphics[width=0.8\linewidth]{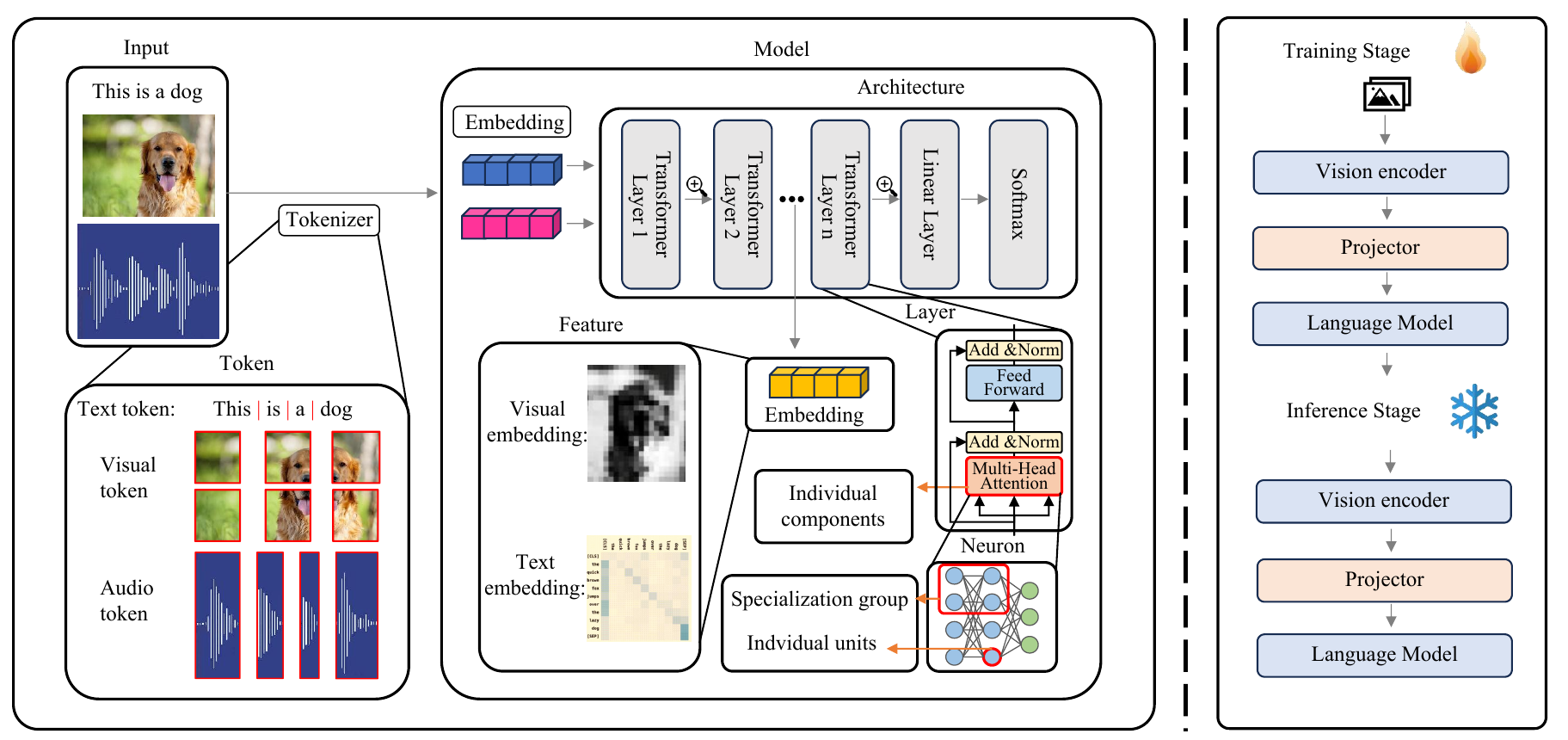}
\caption{\textbf{Overview of our framework.} The framework illustrates how input modalities like images, videos, or audio are tokenized into visual or textual tokens and then transformed into embeddings. The architecture includes individual neurons and neuron groups across layers, analyzed through architecture analysis and design. The workflow concludes with training and inference phases.}
\label{fig:overall-framework}
\end{figure*}

\subsection{Benchmark}
\label{section-data-benchmark}

Recent advancements in MLLMs have provided transformative insights into processing and aligning visual and textual data~\cite{li2024survey,huang2024survey}. As these models become central to diverse applications, understanding their decision-making is crucial for transparency, trust, and robustness. This paper explores benchmarks, evaluation frameworks, and interpretability methods to address challenges in alignment, robustness, and domain-specific explainability, emphasizing the importance of explainability in enhancing the reliability of MLLM datasets.

\textbf{Alignment and Robustness.} Efforts to improve transparency in visual-linguistic alignment have introduced new benchmarks and datasets. VISTA~\cite{tasdizen2024vista} aligned with human visual attention data, which compares internal heatmaps of vision-language models with human attention patterns to enhance model trustworthiness. Addressing robustness under distribution shifts, Cai et al.~\cite{cai2023benchlmm} developed BenchLMM, enabling models to detect image styles and explain errors under stylistic variations. Similarly, Mao et al.~\cite{mao2023coco} introduced COCO-O, designed to evaluate the robustness of object detectors against natural distribution shifts, underscoring explainability's role in identifying vulnerabilities. Broadening this perspective, the Multimodal Uncertainty Benchmark (MUB)~\cite{dang2024exploring} assessed the vulnerability of MLLMs to explicit and implicit misleading instructions. Zhang et al.~\cite{zhang2024benchmarking} further developed MultiTrust, a comprehensive evaluation framework spanning dimensions like truthfulness, safety, robustness, fairness, and privacy, revealing cross-modal explainability challenges. 

\textbf{Image-Text Tasks.} In image-text tasks, Madhyastha et al.~\cite{madhyastha2018defoiling} demonstrated the impact of explicit object information in identifying semantically incorrect captions, highlighting the importance of accurately encoding descriptive image features to enhance explainability. In language representation evaluation, Hewitt and Liang~\cite{hewitt2019designing} introduced control tasks to verify whether linguistic probes genuinely capture underlying structures or merely memorize tasks, underscoring the need for selectivity in explainability assessments. In text-to-image generation, Hu et al.~\cite{hu2023tifa} proposed TIFA, an evaluation metric that uses visual question answering (VQA) to assess the faithfulness of generated images. By correlating model accuracy on generated question-answer pairs with human judgments, TIFA offers fine-grained assessments for explainability. Further exploring domain-specific explainability, Verma et al.~\cite{verma2024cross} examined the impact of visual attributes on model behavior, while Tiong et al.~\cite{tiong2024we} introduced six explainability factors to evaluate how vision-language models represent basic concepts.

\textbf{Task-Specific Multimodal explainability.} In the context of VQA, Alipour et al.~\cite{alipour2020study} demonstrated that multimodal explanations enhance user accuracy, confidence, and understanding, especially when the model's response is inaccurate. Their introduction of active attention as a novel approach for examining causal effects highlights the role of transparency in building trust. For zero-shot learning, Liu et al.~\cite{liu2020deep} developed the explainable zero-shot learning (XZSL) framework, which integrates visual and textual explanations into classification decisions via the Deep Multi-Modal Explanation (DME) model. Lastly, ScienceQA~\cite{lu2022learn}, a multimodal dataset, promotes explainability through chain-of-thought reasoning, enabling structured lectures and explanations to improve both question-answering performance and learning efficiency.
Schwettmann et al.~\cite{schwettmann2024find} introduced FIND, a benchmark suite for automated explainability, generating and validating descriptions of black-box functions to improve understanding of neural network behavior.

\subsection{Application}
\label{section-more-modalities}
In recent years, the growing complexity of multimodal AI models has emphasized the need for explainability, which can provide insights into more applications across domains like medical, video processing, autonomous driving, audio processing, and music~\cite{li2024survey,huang2024survey,yin2023survey}. This survey reviews state-of-the-art explainability techniques for multimodal AI applications, highlighting advancements, challenges, and methodologies that address various needs for model transparency.

\textbf{Agent Explainability.} Xie et al. ~\cite{xie2024large} explored a systematic review of large multimodal agents (LMAs), focusing on the essential components, categorization, collaborative frameworks, evaluation methodologies, and real-world applications while highlighting key advancements in multimodal explainability and proposing future research directions.
Explainability agents are pivotal in analyzing model behaviors, automating feature interpretation, and identifying failure modes. MAIA~\cite{shaham2024multimodal} represented a multimodal automated explainability agent aimed at explaining vision-language models and deep networks. This agent uses neural models for feature interpretation and discovery of failure modes, enhancing the understanding of complex model behaviors in multimodal settings. Similarly, Cuadra et al.~\cite{cuadra2024digital} introduced a multimodal LLMs agent to improve accessibility in digital form completion, especially for older adults and individuals with sensory impairments, thereby tailoring explainability to enhance usability and inclusivity in human-computer interaction.


\textbf{Medical Explainability.} Explainability in multimodal AI models is crucial in healthcare, where transparency can directly impact clinical decisions.
UnitedNet~\cite{tang2023explainable} is an explainable multi-task deep neural network designed for biological data, which reveals relationships between gene expression and other modalities, supporting explainable biological data analysis. Rawls et al.~\cite{rawls2021integrated} employed Causal Discovery Analysis (CDA) to model Alcohol Use Disorder (AUD) pathways, underscoring the influence of cognitive, social, and psychiatric factors on AUD severity. Furthermore, MCAN~\cite{liu2024mcan} bridged fMRI and EEG data using a Multimodal Causal Adversarial Network, facilitating dynamic brain network structure estimation and revealing abnormal activity patterns. 
Amara et al.~\cite{amara2024enhancing} proposed a novel framework combining ontologies with MLLMs to enhance explainability in domain-specific tasks, using ontology-based guidance and evaluation to improve model alignment with domain concepts, particularly for plant disease classification.
MedRegA~\cite{wang2024interpretable} introduced a region-aware medical multimodal language model that aligns with clinical workflows by enabling region-specific identification and report generation across modalities, thereby enhancing explainability in clinical practice.

\textbf{Video Explainability.} The field of video analysis benefits from interpretable multimodal models that clarify decisions in complex visual-linguistic tasks. Kanehira et al.~\cite{kanehira2019multimodal} proposed a counterfactual explanation method for video classification, enhancing interpretability by improving visual-linguistic compatibility and understanding. In VideoQA, Zang et al.~\cite{zang2023discovering} developed the Multimodal Causal Reasoning (MCR) framework, which separates causal and confounding features to improve robustness in answering video-related questions. Similarly, Flipped-VQA\cite{ko2023large} addresses linguistic bias in LLMs by predicting reciprocal video, question, and answer pairs, enhancing VideoQA explainability. Holmes-VAD\cite{zhang2024holmes} tackles video anomaly detection by using a multimodal dataset with single-frame annotations, facilitating detailed anomaly explanations. TV-TREES~\cite{sanders2024tv} contributes an entailment tree generator for logical video-language understanding, allowing human-interpretable proofs and achieving state-of-the-art performance in zero-shot scenarios on the TVQA benchmark.

\textbf{Autonomous driving Explainability.}  In autonomous driving, explainability is essential for understanding complex decision-making processes and ensuring safety~\cite{cui2024survey}. Hu et al.~\cite{hu2019multi} proposed a probabilistic multimodal method for predicting vehicle behavior, addressing uncertainties and enhancing explainability. DriveGPT4~\cite{xu2024drivegpt4} processes video inputs to provide natural language explanations and low-level vehicle controls, improving understanding of autonomous driving systems. Reason2Drive~\cite{nie2025reason2drive} introduced a novel dataset with chain-based reasoning metrics, clarifying decision-making. Cog-GA~\cite{li2024cog} added cognitive mapping and dual-channel scene descriptions to support interpretable vision-language navigation, offering transparency in scene understanding and adaptive planning for autonomous driving.

\textbf{Audio Explainability.}  Audio processing models require interpretability to effectively recognize emotional and contextual cues. GBAN~\cite{liu2022group} employs a gated bidirectional alignment network to align speech and text modalities, enhancing both explainability and emotional recognition accuracy. Qwen-Audio~\cite{chu2023qwen} extended audio-language model capabilities with multi-turn dialogue support, improving explainability in audio-centered scenarios. Multimodal Attention Merging (MAM)~\cite{sundar2024multimodal} facilitated knowledge transfer from text and image models to audio models without additional fine-tuning, while Jalal et al.~\cite{jalal2020empirical} use attention models in speech emotion recognition to map vowel and word cues, revealing emotional patterns and improving the explainability of acoustic-based models. Zadeh et al.~\cite{zadeh2018multimodal} presented the CMU-MOSEI dataset for multimodal sentiment and emotion recognition and introduced the Dynamic Fusion Graph (DFG), which enables detailed analysis of cross-modal interactions by visualizing the interactions between language, visual, and acoustic modalities.

\textbf{Music Explainability.} In music information retrieval, interpretability improves the analysis of complex audio features, enhancing transparency in music classification tasks. Won et al.~\cite{won2019toward} developed a self-attention-based model for music tagging, visualizing attention maps to capture dependencies between musical components. Lyberatos et al.~\cite{lyberatos2024perceptual} combined perceptual feature extraction with explainability techniques like SHAP to clarify ambiguous labels in music tagging. PECMAE\cite{alonso2024leveraging} used a prototype-based model with a diffusion decoder for music classification, enabling explainability in genre and instrument detection. Concept-based methods by Foscarin et al.~\cite{foscarin2022concept} provided the post-hoc explanations that relate high-level musical concepts to model predictions, facilitating musicological analysis. Finally, MUSICLIME~\cite{sotirou2024musiclime} provided model-agnostic explanations in multimodal music models, showing how audio and lyrical features contribute to predictions for a well-rounded understanding of model decision-making.



\section{Model}
\label{section-model}
This section delves into the mechanisms underpinning MLLMs, exploring how their internal representations are interpreted, components such as tokens, embedding, neurons, and layers are analyzed, and architecture is understood.  As is shown in Figure~\ref{fig:fusion tree}, the discussion is structured as follows:

\textbf{Token Interpretability} (\cref{section-representation-token}): examines interpretability at the token level, focusing on visual, textual, and visual-textual tokens.
\begin{itemize} 
    \item \textbf{Visual Tokens} (\cref{section-representation-visual-token}): Explores their role in decision-making, focusing on methods like basis decomposition, attention mechanisms, and token redundancy reduction.
    \item \textbf{Visual-textual Tokens} (\cref{section-representation-visual-text-token}): Explores visual-textual alignment, mitigating hallucination, and improving visual-language integration.
\end{itemize} 

\textbf{Feature Interpretability} (\cref{section-representation-feature}): Focuses on coarse-grained analyses of multimodal embeddings and latent spaces.
\begin{itemize} 
    \item \textbf{Visual Embeddings} (\cref{section-representation-visual-embedding}): Explores human-understandable visual embeddings, internal representations, and dynamic processes in generative models.
    \item \textbf{Visual-Textual Embeddings} (\cref{section-representation-cross-modal-embedding}): Discusses interpretable cross-modal embeddings and techniques for improving alignment and representational capacity. 
\end{itemize} 

\textbf{Neuron Interpretability} (\cref{section-model-neuron}): Investigates the interpretability of individual neurons in multimodal models.
    \begin{itemize} 
    \item \textbf{Individual Units} (\cref{section-Individual Units}):  Explores the roles and semantic concepts of individual neurons in MLLMs.
    \item \textbf{Specialization Groups} (\cref{section-Specialization Groups}): Highlights neuron groups specialized in cross-modal or domain-specific tasks.
    \end{itemize}
    
\textbf{Layer Interpretability} (\cref{section-model-layer}): Analyzes the role of layers in neural networks and their decision-making processes.
    \begin{itemize}
        \item \textbf{Individual Components} (\cref{section-representation-visual-embedding}): Examines the functions of attention heads, MLP layers, and other components.
        \item \textbf{Decision-Making Workflow} (\cref{section-representation-visual-embedding}): Tracks representation transformations and information flow across layers.
    \end{itemize}
    
\textbf{Architecture Interpretability} (\cref{section-model-arch}): Explores model architecture as a whole to explain decision-making processes.

\begin{itemize}  
    \item \textbf{Architecture Analysis} (\cref{section-model-arch-analysis}): We present methods that analyze a model's characteristics or explainability. Most of these methods typically offer visual or textual explanations. We also use the type of explanation as a categorization criterion to organize these methods.
    \item \textbf{Architecture Design} (\cref{section-model-arch-design}): These methods primarily focus on designing specific modules or entire frameworks to enhance the inherent explainability of the model architecture. Typically, they do not provide explicit explanations. We categorize these methods based on their distinct characteristics.
\end{itemize}


\begin{figure*}[!t]
\centering
\resizebox{0.7\linewidth}{!}{%
    \includegraphics{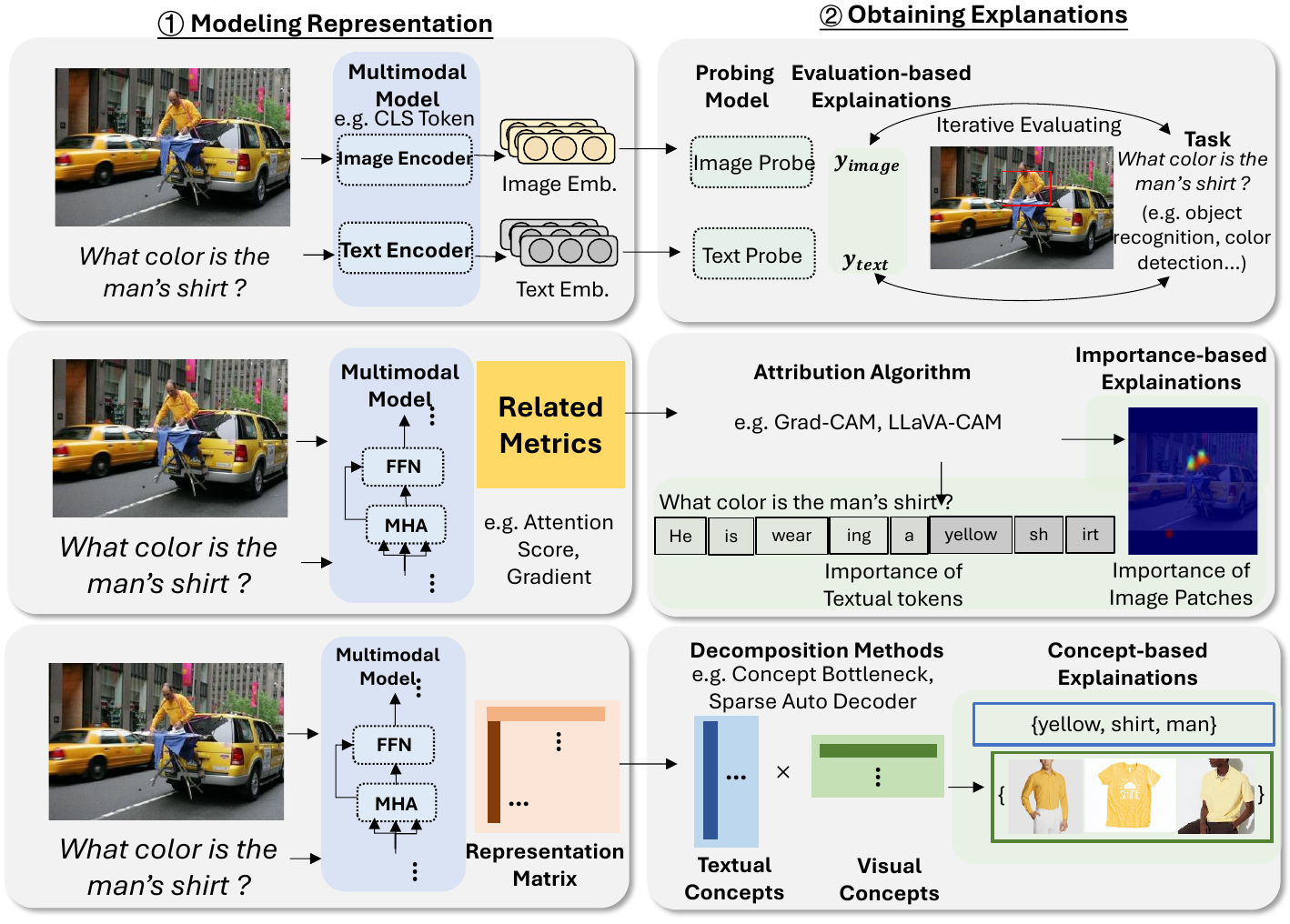}
}
\caption{
\textbf{Illustration of three key methodologies for embedding interpretability.} Probing-based Interpretation: Evaluates representation quality by training a probing model; its performance reflects the utility of the representations for specific tasks.  Attribution-based Interpretation: Assesses input contributions to model outputs using metrics like attention scores or gradients.  
Decomposition-based Interpretation: Analyzes representations by breaking them into meaningful features, often using sparse auto-encoders or similar tools.
}
\label{fig:rep}
\end{figure*}

\begin{table*}[h]
    \centering
    \caption{\textbf{Overview of Embedding-Level Methods.} The papers are categorized by perspective, models, interpretation methods, and tasks.}
    \begin{tabular}{%
        p{0.1\textwidth}%
        p{0.15\textwidth}%
        p{0.06\textwidth}%
        p{0.25\textwidth}%
        p{0.2\textwidth}%
    }
    \toprule
    Perspective & Method & Models & Interpretation Method & Tasks \\
    \midrule
    \multicolumn{5}{c}{\textit{Token Interpretability}} \\
    \midrule
    \multirow{5}{*}{Visual} & Wang et al.~\cite{wang2023visual} & CLIP & Information Bottleneck & Image Captioning \\
    & Neo et al.~\cite{neo2024towards} & MLLMs & Visual Token Localization & Object Identification \\
    & Zhang et al.~\cite{zhang2024redundancy} & MLLMs & Information Flow Analysis & VQA, Image Captioning \\
    & Chen et al.~\cite{chen2024image} & MLLMs & Pruning Irrelevant Visual Tokens & Image, Video Understanding \\    
    & Yao et al.~\cite{yao2024deco} & MLLMs & Understanding Projector Module & VQA  \\
    \midrule
    \multirow{8}{*}{Visual-Textual} & Gandelsman~\cite{gandelsman2023interpreting} & CLIP & Patch, Attention Decomposition & Zero-Shot Image Segmentation \\
    & Bi et al.~\cite{bi2023vl} & MLLMs & Token Detection, Hard Negatives & Image-Text Matching \\
    & Zhao et al.~\cite{zhao2024first} & MLLMs & First-Token Logit Analysis & Logit Analysis \\
    & Li et al.~\cite{li2024unified} & MLLMs & Patch-Level Alignment Metric & Cross-Modal Retrieval \\
    & Dai et al.~\cite{dai2022plausible_hjh3} & MLLMs & Token Alignment & Hallucination \\
    & Huang et al.~\cite{huang2024opera} & MLLMs & Penalizing Overconfidence & Confidence Calibration \\
    & Wei and Zhang~\cite{wei2024dopra} & MLLMs & Dynamic Token Penalization & Token Penalization \\
    & Tang et al.~\cite{tang2022daam} & Diffusion & Pixel-Level Attribution Maps & Attention Visualization \\
    \midrule
    \multicolumn{5}{c}{\textit{Embedding-Level Interpretability}} \\
    \midrule
    \multirow{4}{*}{Visual} & Verma et al.~\cite{verma2024cross} & MLLMs & Analyzing Visual Attributes & Attribute Analysis \\
    & Shi et al.~\cite{shi2024eagle} & MLLMs & Multiple Vision Encoders & Encoder Analysis \\
    & Park et al.~\cite{park2024explaining} & Diffusion & Visualizing Diffusion Process & Diffusion Process Visualization \\
    & Prasad et al.~\cite{prasad2024tree} & Diffusion & Tree of Diffusion & Diffusion Interpretation \\
    \midrule
    \multirow{2}{*}{Textual} & Wolfe~\cite{wolfe2022contrastive} & CLIP & Contrastive Learning Embeddings & Embedding Analysis \\
    & Moayeri et al.~\cite{moayeri2023text} & CLIP & Text-to-Concept Alignment & Concept Alignment \\
    \midrule
    \multirow{14}{*}{Cross-Modal} & Bhalla et al.~\cite{bhalla2024interpreting} & CLIP & Sparse Combinations of Concepts & Concept Combination \\
    & Frank et al.~\cite{frank2021vision} & CLIP & Exploring Structural Aspects & Structural Analysis \\
    & Ramesh~\cite{ramesh2022investigation} & CLIP & Comparing Explainability Methods & Explainability Comparison \\
    & Derby et al.~\cite{derby2018using} & MLLMs & Sparse, Interpretable Vectors & Vector Interpretation \\
    & Chen et al.~\cite{chen2023stair} & MLLMs & Grounded Representations & Representation Learning \\
    & Dominici et al.~\cite{dominici2023sharcs} & MLLMs & Shared Concept Space & Concept Space Analysis \\
    & Wang et al.~\cite{wangfreebind2024} & MLLMs & Integrating Expert Knowledge & Knowledge Integration \\
    & Parekh et al.~\cite{parekh2024concept} & MLLMs & Dictionary-Learning Framework & Concept Learning \\
    & Lindström~\cite{lindstrom2021probing} & MLLMs & Representation Probing & Probing Tasks \\
    & Salin et al.~\cite{salin2022vision} & MLLMs & Analyzing Fine-Tuning Effects & Fine-Tuning Analysis \\
    & Crabbé et al.~\cite{crabbe2023robust} & MLLMs & SVD and Concept Encoding & Concept Encoding \\
    & Nguyen~\cite{nguyen2019multi} & MLLMs & Multi-Task Learning Framework & Multi-Task Learning \\
    & Kwon et al.~\cite{kwon2022diffusion} & Diffusion & Asymmetric Reverse Process & Diffusion Process Analysis \\
    & Evirgen et al.~\cite{evirgen2024text} & Diffusion & Explanation Methods & Model Explanation \\
    \bottomrule
    \end{tabular}
    \label{table2-representtaion}
\end{table*}

\subsection{Token}
\label{section-representation-token}
In this section, we focus on multimodal explainability at the token level, categorizing tokens into visual and visual-textual tokens. 
Visual tokens are primarily studied to understand their impact on model output and to reduce token redundancy, thereby improving model explainability. 
Visual-textual tokens, on the other hand, are analyzed to explore how their distribution affects model output, with the goal of compressing both visual and textual tokens to further enhance explainability.
\textbf{Visual Tokens} (\cref{section-representation-visual-token}): discusses the importance of visual tokens in the model's decision-making process and how methods like interpretable basis decomposition~\cite{zhou2018interpretable} and analysis of attention mechanisms in visual transformers enhance their interpretability. Additionally, it explores methods to improve model efficiency and interpretability by reducing token redundancy. \textbf{Visual-Textual Tokens} (\cref{section-representation-visual-text-token}): explores the integration of visual and textual modalities, highlighting methods that enhance interpretability through visual-textual token alignment, mitigate undesirable behaviors like hallucination in multimodal models, and improve visual-language alignment.

\subsubsection{Visual Token}
\label{section-representation-visual-token}
The interpretability of visual tokens has become a crucial focus in the fields of computer vision and multimodal learning. 
Visual tokens refer to discrete units derived from an image, often representing specific regions or features, enabling models to handle high-dimensional visual information more efficiently. 
This review provides an in-depth overview of current research on visual tokens, emphasizing their role in enhancing model interpretability, performance, and computational efficiency.

\textbf{Backgrounds.}
Initial studies aimed to clarify how these image tokens contribute to the model's predictions, with Zhou et al.~\cite{zhou2018interpretable} introducing a basis decomposition framework that simplifies complex image representations into basic visual components, thereby elucidating the influence of individual tokens on model predictions.
More recent advancements have centered on interpretability techniques specifically tailored to the unique self-attention structures in ViTs, with an emphasis on the interactions between image patches. For instance, Ma et al.~\cite{ma2023visualizing} developed a method to visualize patch-wise interactions, highlighting how cross-patch correlations and attention distribution affect overall model performance. Extending these interpretability techniques, frameworks like ViT-NeT \cite{kim2022vit} and IA-ViT~\cite{qiang2023interpretability} introduced innovative visualization and training approaches. ViT-NeT utilizes a hierarchical tree structure with prototypes to organize and visualize attention layers, providing an insightful view of token interactions. Meanwhile, IA-ViT employs a joint training strategy for a feature extractor, predictor, and interpreter, ensuring that explanations remain consistent and faithful to the model’s internal processes.

\textbf{Token Efficiency and Redundancy.} A significant line of research has examined the efficiency and redundancy of visual tokens within deep layers of vision-language models. DynamicViT~\cite{rao2021dynamicvit} introduced a token sparsification method, dynamically pruning redundant tokens based on input, which enhances interpretability by focusing on the most informative tokens. In multimodal settings, Zhang et al.~\cite{zhang2024redundancy} further investigated token redundancy in MLLMs, revealing that token contributions converge in the shallow layers and become redundant in the deeper layers, which has implications for model efficiency and interpretability.
Building on token-level analysis, FastV~\cite{chen2024image} introduced a method to prune less relevant visual tokens in MLLMs, enhancing computational efficiency while preserving key interpretive insights. By concentrating on high-attention tokens, FastV highlights which visual elements the model prioritizes during decision-making, thereby clarifying the flow of visual information and enhancing the understanding of MLLMs.
Additionally, vision token-specific analyses reveal further insights into semantic alignment between image and text in vision-language models. Gandelsman et al.~\cite{gandelsman2023interpreting} analyzed CLIP’s image encoder by decomposing image representations into text-interpretable components, such as attention heads and image patches, uncovering specific roles for attention heads, including spatial localization and shape recognition. Similarly, Neo et al.~\cite{neo2024towards} explored the interpretability of visual tokens in the LLaVA model, showing how object-specific information is gradually refined across layers for improved predictions. 
Yao et al.~\cite{yao2024deco} focused on understanding the projector module in MLLMs, tracking how semantic information flows from language tokens back to visual patches.



\subsubsection{Visual-Textual Token}
\label{section-representation-visual-text-token}
The integration of visual and textual modalities in machine learning has significantly advanced the interpretability of complex tasks such as activity recognition, visual question answering, and content moderation.
Aligning visual elements with textual, multimodal interpretability methods provides a more comprehensive understanding of model behaviours and the decision-making process. 
This review discusses recent developments in this area, focusing on methodologies that enhance interpretability by leveraging visual-textual token alignment and related strategies.

Investigating the interpretability of MLLMs, VL-Match~\cite{bi2023vl} emphasized interpretability through a generator-discriminator structure. VL-Match operates by aligning tokens at a fine-grained level, utilizing negative sampling for instance-level alignment and ensuring token-level coherence between visual and textual representations.
Zhao et al.~\cite{zhao2024first} examined the logit distributions of initial tokens in MLLMs to reveal hidden knowledge within these models. Their findings indicate that analyzing these distributions can unveil inappropriate content generation, unanswerable questions, and other undesirable outputs, making the initial token analysis a useful tool for identifying and mitigating content generation issues.
LexVLA~\cite{li2024unified} presented a patch-level interpretability metric that specifically evaluates the alignment between image patch features and category-specific text tokens, providing a fine-grained approach to interpretability by examining the coherence between visual patches and corresponding textual categories. 
Additionally, DAAM~\cite{tang2022daam} proposed a novel method for interpreting large diffusion models like Stable Diffusion by generating pixel-level attribution maps using cross-attention scores, revealing how words in text prompts influence image generation, and analyzing syntactic and semantic phenomena affecting generation quality. 



\subsection{Embedding}
\label{section-representation-feature}
Despite the extensive research on individual tokens, there has also been a focus on more coarse-grained analyses of multimodal embeddings and their latent spaces in MLLMs. Similarly, we categorize these studies into visual, textual, and cross-modal embeddings. Additionally, studies have examined how MLLMs understand linguistic knowledge and their ability to recognize text within images. As shown in figure~\ref{fig:rep}, we illustrate three key methods: probing-based interpretation, attribution-based interpretation, and decomposition-based interpretation. More recently, some works have analyzed and improved the alignment and representational capacity of LLMs and MLLMs \cite{alain2016understanding, kim2018interpretability, qian2024towards, zhang2024better, zhang2024reef, lee2024mechanistic}. The overview of embedding-level methods is summarized in Table~\ref{table2-representtaion}.

\subsubsection{Visual Embedding}
\label{section-representation-visual-embedding}
The interpretability of MLLMs has become a focal point for understanding how these models process and integrate visual information, a crucial aspect as these models scale in complexity and application scope. 
Foundational methods, such as layer-wise evaluation using linear classifiers~\cite{alain2016understanding}, have been instrumental in clarifying the progressive encoding of visual features across model layers, offering insights into the structural evolution of learned representations. Techniques like Integrated Gradients \cite{sundararajan2017axiomatic} provide a structured method for attributing model predictions to specific visual features, enhancing transparency by delineating the contribution of each feature to the output. Other interpretability approaches concentrate on mapping neural representations to human-understandable concepts. For instance, Network Dissection \cite{bau2017network, zhou2018interpreting} assigned semantic labels to individual units in convolutional networks, while Testing with Concept Activation Vectors (TCAV) \cite{kim2018interpretability} quantifies the impact of high-level, user-defined concepts on predictions. In unsupervised settings, methods such as spatial masking \cite{zhu2021and} localized the influence of individual latent dimensions, capturing distinct, interpretable variations within visual data. 

Expanding on interpretability challenges unique to MLLMs, recent research highlights the intricacies of textual and visual representation interactions. 
For instance, \cite{verma2024cross} discovered that domain-specific visual attributes within MLLMs are often represented within the LLMs rather than the cross-modal projection, underscoring the importance of understanding how LLMs encode visual information to improve interpretability. Moreover, shi et al. \cite{shi2024eagle} examined the potential of incorporating multiple vision encoders within MLLMs, demonstrating that straightforward methods like concatenating visual tokens from diverse encoders can significantly boost both interpretability and model performance.
Another area of interpretability research focuses on generative models, particularly diffusion models, where understanding dynamic processes is critical. Park et al. \cite{park2024explaining} visualized the diffusion process within generative models, showing how these models progressively build and refine semantic information by directing attention to relevant visual concepts and regions over successive time steps. Similarly, the Tree of Diffusion Life (TDL) Prasad et al. \cite{prasad2024tree} visualized data evolution in diffusion models through an innovative embedding approach that preserves both semantic relationships and temporal dynamics, offering a more intuitive grasp of the generative process and rendering the complex evolution of model outputs interpretable over time.

\subsubsection{Textual Embedding}
\label{section-representation-textual-embedding}
In recent years, significant advances in understanding the internal structure of embeddings have greatly enhanced the interpretability of multimodal models. For instance, Hennigen et al.~\cite{hennigen2020intrinsic} propose a decomposable multivariate Gaussian probe for intrinsic analysis of text embeddings, uncovering that only a limited subset of neurons encodes core morphosyntactic features. This targeted approach improves interpretability by pinpointing how linguistic information is distributed across neural representations. Building on this, SVO-Probes~\cite{hendricks2021probing} identified specific challenges in verb comprehension for multimodal image-language transformers, particularly in comparison to nouns, thus highlighting critical areas for refinement in model interpretability. Meanwhile, Wolfe et al.~\cite{wolfe2022contrastive} demonstrate that CLIP embeddings achieve reduced anisotropy relative to those of GPT-2, enhancing semantic coherence and interoperability across word and sentence embeddings. In addition, Moayeri et al.~\cite{moayeri2023text} introduced an innovative ``text-to-concept'' alignment method that maps features from pre-trained models into CLIP's embedding space, enabling more direct interpretation of model features through aligned text embeddings.

\subsubsection{Cross-modal Embedding}
\label{section-representation-cross-modal-embedding}
To tackle the challenge of interpretability in MLLMs, researchers have developed various approaches that focus on creating cross-modal embeddings that align with human cognition. Joint Non-Negative Sparse Embedding (JNNSE)~\cite{derby2018using} introduced an early approach, generating sparse, interpretable vectors that capture multimodal semantic information by aligning with human behaviour and neuroimaging data. Building on this, STAIR~\cite{chen2023stair} grounded image and text representations in human-understandable tokens, providing not only improved model explainability but also enhanced retrieval performance, which underscores the dual benefit of interpretability-focused embeddings. Expanding the interpretability of multimodal representations, SHARCS (Shared Concept Space)~\cite{dominici2023sharcs} unified interpretable concepts from diverse modalities into a shared space, thus creating a versatile framework for situations involving missing modalities and enhancing the general applicability of multimodal learning.

Further efforts to refine cross-modal embedding interpretability include SpLiCE~\cite{bhalla2024interpreting}, which transforms CLIP embeddings into sparse, interpretable combinations of human-friendly concepts. By allowing concept-level analysis without explicit concept labels, SpLiCE preserves interpretability and downstream performance, enriching both qualitative and quantitative understanding of multimodal models. Additionally, Gandelsman et al.~\cite{gandelsman2023interpreting} and Frank et al.~\cite{frank2021vision} have explored structural aspects of models like CLIP, revealing insights into information asymmetry and the role of specific attention heads in processing text and visual inputs. Lastly, FreeBind~\cite{wangfreebind2024} introduced ``space bonds" to integrate expert knowledge across multimodal spaces while maintaining unified interpretive coherence, and Parekh et al.~\cite{parekh2024concept} proposed a dictionary-learning framework that further elucidates multimodal concept extraction grounded in visual and textual representations. Together, these advancements reflect a continuous push towards more interpretable and balanced multimodal models.

\textbf{Difussion Interpretability.} Recent advancements in interpretability techniques for MLLMs have enabled more intuitive and controllable generative processes. The Asymmetric Reverse Process (Asyrp)~\cite{kwon2022diffusion} introduced a semantic latent space (h-space) in pretrained diffusion models, facilitating interpretable and precise image editing across different timesteps. Complementing these advances, Evirgen et al.~\cite{evirgen2024text} proposed novel explanation methods specifically for text-to-image systems, allowing users to better understand and utilize these models.

\textbf{Probing explainability.} Understanding how cross-modal embeddings encode and transfer information is essential for achieving deeper interpretability. Lindström et al.~\cite{lindstrom2021probing} provided valuable insights by examining visual-semantic embeddings and revealing how these embeddings capture complementary information from text and image modalities, particularly in tasks involving synonyms and polysemy. Salin et al.~\cite{salin2022vision} further explored this area by employing probing tasks to analyze how fine-tuning affects the interpretability of Vision-Language model embeddings. Expanding on these perspectives, Ramesh et al.~\cite{ramesh2022investigation} compared explainability frameworks—Label Attribution and Optimal Transport—to examine attention interactions in multimodal transformers like CLIP and ViLBERT, addressing the need for unified explainability across different models. Crabbé et al.~\cite{crabbe2023robust} contributed to this field by using Singular Value Decomposition (SVD) and concept encoding to investigate privileged directions and polysemantic features within cross-modal embeddings, showing how these components encode complex information across multiple concepts. 

\textbf{Graph-Based Interpretability.} Beyond embeddings, hierarchical and graph-based methods have been explored to enhance interpretability in multimodal models. 
LaPool~\cite{noutahi2019towards} introduced an interpretable hierarchical graph pooling method that utilizes both node features and graph structure. This method significantly improves molecular representation in Graph Neural Networks (GNNs) and enhances interpretability in molecular tasks such as drug design, demonstrating the applicability of interpretable models in specialized domains.



\subsection{Neuron}
\label{section-model-neuron}

\begin{table*}[h]
\centering
\small
\caption{
\textbf{Overview of Neuron-Level Methods.} The papers are sorted by year, and annotated with their authors, research perspectives, models, and interpretation forms/functions of neurons.
}
\label{table:neuron_overview}
\resizebox{0.8\textwidth}{!}{%
\begin{tabular}{llllll}
\toprule
Perspective & Method & Venue & Models & Interpretation Form / Function \\
\midrule
\multirow{17}{*}{Individual Units} & Network Dissection \cite{bau2017network} & CVPR'17 & CNN & Concept \\
& GAN Dissection\cite{bau2018gan} & ICLR'18 & GAN & Concept \\
& Bau et al.~\cite{bau2020understanding} & PANS'20 & GAN, CNN & Concept \\
& MILAN\cite{hernandez2021natural} & ICLR'22 & GAN, ViT, CNN & Natural Language \\
& Dai et al.~\cite{dai2021knowledge} & ACL'21 & BERT & Factual Knowledge \\
& Goh et al.~\cite{goh2021multimodal} & Distill'21 & CLIP & Concept \\
& CLIP-Dissect\cite{oikarinen2022clip} & ICLR'22 & CNN & Concept \\
& ROME\cite{meng2022locating} & NeurIPS'22 & GPT & Factual Knowledge \\
& HINT\cite{wang2022hint} & CVPR'22 & CNN & Hierarchical Concept \\
& Rosetta Neurons\cite{dravid2023rosetta} & ICCV'23 & CNN, CLIP, DINO & Concept \\
& Bills et al.~\cite{bills2023language} & Openaipublic'23 & GPT & Natural Language \\
& Cones\cite{Liu2023ConesCN} & ICML'23 & Diffusion & Concept \\
& Chen et al.~\cite{chen2024journey} & AAAI'23 & m-BERT, m-GPT & Factual Knowledge \\
& Gao et al.~\cite{gao2024scaling} & arXiv'24 & GPT-4 & Concept \\
& Gandelsman et al.~\cite{gandelsman2024interpreting} & arXiv'24 & CLIP & Concept \\
& DEAN \cite{qian2024dean} & arXiv'24 & LLMs & Concept \\
& NeMo \cite{Hintersdorf2024FindingNL} & NeurIPS'24 & Diffusion & Individual Samples \\
\midrule
\multirow{11}{*}{Specialization Group} & Schubert et al.~\cite{schubert2021high} & Distill'20 & CNN & High-Low Frequency Detection \\
& Curve Detectors\cite{cammarata2020curve} & Distill'20 & CNN & Curve Detection \\
& Zoom In \cite{olah2020zoom} & Distill'20 & CNN & Meaningful Circuit \\
& Mueller et al.~\cite{Mueller2022CausalAO} & CoNLL'22 & m-BERT, XGLM & Multilingual \\
& Schwettmann et al.~\cite{schwettmann2023multimodal} & ICCV'23 & MLLMs & Multimodal Sensing \\
& Pan et al.~\cite{pan2023finding} & ACL'24 & MLLMs & Multimodal Sensing \\
& Tianyi Tang\cite{Tang2024LanguageSpecificNT} & ACL'24 & LLMs & Multilingual \\
& Kojima et al.~\cite{Kojima2024OnTM} & NAACL'24 & LLMs & Multilingual \\
& Gurnee et al.~\cite{gurnee2024universal} & TMLR'24 & GPT & Universal Pattern \\
& MMNeuron \cite{huo2024mmneuron} & EMNLP'24 & MLLMs & Vision Domain Sensing \\
& MINER\cite{Huang2024MINERMT} & arXiv'24 & MLLMs & Modality Sensing \\
\bottomrule
\end{tabular}
}%
\end{table*}

Neurons in pre-trained models have been a key focus for interpretability research, with studies in computer vision (CV) and natural language processing (NLP) analyzing their functions and semantic roles~\cite{cohen2011measuring,sajjad2022neuron}. In multimodal models, efforts extend to exploring neurons tied to specific concepts or domains. This section provides an overview of these studies, detailed shown in Table~\ref{table:neuron_overview}, covering both detailed and broader analyses in multimodal and traditional domains.
\subsubsection{Individual Units}
\label{section-Individual Units}
It has been widely explored how to associate individual neurons in deep neural networks with specific concepts or functions. \par
\textbf{Backgrounds.} Upon the proposal of the transformer, Dai et al.~\cite{dai2021knowledge} introduced the concept of ``knowledge neurons" in transformer models, referring to neurons that activate when expressing factual information. Their work demonstrated that deactivating these knowledge neurons significantly impairs the accuracy of the corresponding facts stored within the transformer model.
Meng et al.~\cite{meng2022locating} further explored knowledge neurons in GPT, discovering that factual associations within GPT can be directly modified by locating and editing specific neurons in the MLP layers. This work provided deeper insights into how knowledge is structured within language models.
Chen et al.~\cite{chen2024journey} expanded on this research by analyzing knowledge neurons in multilingual large language models. They proposed a gradient-based detection method, AMIG, to identify neurons storing specific knowledge. Chen et al. categorized these knowledge neurons into two types: language-independent neurons and degenerate neurons, which were determined by whether the stored knowledge was shared across languages or specific to one language input. Qian et al.~\cite{qian2024dean} identified that fairness and privacy-related neurons are coupled in LLMs. A simple and effective decoupled operation mitigated the fairness-privacy conflicts.

In the vision domain, Bau et al.~\cite{bau2017network} proposed the ``network dissection" task, designed to identify and label the concepts captured by neurons in CNNs. Manually assigning language-based labels to each neuron based on activation patterns provided a foundation for understanding the role of individual neurons, although the method was labor-intensive. Network dissection has also been applied to interpret neurons in GANs \cite{bau2018gan}, with further refinements in subsequent studies by Bau et al.~\cite{bau2020understanding}. Furthermore, MILAN \cite{hernandez2021natural} introduced a mutual-information-guided linguistic annotation method that automates neuron annotation by maximizing pointwise mutual information between neuron activations and image regions, offering fine-grained, natural language descriptions of neurons. Oikarinen et al.~\cite{oikarinen2022clip} moved forward proposing an alternative for generating concept descriptions for individual visual neurons by comparing their activation vectors with a text matrix in CLIP. Wang et al.~\cite{wang2022hint} extended the network dissection approach to construct bidirectional hierarchical connections between neurons and concepts, which they validated through weakly-supervised object localization. Dravid et al.~\cite{dravid2023rosetta} also explored ``rosetta" neurons, referring to neurons that universally exist across different visual networks and capture similar features. More recently, Bills et al.~\cite{bills2023language} and Gao et al.~\cite{gao2024scaling} utilized sparse autoencoders to interpret neurons in LLMs, providing new perspectives on neuron functionality.

\textbf{Individual Neurons.} In the multimodal domain, there has been notable work on concept neurons in multimodal networks. Goh et al.~\cite{goh2021multimodal} introduced the concept of multimodal neurons in CLIP, which respond to concepts present in both real and textual images. Schwettmann et al. Gandelsman et al.~\cite{gandelsman2024interpreting} decomposed CLIP representations to analyze the indirect influence of individual neurons. By projecting decomposed embeddings into vocabulary space, they reveal the secondary semantic effects of neurons in CLIP. As for multimodal generative AI, Liu et al.~\cite{Liu2023ConesCN} introduced ``Cones", a method to detect and edit concept neurons in diffusion models. By enabling selective activation or concatenation of concept neuron clusters, they managed to manipulate specific subjects in the generated images. Hintersdorf et al.~\cite{Hintersdorf2024FindingNL} developed NEMO, a method to pinpoint and manage memorization neurons in cross-attention layers of diffusion models. 

\subsubsection{Specialization Group}
\label{section-Specialization Groups}
While some researchers focus on analyzing the function of individual neurons, there is also a perspective that considers groups of neurons as collectively responsible for specific tasks. \par
\textbf{Backgrounds.} In visual networks, Cammarata et al.~\cite{cammarata2020curve} discovered neurons specialized in detecting curves within images. Schubert et al.~\cite{schubert2021high} extended these findings, identifying visual neurons responsible for detecting high- and low-frequency features. Olah et al.~\cite{olah2020zoom} further confirmed that features and circuits similar to those in \cite{cammarata2020curve, schubert2021high} are universal across visual networks, providing insights into how these networks interpret images. There is also relevant research in language models. For example, Gurnee et al.~\cite{gurnee2024universal} discussed universal neurons in GPT-2, noting that these neurons show consistent activation patterns across various model instances. These universal neurons are essential for adjusting prediction uncertainty and managing attention for specific tokens. Mueller et al.~\cite{Mueller2022CausalAO} used a causal-based method to examine neurons responsible for syntactic agreement across different languages, finding that these neurons show greater overlap in autoregressive models compared to masked language models. Tang et al.~\cite{Tang2024LanguageSpecificNT} introduced the concept of language-specific neurons (LSNs) and demonstrated that activating or deactivating these LSNs can influence the output language of multilingual large language models (XLLMs). Kojima et al.~\cite{Kojima2024OnTM} found that LSNs are primarily located in the top and bottom layers of XLLMs and exhibit minimal overlap across languages.

\textbf{Multimodal Neuron Groups.} Research on multimodal models, however, often emphasizes neurons that bridge text and image features. \cite{schwettmann2023multimodal} expanded this concept, detecting multimodal neurons even in the text-only language component of MLLMs. Similarly, Pan et al.~\cite{pan2023finding} identified multimodal neurons in pre-trained transformers, assessing their sensitivity, specificity, and causal effects. Huo et al.~\cite{huo2024mmneuron} proposed the Domain Activation Probability Entropy (DAPE) score to identify domain-specific neurons and evaluated their impact on VQA tasks. Recently, Huang et al.~\cite{Huang2024MINERMT} extend this concept to modality-specific neurons and propose an importance score-based method to detect neurons specific to different modalities.

\subsection{Layer}
\label{section-model-layer}
Classic deep neural networks, particularly transformer architectures, consist of stacked hidden layers connected by skip connections, such as decoder-based (e.g., GPT), encoder-based (e.g., BERT), or encoder-decoder models (e.g., T5 \cite{raffel2020exploring}). These layers can be further broken into components like MLP, multi-head attention (MHA), and layer normalization. This section \cref{section-model-neuron} reviews the literature on these layer-level elements from two perspectives: first, the function of specific layers (e.g., attention heads, MLP layers) and their contribution to model decisions; second, the overall decision-making process across layers, focusing on representation transformation.
The methods summarized are presented in Table~\ref{tab:my-table-layer-level}.

\subsubsection{Individual Components} Numerous studies have attempted to interpret the roles of different layers in deep neural networks across domains such as CV, NLP, and multimodal applications. \par \textbf{Backgrounds.} Mahendran et al.~\cite{mahendran2015understanding} explored the invertibility of hidden states in CNNs, showing that photographically accurate information about images is retained across several layers. Dosovitskiy et al.~\cite{dosovitskiy2015inverting} inverted CNNs hidden states using an up-convolutional neural network, finding that colors and rough contours of input images could be reconstructed from activations in higher network layers, even from predicted class probabilities.

With the rise of transformers, several studies have examined the functions of attention heads and MLP layers within these models. Cordonnier et al.~\cite{cordonnier2019relationship} argued that attention layers can effectively perform convolution and often learn to do so in practice. They also proved that a multi-head self-attention layer with a sufficient number of heads is at least as expressive as any convolutional layer. Sukhbaatar et al.~\cite{sukhbaatar2019augmenting} proposed augmenting the self-attention layer with a persistent memory vector, suggesting that these memory vectors could replace the transformer’s MLP layers. Geva et al.~\cite{geva2020transformer} demonstrated that feed-forward layers in transformer-based language models act as key-value memories, with each key linked to textual patterns in training examples, while each value influences output vocabulary distributions. Michel et al.~\cite{michel2019sixteen} conducted ablation studies on transformer attention heads, showing that many heads could be removed during testing without substantial performance loss. Voita et al.~\cite{voita2019analyzing} analyzed individual attention heads, finding that only a few critical heads have interpretable functions, such as attending to adjacent words and tracking specific syntactic relationships.

In NLP, the layers of pre-trained language models have been extensively analyzed. Clark et al.~\cite{clark2019does} examined BERT’s attention heads, observing patterns such as attention to delimiter tokens, specific positional offsets, and broad sentence-wide attention, often with similar behaviors among heads within the same layer. Kovaleva et al.~\cite{kovaleva2019revealing} explored the information encoded by BERT’s individual heads, revealing a limited set of attention patterns repeated across different heads. Htut et al.~\cite{htut2019attention} investigated BERT and RoBERTa’s ability to implicitly capture syntactic dependencies, finding some specialized attention heads for specific dependency types, although no generalist head was identified for holistic parsing. Ren et al.~\cite{ren2024identifying} analyzed the attention heads to explain LLMs' in-context learning.

\textbf{Multimodal Components.} Cao et al.~\cite{cao2020behind} analyzed multimodal pre-training through probing tasks across various model architectures, identifying a subset of attention heads optimized for cross-modal interactions and effectively encoding linguistic knowledge. Gandelsman et al.~\cite{gandelsman2023interpreting} decomposed CLIP’s image representation as a sum across individual image patches, layers, and attention heads, using CLIP’s text representation to interpret these components. They found that each attention head's role could be characterized through text representations spanning its output space, concluding that MLPs in CLIP have minimal direct impact. Quantmeyer et al.~\cite{quantmeyer2024and} applied interpretability methods from language models, such as causal tracing, to multimodal models, isolating parts of CLIP’s text encoder that handle negation and analyzing the roles of attention heads in this task. Si et al.~\cite{si2024freeu} examined U-Net’s contributions to the denoising process, revealing that its backbone primarily aids in denoising, while skip connections introduce high-frequency features into the decoder.

\subsubsection{Decision-Making Workflow}

\begin{table*}[t]
\centering
\caption{
\textbf{Overview of Layer-Level Methods.} The table summarizes key research contributions by perspectives (components or workflow), models, analysis topics, and application fields.
}
\label{tab:my-table-layer-level}
\begin{tabular}{llllll}
\toprule
Perspective & Method & Venue & Models & Topics & Field \\
\midrule
\multirow{15}{*}{Components} 
& Mahendran et al.~\cite{mahendran2015understanding} & CVPR'14 & CNN & Convolutional Layer & CV \\ 
& Dosovitskiy et al.~\cite{dosovitskiy2015inverting} & CVPR'15 & CNN & Convolutional Layer & CV \\ 
& Cordonnier et al.~\cite{cordonnier2019relationship} & ICLR'19 & CNN, Transformer & Convolutional Layer, Self-Attention & CV \\ 
& Sukhbaatar et al.~\cite{sukhbaatar2019augmenting} & arXiv'19 & Transformer & Self-Attention & NLP \\ 
& Geva et al.~\cite{geva2020transformer} & EMNLP'20 & Transformer & FFN & NLP \\ 
& Michel et al.~\cite{michel2019sixteen} & NeurIPS'19 & Transformer & Multi-Head Attention & NLP \\
& Voita et al.~\cite{voita2019analyzing} & ACL'19 & Transformer & Multi-Head Self-Attention & NLP \\
& Clark et al.~\cite{clark2019does} & BlackboxNLP'19 & BERT & Attention & NLP \\
& Kovaleva et al.~\cite{kovaleva2019revealing} & EMNLP'19 & BERT & Attention & NLP \\
& Htut et al.~\cite{htut2019attention} & arXiv'19 & BERT & Attention & NLP \\
& Ren et al.~\cite{ren2024identifying} & ACL'24 & LLMs & Attention & NLP \\
& Cao et al.~\cite{cao2020behind} & ECCV'20 & ViLBERT, LXMERT & Attention & MM \\
& Gandelsman et al.~\cite{gandelsman2023interpreting} & ICLR'23 & CLIP & FFN, Attention & MM \\
& Quantmeyer et al.~\cite{quantmeyer2024and} & ALVR'24 & CLIP & Attention & MM \\
& FreeU~\cite{si2024freeu} & CVPR'23 & Diffusion & U-Net & MM \\ \midrule
\multirow{9}{*}{Workflow}
& VCC~\cite{kowal2024visual} & CVPR'24 & CNN & Inner Connection & CV \\ 
& Van Aken et al.~\cite{van2019does} & CIKM'19 & BERT & Representation Transformation & NLP \\
& Tenney et al.~\cite{tenney2019bert} & ACL'19 & BERT & Pipeline & NLP \\ 
& Wolfe et al.~\cite{wolfe2022contrastive} & ACL'22 & GPT, CLIP & Representation Transformation & MM \\ 
& Xu et al.~\cite{xu2023bridging} & ICCV'23 & Dual-stream VLM & Modality Alignment & MM \\ 
& Palit et al.~\cite{palit2023towards} & ICCV'23 & BLIP & Causal Relevance & MM \\ 
& MMNeuron~\cite{huo2024mmneuron} & EMNLP'24 & MLLMs & Modality Alignment & MM \\ 
& Zhang et al.~\cite{zhang2024redundancy} & arXiv'24 & MLLMs & Information Flow & MM \\ 
& Prasad et al.~\cite{prasad2023unraveling} & arXiv'23 & Diffusion & U-Net & MM \\ \bottomrule
\end{tabular}%
\end{table*}

Beyond analyzing the functions of different layers in neural networks, it is equally crucial to interpret the decision-making process of models across layers, from shallow to deep. This involves uncovering how pre-trained models perceive inputs and make decisions, providing deeper insights into their understanding and reasoning mechanisms.\par
\textbf{Backgrounds.} Understanding how learned representations transform within deep neural networks is another vital challenge in interpretability. Kowal et al.~\cite{kowal2024visual} proposed the ``Visual Concept Connectome" (VCC) method, which identifies human-interpretable concepts and inter-layer connections in visual networks, quantifying the contributions of these concepts without needing labeled datasets. In NLP, Van Aken et al.~\cite{van2019does} applied general and QA-specific probing tasks to reveal information stored in each representation layer, showing that transformations in BERT follow phases related to traditional pipeline tasks. Tenney et al.~\cite{tenney2019bert} discovered that BERT represents the steps of the traditional NLP pipeline in an interpretable, localized manner. \par
\textbf{Workflow of Multimodal Models.} Recent work has explored the decision-making workflow of multimodal models. Xu et al.~\cite{xu2023bridging} combined convolutional and attention mechanisms with an adapter module, finding that placing this module in shallower layers enhanced the vision-language model performance more effectively than placing it in top layers. Wolfe et al.~\cite{wolfe2022contrastive} observed that CLIP’s sentence embeddings became progressively less self-similar across layers, indicating that contrastive pre-training objectives drive the formation of fine-grained semantic sentence representations. Palit et al.~\cite{palit2023towards} adopt a causal tracing tool for mechanistic interpretability in vision-language models, elucidating the causal role of representations in later layers during image-conditioned text generation, offering insights into the underlying mechanisms beyond simple input-output correlations. Huo et al.~\cite{huo2024mmneuron} proposed a three-stage hypothesis for how multimodal language models process visual embeddings, which they verified using the logit lens method. Zhang et al.~\cite{zhang2024redundancy} used LLaVA-CAM and attention scores to visualize information flow in reasoning processes across layers in MLLMs, finding that information converges in shallow layers and diverges in deep layers. Tao et al.~\cite{tao2024probing} noted that intermediate layers of models encode more global semantic information, making them better suited for visual-language entailment tasks than top layers. Prasad et al.~\cite{prasad2023unraveling} evaluated time-step and U-Net component impacts on Stable Diffusion’s final output, showing that lower layers primarily contribute to semantic alterations, while higher layers focus on denoising, especially after the initial generation phase.
Nguyen et al.~\cite{nguyen2019multi} propose a multi-task learning framework that leverages Dense Co-attention layers to jointly learn hierarchical vision-language representations. By using task-specific decoders and attention map visualizations, their approach enhances interpretability through explicit modeling of cross-modal interactions.

\subsection{Architecture}
\label{section-model-arch}
In \cref{section-model-neuron} and \cref{section-model-layer}, we examined interpretability at the fine-grained neuron and layer levels. However, some studies explore the interpretability of MLLMs at a more coarse-grained architecture level. We will provide a detailed definition of the architecture level and subsequently introduce and categorize these related works. Unlike previous methods that focus on the specific components of MLLMs, this subsection will treat the MLLMs model as a whole. We also aim to explore whether we can explain the decision-making process of MLLMs in this manner. We categorize these works into two groups:
\begin{itemize}
    \item \textbf{Architecture Analysis:} (\cref{section-model-arch-analysis}) This approach is independent of any specific model structure or internal mechanism, such as attention operations in transformers or convolutional units in CNNs, enabling us to apply it for explanations of any MLLMs.
    \begin{itemize}
        \item \textbf{Feature Attribution:} We introduce classic explanation methods that attribute importance scores to features, forming the basis for subsequent approaches.
        \item \textbf{Uni-modal Explanation:} Here, we include methods that provide single-modality explanations (mostly for the image modality), offering a comprehensive global perspective.
        \item \textbf{Multi-modal Explanation:} There are also methods that provide multi-modal explanations (e.g., combining image and text modalities), offering users a more comprehensive perspective.
        \item \textbf{Interactive Explanation:} Methods that provide explanations based on human commands or preferences are grouped here under the category of \textit{interactive explanation}.
        \item \textbf{Others:} Architecture-level model analysis methods, which offer insights into model characteristics through model comparisons, are also included here for reference.
    \end{itemize}
    \item \textbf{Architecture Design:} (\cref{section-model-arch-design}) These methods enhance model explainability by modifying the architecture with highly interpretable modules. Unlike architecture analysis, they do not generate explicit explanation outputs but focus on specific model types, leveraging unique structures or parameters to explore internal mechanisms and produce detailed insights.
    \begin{itemize}
        \item \textbf{Surrogate Model:} A simpler model, such as a linear model or decision tree, is used to approximate the performance of a complex model.
        \item \textbf{Concept-based:} This approach enables the model to learn human-understandable concepts, which are then used to generate predictions.
        \item \textbf{Causal-based:} These methods incorporate concepts from causal learning into architecture design, such as causal reasoning or causal frameworks.
        \item \textbf{Others:} We include here methods related to other modules in the architecture that cannot be categorized into the classes mentioned above.
    \end{itemize}
\end{itemize}

We will then provide a detailed explanation of the methods within these categories.

\begin{table*}[]
\centering
\caption{
\textbf{Overview of Architecture Analysis Methods.} This table categorizes analysis methods into feature attribution, uni-modal, multi-modal, and interactive explanations, with details in \cref{section-model-arch-analysis}. It includes key highlights, architectures, tasks, and a comparison of explanation types and control signals.
}
\label{tab:my-table-Overview of Architecture Analysis Methods}
\resizebox{0.95\textwidth}{!}{%
\begin{tabular}{@{}llllccll@{}}
\toprule
\textbf{} & Method & Venue & Key Points & Explanation Type & Control Signal & Architecture & Task \\ \midrule
\multirow{3}{*}{\begin{tabular}[c]{@{}l@{}}Feature \\ Attribution\end{tabular}} 
    & LIME\cite{ribeiro2016should} & KDD'16 & \begin{tabular}[c]{@{}l@{}}Local Space\\ Decision Analysis\end{tabular} & -- & -- & Any Model & \begin{tabular}[c]{@{}l@{}}Classification\\ Regression\end{tabular} \\ \cmidrule(l){2-8} 
    & DeepLIFT\cite{shrikumar2017learning} & ICML'17 & \begin{tabular}[c]{@{}l@{}}Reference Activation\\ Importance Backpropagation\end{tabular} & -- & -- & Vision Model & Classification \\ \cmidrule(l){2-8} 
    & SHAP\cite{lundberg2017unified} & NIPS'17 & \begin{tabular}[c]{@{}l@{}}Unified Framework\\ Additive Feature Attribution\end{tabular} & -- & -- & Vision Model & Classification \\ \midrule
\multirow{5}{*}{\begin{tabular}[c]{@{}l@{}}Uni-modal \\ Explanation\\ (CAM family)\end{tabular}} 
    & CAM\cite{zhou2016learning} & CVPR'16 & \begin{tabular}[c]{@{}l@{}}Class Activation Mapping\\ Global Average Pooling\end{tabular} & Saliency Map & -- & Vision Model & \begin{tabular}[c]{@{}l@{}}Classification\\ Localization\end{tabular} \\ \cmidrule(l){2-8} 
    & Grad-CAM\cite{selvaraju2017grad} & ICCV'17 & \begin{tabular}[c]{@{}l@{}}Gradient-based\\ Without Altering of Network\end{tabular} & Saliency Map & -- & Vision Model & \begin{tabular}[c]{@{}l@{}}Classification\\ Image Captioning\\ VQA\end{tabular} \\ \cmidrule(l){2-8} 
    & U-CAM\cite{patro2019u} & ICCV'19 & Estimate Uncertainty & Saliency Map & -- & Vision Model & VQA \\ \cmidrule(l){2-8} 
    & Score-CAM\cite{wang2020score} & CVPR-W'20 & Without Gradient & Saliency Map & -- & Vision model & \begin{tabular}[c]{@{}l@{}}Image Recognition\\ Localization\end{tabular} \\ \cmidrule(l){2-8} 
    & gScore-CAM\cite{chen2022gscorecam} & ACCV'22 & Gradient-based & Saliency Map & -- & CLIP & Object Detection \\ \midrule
\multirow{4}{*}{\begin{tabular}[c]{@{}l@{}}Uni-modal \\ Explanation\\ (Others)\end{tabular}} 
    & Simonyan et al.~\cite{simonyan2013deep} & ICLR-W'14 & \begin{tabular}[c]{@{}l@{}}Activation Maximization\\ Attribution\end{tabular} & \begin{tabular}[c]{@{}l@{}}Saliency Map\\ Synthetic Image\end{tabular} & -- & CNN & Classification \\ \cmidrule(l){2-8} 
    & Binder et al.~\cite{binder2016layer} & ICANN'16 & \begin{tabular}[c]{@{}l@{}}LRP\\ Taylor Expansion\end{tabular} & Saliency Map & -- & Vision Model & Classification \\ \cmidrule(l){2-8} 
    & Montavon et al.~\cite{montavon2017explaining} & PR'17 & Deep Taylor Decomposition & Saliency Map & -- & Vision Model & Classification \\ \cmidrule(l){2-8} 
    & I-GOS\cite{qi2019visualizing} & AAAI'20 & \begin{tabular}[c]{@{}l@{}}Integrated Gradients\\ Mask Optimization\end{tabular} & Saliency Map & -- & DNN & Classification \\ \midrule
\multirow{5}{*}{\begin{tabular}[c]{@{}l@{}}Multi-modal \\ Explanation\end{tabular}} 
    & Wu et al.~\cite{wu2018faithful} & ACL-W'19 & Faithful Explanation & \begin{tabular}[c]{@{}l@{}}Segmentation Mask\\ Text Explanation\end{tabular} & -- & VLM & VQA \\ \cmidrule(l){2-8} 
    & CCM\cite{goyal2016towards} & WACV'20 & Robust Explanation & \begin{tabular}[c]{@{}l@{}}Saliency Map\\ Text Explanation\end{tabular} & -- & VLM & VQA \\ \cmidrule(l){2-8} 
    & Chefer et al.~\cite{chefer2021generic} & ICCV'21 & \begin{tabular}[c]{@{}l@{}}Any Transformer Architecture\\ Relevance Score\end{tabular} & \begin{tabular}[c]{@{}l@{}}Saliency Map\\ Text Relevancy\end{tabular} & -- & Transformer & Classification \\ \cmidrule(l){2-8} 
    & DIME\cite{lyu2022dime} & AIES'22 & \begin{tabular}[c]{@{}l@{}}Local Explanation\\ Disentangling\end{tabular} & \begin{tabular}[c]{@{}l@{}}Image Localization\\ Text Distribution\end{tabular} & -- & \begin{tabular}[c]{@{}l@{}}MLP\\ MDETR\\ LXMERT\end{tabular} & \begin{tabular}[c]{@{}l@{}}Classification\\ Visual Reasoning\end{tabular} \\ \cmidrule(l){2-8} 
    & VALE\cite{natarajan2024vale} & arXiv'24 & SHAP + SAM + VLM & \begin{tabular}[c]{@{}l@{}}Segmentation Mask\\ Text Explanation\end{tabular} & -- & Vision Model & Classification \\ \midrule
\multirow{4}{*}{\begin{tabular}[c]{@{}l@{}}Interactive \\ Explanation\end{tabular}} 
    & Olah et al.~\cite{olah2018building} & Distill'18 & Multiple Interaction Modes & \begin{tabular}[c]{@{}l@{}}Saliency Map\\ Synthetic Image\\ Text Distribution\end{tabular} & Select Pixel & GoogLeNet & Classification \\ \cmidrule(l){2-8} 
    & Diffusion Explainer\cite{lee2023diffusion} & arXiv'23 & Interactive & Generated image & Prompt & Diffusion Model & Text to Image \\ \cmidrule(l){2-8} 
    & MAIA\cite{shaham2024multimodal} & ICML'24 & \begin{tabular}[c]{@{}l@{}}Use Tools\\ Neuron-level Explanation\end{tabular} & \begin{tabular}[c]{@{}l@{}}Image Localization\\ Text Explanation\end{tabular} & Prompt & VLM & Vision Tasks \\ \cmidrule(l){2-8} 
    & LVLM-Interpret\cite{stan2024lvlm} & CVPR-W'24 & \begin{tabular}[c]{@{}l@{}}Interactive\\ Attention Mechanisms\end{tabular} & \begin{tabular}[c]{@{}l@{}}Saliency Map\\ Text Distribution\end{tabular} & Select Token & VLM & VQA \\ \bottomrule
\end{tabular}%
}
\end{table*}

\begin{figure*}[h]
\centering
\includegraphics[width=0.8\linewidth]{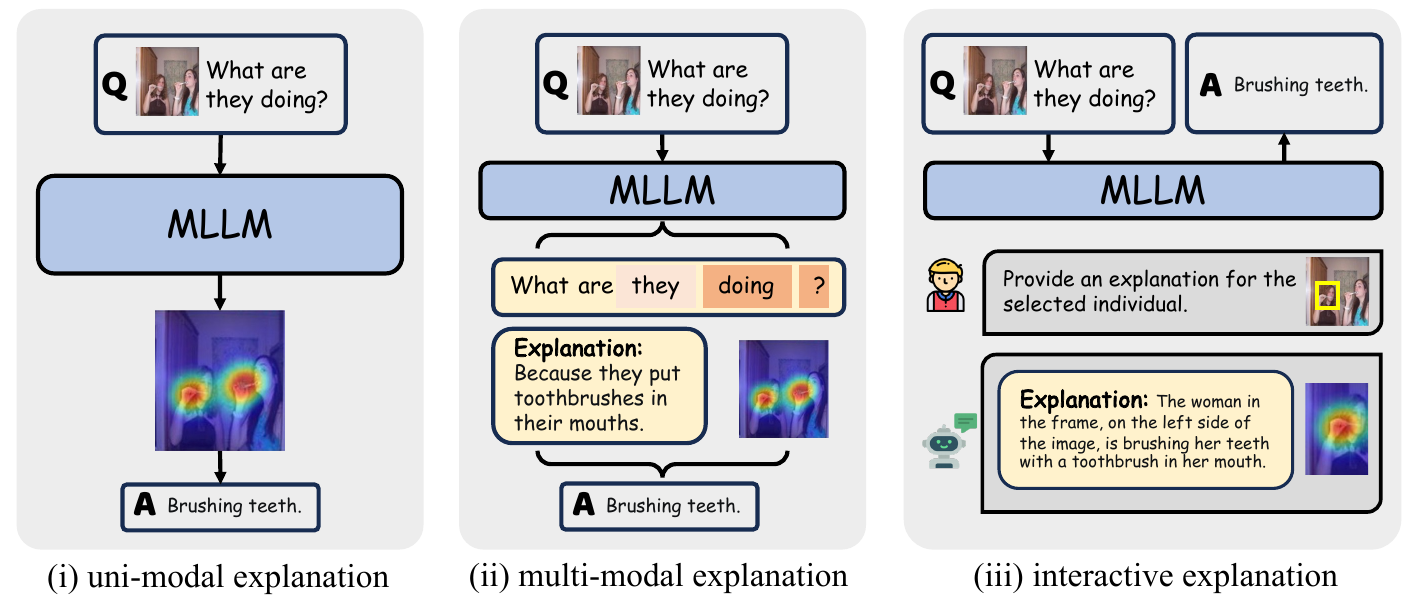}
\caption{
\textbf{Architecture Analysis.}\,\, We classify architecture analysis methods into three types: \textit{uni-modal}, \textit{multi-modal}, and \textit{interactive} explanations, based on explanation modalities and control signal acceptance.
}
\label{figure5-architecture}
\end{figure*}

\subsubsection{Architecture Analysis}
\label{section-model-arch-analysis}

Unlike the analysis of neurons, layers, or modules mentioned earlier, this section introduces works that utilize the entire model architecture to provide explanations. As is shown in Figure~\ref{figure5-architecture}, we categorize these methods based on an intuitive perspective—the type of explanation output: uni-modal explanations, multi-modal explanations, and interactive explanations.
The methods summarized are presented in Table~\ref{tab:my-table-Overview of Architecture Analysis Methods}.

\textbf{Feature Attribution.}\,\, These methods, primarily proposed in earlier years, were used as explanation techniques for CNNs or other models. Although they are not directly related to MLLMs, we include them here as background to provide readers with a more comprehensive understanding of the timeline in XAI development. We also hope these methods can inspire new explanation techniques for MLLMs.

\cite{ribeiro2016should,shrikumar2017learning,lundberg2017unified} explained the decision-making process of models by assigning contribution values to sparse features. LIME \cite{ribeiro2016should} trained simple linear models in local subspaces of the input space, approximating the behavior of complex models in those regions. Experiments show that LIME is effective for both expert and non-expert users, enhancing interpretability across various tasks, such as model comparison, trust assessment, improving unreliable models, and gaining insights into predictions. DeepLIFT \cite{shrikumar2017learning} propagated the contributions of each neuron in the network back to the input features to decompose the predicted output. It compares each neuron's activation value with its ``reference activation value" to allocate contribution scores based on the differences. SHAP \cite{lundberg2017unified} unified six existing methods for explaining model predictions (including LIME and DeepLIFT) by introducing the concept of ``additive feature attribution," assigning contribution values to each feature to elucidate predictions for individual samples.

\textbf{Uni-modal Explanation.}\,\, Building on Class Activation Mapping (CAM) introduced in \cite{zhou2016learning}, several follow-up methods \cite{selvaraju2017grad,patro2019u,wang2020score,chen2022gscorecam}, collectively known as the CAM Family, identify critical regions in input images and display them as class activation maps (heatmaps). CAM \cite{zhou2016learning} originally demonstrated object localization in CNNs without bounding box annotations by highlighting important areas. Grad-CAM \cite{selvaraju2017grad}, a CAM variant, extends this to multiple CNNs architectures without modification. U-CAM \cite{patro2019u} targets VQA tasks, generating visual attention maps through gradient-based estimates. Score-CAM \cite{wang2020score} improved on CAM by using forward scores to compute activation weights, making it gradient-independent and effective in recognition and localization. Lastly, gScore-CAM \cite{chen2022gscorecam} enhances CLIP interpretability, using gradient sampling groups to produce reliable attention maps while reducing complexity and avoiding distractions from text in images.

\cite{qi2019visualizing,simonyan2013deep} identify the most important image regions for the decision-making process by optimizing an objective function, typically aiming to minimize classification accuracy, and then use heatmaps or saliency maps to highlight significant areas of the image. I-GOS \cite{qi2019visualizing} determines the most important regions in an image by minimizing classification accuracy and employing integrated gradients instead of traditional gradients to calculate the descent direction. The paper \cite{simonyan2013deep} mentions two visualization techniques for CNNs: (1) using activation maximization to illustrate the category concepts captured by the CNN, and (2) calculating saliency maps for specific images and categories through backpropagated gradients. \cite{bach2015pixel,binder2016layer,montavon2017explaining} are based on relevance propagation methods that define the correlation between features and classification outcomes and reverse this correlation back to the input image through various propagation strategies. \cite{bach2015pixel,binder2016layer} discusses two strategies: (1) Taylor decomposition, which linearly approximates the classification function by performing a Taylor expansion around a neutral data point (one that does not belong to any category) to identify each pixel's contribution; (2) Layer-wise Relevance Propagation (LRP), which propagates the classification relevance from each layer to the previous one. \cite{montavon2017explaining} extends Deep Taylor Decomposition (DTD) to multilayer neural networks, generating heatmaps to assess the importance of individual pixels in classification.

\textbf{Multi-modal Explanation.}\,\, \cite{chefer2021generic,lyu2022dime,goyal2016towards,wu2018faithful,natarajan2024vale} offer multimodal interpretability, such as image and text explanations, providing more detailed and systematic insights compared to single-modal methods. \cite{chefer2021generic} introduces a general transformer explanation framework capable of interpreting (i) self-attention architectures, (ii) hybrid self-attention and cross-attention models, and (iii) encoder-decoder attention designs. DIME \cite{lyu2022dime} enhances fine-grained interpretability by decoupling information flows of different modalities, clarifying how each modality contributes to the model's decisions. CCM \cite{goyal2016towards} improves answer explanations for VQA models with a collaborative correlation module that strengthens the link between answers and explanations and enhances the quality of visual and textual outputs. \cite{wu2018faithful} integrates textual and visual explanations in a VQA system, presenting answers in a human-like style for better clarity and comprehension. VALE \cite{natarajan2024vale} combines SHAP for identifying influential image regions with SAM and pre-trained vision-language models (VLMs) to generate visual (e.g., heatmaps) and natural language explanations, offering a comprehensive view of the model’s reasoning.

\textbf{Interactive Explanation.}\,\, The field of explainable AI has recently grappled with the question of whether the decision-making process of deep neural networks (DNNs) can be interpreted in terms of a set of sparse symbolic concepts. A body of work has explored different types of interactions between a DNN's input variables as a means of interpreting its inner workings. Sundararajan et al.~\cite{sundararajan2020shapley}, Janizek et al.~\cite{janizek2021explaining}, and Tsai et al.~\cite{tsai2023faith} have each proposed distinct approaches to modeling these input-level interactions. Building on this, Ren et al.~\cite{ren2023defining} utilized the Harsanyi dividend to represent the AND-type interactions encoded by DNNs. Interestingly, their experimental findings suggest that DNNs tend to rely on a sparse set of such interactions between input variables. Further advancing this line of inquiry, Li et al.~\cite{li2023does} revealed that low-order interactions exhibit higher transferability across diverse input samples in discriminative neural networks. Complementing this, Ren et al.~\cite{renwe} formally derived the common conditions under which the sparsity of interactions can be guaranteed. Additionally, Ren et al.~\cite{ren2021can} introduced a method to learn optimal masked states of input variables based on their interactions, mitigating the bias introduced by sub-optimal masking in Shapley value-based interpretations. Taking a broader view, Chen et al.~\cite{chen2024defining} extracted common interaction patterns shared across different neural network architectures, suggesting that such generalizable interactions may underpin the networks' inference mechanisms. Moreover, Cheng et al.~\cite{cheng2024layerwise} proposed an approach to extract interactions from a DNN's intermediate layers, shedding light on how these inference patterns are gradually learned and forgotten during the forward propagation process. GANSpace~\cite{harkonen2020ganspace} employed principal component analysis (PCA) to uncover the key directions within the latent space. By selectively perturbing the layers along these primary axes, they achieved a high degree of explainability and control over the generated images.

The Harsanyi interaction theory \cite{ren2023we} offers a compelling perspective on the representational capabilities of neural networks, providing insights into their behavior and learning processes. This can be detailed as follows. Wang et al.~\cite{wang2020unified} established and mathematically verified an inverse relationship between the adversarial transferability of deep neural networks (DNNs) and the interactions present within adversarial perturbations. Ren et al.~\cite{ren2021towards} highlighted that adversarial attacks predominantly target higher-order interactions rather than lower-order ones. Similarly, Zhou et al.~\cite{zhou2024explaining} demonstrated that lower-order interactions exhibit superior generalization properties compared to their higher-order counterparts. Liu et al.~\cite{liu2024towards} provided an explanation for the observation that DNNs are more adept at learning lower-order interactions than higher-order ones. Deng et al.~\cite{deng2021discovering} identified a surprising bottleneck: neural networks often fail to encode middle-order bivariate interactions effectively. Ren et al.~\cite{ren2023bayesian} showed that Bayesian neural networks (BNNs) are less likely than standard neural networks to capture complex Harsanyi interactions. Additionally, Ren et al.~\cite{ren2024towards} and Zhang et al.~\cite{zhang2024two} uncovered a two-phase learning dynamic in neural networks' interaction acquisition, a phenomenon consistently observed across various architectures and tasks. Finally, Deng et al.~\cite{deng2024unifying} unified a range of classical attribution methods by proving that their underlying mechanisms could be reformulated as distinct ways of redistributing interaction effects among input variables.

Some explainability methods support interactive explanations, offering flexible, fine-grained analyses based on user preferences, such as focusing on specific image areas or neurons in MLLMs. These systems integrate various modalities, like text, images, and charts, to create comprehensive explanation frameworks. Their flexibility and depth provide users with broader insights, aiding informed decision-making in complex scenarios. \cite{olah2018building,shaham2024multimodal,stan2024lvlm} proposed the use of comprehensive explanation systems or agents to provide multimodal insights into model behavior. \cite{olah2018building,shaham2024multimodal} combined multiple explainability tools to enhance the understanding of complex neural networks. Specifically, \cite{olah2018building} unified various explainability techniques into a coherent syntax and created richer, more effective user interfaces. These interfaces, especially in visual tasks, help users better grasp the internal workings of neural networks. \cite{shaham2024multimodal} introduced MAIA, which integrates and automates a series of explainability tools. MAIA addresses two key challenges: (1) reducing sensitivity to spurious features and (2) automatically detecting potentially misclassified inputs, offering deeper insights into complex neural models. LVLM-Interpret \cite{stan2024lvlm} focused on identifying the critical image patches influencing model outputs. It proposes a novel interactive application that enhances the explainability of these patches, helping users understand the internal mechanics of LVLMs. Together, these works emphasize the importance of combining multiple explainability tools and user interaction to achieve more comprehensive and accessible model explanations. Diffusion Explainer \cite{lee2023diffusion} is an interactive visualization tool for diffusion, designed to explain how stable diffusion transforms text prompts into images. By comparing the image generation results of different text prompts, users can identify how changes in keywords impact the generated images.

\textbf{Others.} Additionally, some studies analyze model properties from an architectural perspective. Tran et al.\cite{tran2018importance} compared recurrent (RNN) and non-recurrent (Transformer) structures in modeling hierarchical information, showing that recurrent structures are advantageous for capturing hierarchy and offering insights for interpretability research. Yang et al.\cite{yang2024law} and Ramesh et al.\cite{ramesh2022investigation} focus on analyzing MLLMs: Yang et al.\cite{yang2024law} propose the ``Law of Vision Representation," revealing a strong link between cross-modal alignment, vision consistency, and model performance, while Ramesh et al.~\cite{ramesh2022investigation} examined various interpretability methods (e.g., attention weights, gradient-based approaches), evaluating their strengths and limitations across multimodal tasks and providing recommendations for improvement.

\begin{figure*}[h]
\centering
\includegraphics[width=1\linewidth]{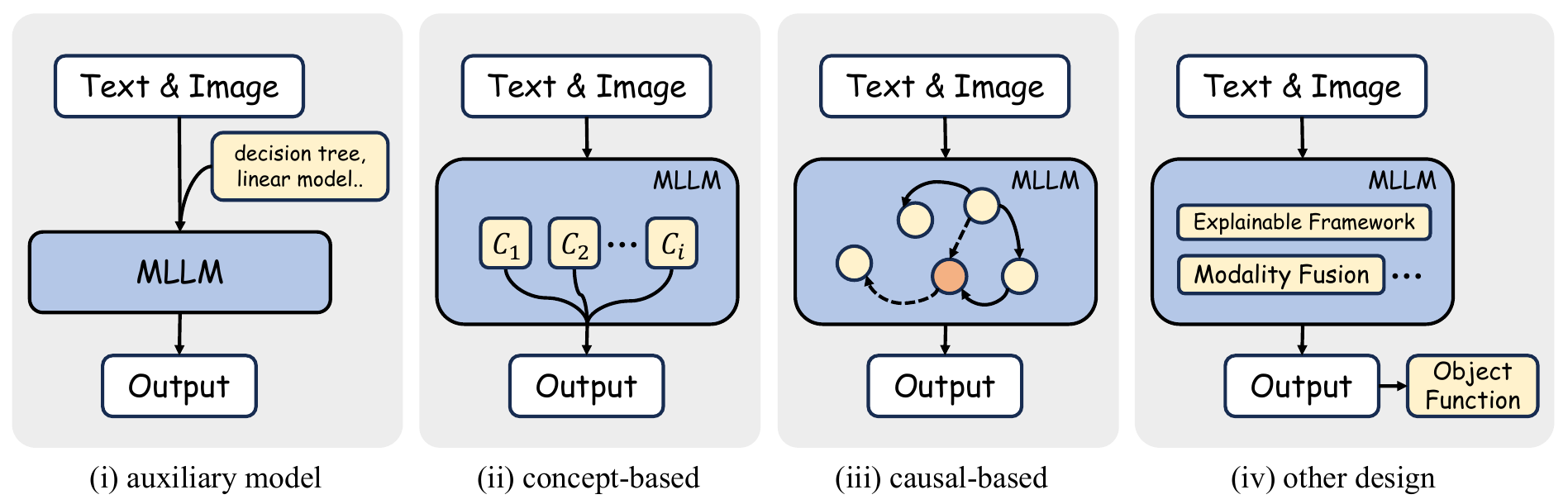}
\caption{
\textbf{Architecture Design.}\,\, This category focuses on modifying modules to improve explainability without generating explicit explanations. Methods include \textit{auxiliary models}, \textit{concept-based}, \textit{causal-based}, and \textit{other} approaches. 
}
\label{figure-6-adesign}
\end{figure*}

\begin{table*}[]
\centering
\caption{
\textbf{Overview of Architecture Design Methods.} This table categorizes design methods into surrogate, concept-based, causal-based, and others (details in \cref{section-model-arch-design}). It includes publication venues, highlights, architectures, tasks, and \checkmark indicators for causal or concept learning.
}
\label{tab:my-table-2-Architecture Design Methods}
\resizebox{0.95\textwidth}{!}{%
\begin{tabular}{@{}lllcccll@{}}
\toprule
Method & Venue & Key Points & Causal & Concept & Auxiliary Model & Architecture & Task \\ \midrule
CAN\cite{tang2021interpretable} & ICANN'21 & \begin{tabular}[c]{@{}l@{}}Modality Fusion Module\\ Inference Module\end{tabular} & -- & -- & -- & Vision Model & VCR \\ \midrule
IA-ViT\cite{qiang2023interpretability} & arXiv'23 & \begin{tabular}[c]{@{}l@{}}Attention\\ Training Objective\end{tabular} & -- & -- & -- & ViT-based & Classification \\ \midrule
MultiModN\cite{swamy2024multimodn} & NeurIPS'23 & Module Separation & -- & -- & -- & MM & \begin{tabular}[c]{@{}l@{}}Classification\\ Regression\end{tabular} \\ \midrule
VL-MoE\cite{shen2023scaling} & arXiv'23 & MoE & -- & -- & -- & VLM & \begin{tabular}[c]{@{}l@{}}Classification\\ Natural Language Inference\\ VQA\\ Image-Text Retrieval\end{tabular} \\ \midrule
Liu et al.~\cite{liu2018improving} & ICDM-W'18 & Knowledge Distillation & -- & \checkmark & Decision Tree & Vision Model & Classification \\ \midrule
Wong et al.~\cite{wong2021leveraging} & ICML'21 & \begin{tabular}[c]{@{}l@{}}Elastic Net\\ Sparse Decision Layer\end{tabular} & -- & \checkmark & Linear Layer & Vision Model & Classification \\ \midrule
NBDT\cite{wan2020nbdt} & ICLR'21 & \begin{tabular}[c]{@{}l@{}}Sequential,\\ Discrete Decisions\end{tabular} & -- & \checkmark & Decision Tree & Vision Model & Classification \\ \midrule
CBM\cite{koh2020concept} & ICML'20 & Bottleneck & -- & \checkmark & -- & Vision Model & Classification \\ \midrule
PCBM\cite{yuksekgonul2022post} & ICLR'23 & \begin{tabular}[c]{@{}l@{}}Post-hoc\\ Concept Transfer\end{tabular} & -- & \checkmark & SVM & Vision Model & Classification \\ \midrule
LaBo\cite{yang2023language} & CVPR'23 & \begin{tabular}[c]{@{}l@{}}LLMs Generated Concepts\\ CBM-based\end{tabular} & -- & \checkmark & LLMs & Vision Model & Classification \\ \midrule
TRACE\cite{guo2024trace} & arXiv'24 & Causal Event Modeling & \checkmark & -- & -- & Video LLMs & VTG \\ \midrule
LLCP\cite{chenllcp} & ICLR'24 & Latent Causal Process & \checkmark & -- & -- & MM & VideoQA \\ \midrule
MGCE\cite{li2024multimodal} & TKDE'24 & Causal Graph & \checkmark & -- & -- & MM & Recommendation \\ \midrule
Li et al.~\cite{li2024steering} & arXiv'24 & Debiasing Framework & \checkmark & -- & -- & LLMs & Debiasing \\ \midrule
Liu et al.~\cite{liu2024revealing} & arXiv'24 & Disentangled Representations & \checkmark & -- & -- & MM & Classification \\ \bottomrule
\end{tabular}%
}
\end{table*}

\subsubsection{Architecture Design}
\label{section-model-arch-design}

As is shown in Figure~\ref{figure-6-adesign}, this class of methods focuses on designing specific modules within the model's architecture to enhance its inherent interpretability. Many approaches leverage simple yet interpretable \textit{surrogate models}, such as decision trees or linear models, integrating them into the architecture to improve explainability. Some methods first predict human-understandable \textit{concepts} and then use these concepts to generate predictions, making the results easier to interpret. Additionally, numerous works build architectures based on \textit{causal frameworks} to embed explainability. Methods that do not fit into the above categories are grouped under the \textit{other designs} category.
The methods summarized are presented in Table~\ref{tab:my-table-2-Architecture Design Methods}.

\textbf{Surrogate Model.}\,\, A common approach is to use a Surrogate Model to serve as a stand-in for complex models during the explanation or decision-making process. The chosen surrogate model can be a decision tree or a simple linear model. These simpler models offer greater transparency in their decision-making processes and can effectively approximate the behavior of complex models, thereby retaining a degree of the high accuracy characteristic of deep neural networks. This method enhances the interpretability of the model, allowing users to better understand its decision mechanisms. Liu et al.\cite{liu2018improving} and Wan et al.\cite{wan2020nbdt} propose using \textit{decision trees} to approximate the behavior of complex models, effectively balancing interpretability and high accuracy by leveraging the strengths of both neural networks and decision trees. Liu et al.\cite{liu2018improving} transfer the knowledge of deep neural networks to decision trees through knowledge distillation. NBDT by Wan et al.\cite{wan2020nbdt} employs differentiable decision sequences and an alternative loss function to replace the final linear layer of the neural network, encouraging the model to learn higher-level concepts and reducing reliance on uncertain decisions. Wong et al.~\cite{wong2021leveraging} utilize elastic net regularization to train a sparse \textit{linear decision layer} on the deep features of pre-trained deep networks, allowing for model behavior debugging by examining less important features and their linear coefficients.

\textbf{Concept-based.}\,\, Koh et al.\cite{koh2020concept} and Yüksekgönül et al.\cite{yuksekgonul2022post} enhanced interpretability by enabling models to predict human-understandable concepts. CBM \cite{koh2020concept} employs interpretable concepts to predict final outputs, applicable to any network architecture, by simply adjusting the number of neurons in a given layer to match the number of concepts while constraining the layer’s output with a loss function. However, CBM has two main drawbacks: (1) it requires dense concept annotations and (2) it may reduce the model's accuracy. PCBM \cite{yuksekgonul2022post} improves upon CBM by (1) allowing concept transfer from other datasets or through multimodal models without needing dense annotations, and (2) incorporating SVM to introduce interpretability while maintaining model performance, addressing both of CBM's limitations. LaBo \cite{yang2023language}, a language-guided concept bottleneck model (CBM) approach, uses GPT-3 to automatically generate interpretable bottleneck concepts aligned with visual data through CLIP, achieving performance comparable to or better than black box models, especially in low-data settings, while maintaining interpretability.

\textbf{Causal-based.}\,\, Some works have introduced causal learning to enhance model interpretability: Li et al.~\cite{li2024steering} proposes a causality-based debiasing framework that guides the design and selection of debiasing prompts using causal insights from the training corpus and LLMs inference. \cite{guo2024trace,chenllcp,li2024multimodal,liu2024revealing} present methods for multimodal tasks, with Chen et al.~\cite{chenllcp} addressing the challenges of extensive data annotation and limited causal reasoning in video QA by introducing the LLCP framework, which analyzes the spatial-temporal dynamics of objects in events. TRACE \cite{guo2024trace} presented a causal event modeling framework that represents videos as event sequences, leveraging prior temporal information, video inputs, and textual instructions to predict current events. MGCE \cite{li2024multimodal} utilized multimodal causal embedding learning networks to enhance the learning of high-quality causal embeddings at both structural and feature levels. Liu et al.~\cite{liu2024revealing} presented a unified causal model specifically designed for multimodal data, showcasing the advantages of multimodal contrastive representation learning in identifying latent coupled variables.

\textbf{Others.}\,\, Before the emergence of MLLMs, several works focused on injecting interpretability into module design: Cognitive Attention Network (CAN) \cite{tang2021interpretable} was introduced to address the Visual Commonsense Reasoning (VCR) task. CAN comprises two key modules: (1) Image-text fusion module: This module integrates information from both images and text, enhancing the model's ability to process multimodal inputs. (2) Inference module: It encodes commonsense relationships among the image, query, and response, enabling the model to reason beyond mere object recognition by understanding relationships between elements using commonsense knowledge. Many works also apply this idea to multimodal models. IA-ViT \cite{qiang2023interpretability} designed a Vision Transformer with enhanced interpretability by analyzing image patches, improving explainability across various visual tasks. MultiModN \cite{swamy2024multimodn} utilized a modular and sequential fusion architecture, enabling clear tracking of each modality's contribution and enhancing both interpretability and robustness against biases from missing data. VL-MoE \cite{shen2023scaling} utilized sparsely-gated Mixture-of-Experts (MoE) to improve efficiency and interpretability in vision-language tasks by dynamically scaling the model based on input modality, offering insights into handling trade-offs between model complexity and performance. IMKGA-SM~\cite{wen2023imkga} enhanced interpretability in multimodal knowledge graph link prediction by employing a sequence modeling approach with fine-grained multimodal fusion and perceptual interaction-based reward mechanisms, achieving efficient and interpretable reasoning in complex multimodal settings.

\section{Training and Inference}
\label{section-training-inference}
We examine training strategies, mechanisms, and inference methods to enhance and analyze the explainability of MLLMs. Training strategies are pivotal for improving model explainability by influencing weight distributions and uncovering feature interactions within the model. Optimizing these strategies lays a solid foundation for future research on explainable AI.
In training, pre-training methods reveal how attention mechanisms and cross-modal alignment enhance understanding.
During inference, techniques such as Chain-of-Thought (CoT) reasoning and in-context learning (ICL) provide structured, interpretable outputs. CoT facilitates step-by-step explanations to minimize hallucinations, while multimodal ICL highlights key representation dynamics, enabling robust, real-time explainability. Together, these approaches enhance the transparency and reliability of MLLMs, fostering their adoption in real-world applications.

\subsection{Training}
\label{section-training-traning}
\textbf{Pre-trained Explainability.} The explainability of pre-trained VLMs is foundational to building robust, transparent AI systems, especially as they are deployed in real-world scenarios requiring explainability. Early research by Value~\cite{cao2020behind} explored how attention mechanisms, through tasks like Visual Coreference Resolution and Visual Relation Detection, contribute to cross-modal and modality-specific alignment, shedding light on the critical role of self-attention patterns in model explainability. Building on this, Salin et al.~\cite{salin2022vision} analyzed pre-trained and fine-tuned VLM representations, revealing inherent biases—particularly concerning object positioning and size—thereby underscoring the need for frameworks that can identify and address these biases for equitable AI applications. Further advancing these explainability efforts, Concept Discovery and Learning (CDL)~\cite{zang2024pre} introduced methods to identify and rank visual concepts based on multimodal data, enhancing models' capacity for interpretable object recognition and broadening their usability in tasks demanding contextual understanding.
Yun et al.~\cite{yun2022vision} investigated how pre-trained models learn basic concepts such as colors and shapes and introduce Compositional Concept Mapping (CompMap) to evaluate explainability in predicting composite concepts. 
LIMA~\cite{zhou2024lima} demonstrated that foundational knowledge and generalization capabilities are established during pre-training, with targeted fine-tuning improving explainability by refining, rather than overhauling, core knowledge. To further address the challenges of large-scale training, DistTrain~\cite{zhang2024disttrain} enhanced explainability and efficiency in multimodal LLMs training by tackling model and data heterogeneity, optimizing resource allocation, and minimizing computational inefficiencies across large-scale clusters. Neo et al.~\cite{neo2024towards} expanded this line of inquiry by examining internal visual processing within VLMs, providing insights into their interpretive mechanisms and further clarifying how these models understand and represent visual information. 

\textbf{Alignment Interpretability.} Effective alignment of vision-language representations is essential for reducing issues like hallucinations and enhancing the reliability of multimodal models. Factually Augmented RLHF~\cite{sun2023aligning} addressed this by incorporating factual data into the training process, minimizing hallucinations, and generating more accurate, interpretable outputs. The ViGoR framework~\cite{yan2024vigor} enhanced interpretability by using fine-grained reward modeling to improve visual grounding, supported by both human and automated evaluations for better accuracy in multimodal tasks.
RLHF has emerged as an effective method for aligning MLLMs with human expectations, thereby improving interpretability. The LLaVA-RLHF model~\cite{sun2023aligning} illustrated this approach by combining human feedback with factual augmentation to reduce hallucinations and enhance model transparency. Expanding on this methodology, RLHF-V~\cite{yu2024rlhf} integrated Dense Direct Preference Optimization (DDPO) to refine policy models further, effectively mitigating hallucinations and boosting robustness in complex multimodal scenarios. Addressing hallucinations is essential for achieving reliable interpretability. HA-DPO~\cite{zhao2023beyond} reduced hallucinations by creating style-consistent hallucination sample pairs and focusing on preference learning, allowing models to prioritize factual accuracy and thereby enhancing the interpretability of outputs. Similarly, Silkie~\cite{li2023silkie} utilized Reinforcement Learning from AI Feedback (RLAIF), drawing on preferences distilled from stronger MLLMs to reinforce faithful and interpretable outputs. RLAIF-V~\cite{yu2024rlaif} also emphasized trustworthiness by aligning MLLMs with open-source feedback and iteratively improving feedback quality through deconfounded response generation. POVID~\cite{zhou2024aligning} innovatively incorporated AI-generated dispreferred data to introduce plausible hallucinations, fostering a nuanced preference optimization framework without human intervention. 

\textbf{Gradient Interpretability.} Gradient-based methods offer a significant approach to enhancing the interpretability of multimodal models by focusing on how models attribute importance to different modalities. SMOOTHGRAD~\cite{smilkov2017smoothgrad} refined gradient-based sensitivity maps by averaging noise-perturbed versions to sharpen the visualization of pixel importance, which is particularly effective in image classification tasks. Building on this, Multimodal Routing~\cite{tsai2020multimodal} enhanced interpretability by dynamically adjusting weights between input modalities and output predictions, enabling both local and global insight into modality-prediction relationships. IFI~\cite{mallick2024ifi} further improved interpretability in transformer-based models by refining feature selection for video and sensor data fusion, which enhances model performance in specific applications such as risk detection and video classification. 


\textbf{Hallucination Explainability.} Recent advancements also focus on reducing undesirable behaviours such as hallucination in multimodal models. Dai et al.~\cite{dai2022plausible_hjh3} introduced a model that mitigates hallucination by utilizing smaller, patch-based features and a novel object-masked language modeling loss. These design choices not only enhance interpretability by reducing model misalignment with reality but also contribute to improved performance, offering a balance between accuracy and clarity in model outputs.
Furthermore, OPERA~\cite{huang2024opera} addressed overconfidence issues by analyzing token interactions by introducing a penalty for excessive trust in summary tokens, effectively reducing hallucinations during decoding and leading to more accurate and reliable interpretations. 
Complementing this, DOPRA~\cite{wei2024dopra} dynamically penalizes excessive token accumulation and employs a backtracking reallocation strategy to align generated content more closely with image data without relying on external resources.


\subsection{Inference}
\label{section-inference}
Recent works~\cite{huang2023survey,liu2024survey,rawte2023survey} have explored the phenomenon of hallucination in LLMs, a pressing issue that affects the reliability of models used in multimodal applications. Hallucination refers to instances where models generate information that appears plausible but is actually incorrect or unsupported by input data. Such issues are particularly complex in MLLMs due to the integration of information from both text and visual modalities. 

\textbf{COT Explainability.} CoT reasoning has emerged as a powerful technique for enhancing interpretability in reasoning tasks, especially within multimodal models. Multimodal-CoT~\cite{zhang2023multimodal} contributes to this area by integrating text and visual information, generating coherent rationales that improve inference accuracy and reduce hallucination. 
Further advancements in CoT reasoning have been made by models that explicitly decouple reasoning steps. Studies like~\cite{ge2023chain,yao2023thinking,zheng2023ddcot} introduce manual separation of CoT reasoning steps, facilitating more nuanced multimodal interactions and enhancing model interpretability. In addition, Visual CoT~\cite{shao2024visual} presents a unique dataset and multi-turn processing pipeline, focusing dynamically on key visual areas to support interpretable reasoning steps in VQA tasks.
More sophisticated CoT frameworks integrate external knowledge structures. For instance, KAM-CoT~\cite{mondal2024kam} leveraged knowledge graphs within CoT reasoning across multiple modalities, dynamically emphasizing critical information to foster transparency in inference. M\textsuperscript{3}CoT~\cite{chen2024m} benchmark addresses challenges across multiple domains and multi-step reasoning scenarios, delivering a robust evaluation framework for understanding complex reasoning. Visual Chain-of-Thought (VCOT)~\cite{rose2023visual} further advances interpretability by generating multimodal synthetic infillings that provide human-interpretable insights, effectively bridging logical gaps in sequential reasoning tasks and improving performance.
Tree-augmented Vision-Language (3VL)~\cite{yellinek20233vl} model, enhancing interpretability and compositional reasoning in vision-language models through a hierarchical tree structure for text representation, along with Anchor and Differential Relevance (DiRe) tools that clarify model behavior by visualizing successes and failures in compositional understanding

\textbf{ICL Explainability.} In-context learning (ICL) capabilities in LLMs offer a unique approach for real-time, contextually relevant responses without the need for retraining~\cite{dong2022survey}. However, challenges remain in achieving uniform interpretability across all model components. For example,~\cite{bansal2022rethinking} explored only specific attention heads and feed-forward networks that contribute significantly to performance.
Addressing explainability in multimodal ICL,~\cite{miyanishi2024multimodal} introduced a multimodal contrastive ICL framework to enhance explainability by employing contrastive learning techniques to reveal key representational dynamics. 

\textbf{Hallucination Explainability.} To address this challenge, a detailed survey~\cite{bai2024hallucination} invsigated hallucination in MLLMs, reviewing its root causes, current evaluation benchmarks, and available mitigation strategies.
To tackle hallucination during inference without requiring additional data or retraining, OPERA~\cite{huang2024opera} introduced an over-trust penalty mechanism and enhanced both interpretability and performance, providing a promising approach for reducing hallucination in MLLMs.
Visual Contrastive Decoding (VCD)~\cite{leng2024mitigating} employed a training-free technique that compares output distributions generated from original versus distorted visual inputs. By emphasizing discrepancies in output consistency, VCD effectively reduced object hallucinations, thereby improving the reliability and explainability of MLLMs outputs.



\section{Future Direction}
\subsection{Dataset and More Modalities}
Future work in multimodal explainability should focus on improving input-output data representation and benchmarking. For input data, standardized preprocessing and annotation pipelines are needed to ensure consistency across modalities like text, images, video, and audio while preserving essential modality-specific features. For outputs, frameworks should generate multimodal explanations, such as natural language rationales with visual or temporal highlights, aligned with human understanding. On benchmarks, future efforts should create task-specific datasets and evaluation protocols that assess explainability across fidelity, comprehensibility, and bias detection, while reflecting real-world complexities, including diverse domains and multilingual datasets.
\subsection{Multimodal Embeddings}
Future work on token-level and embedding-level interpretability in multimodal models should aim to bridge fine-grained interpretability with overall system transparency. At the token level, research should focus on tracing and attributing predictions to specific input tokens across modalities, exploring dynamic token importance mechanisms, and aligning attributions with human reasoning. At the feature level, efforts should enhance the interpretability of intermediate representations, such as visual embeddings and latent textual features, by uncovering meaningful patterns and correlations. Integrating token- and feature-level insights into unified frameworks could provide a comprehensive understanding of how models process multimodal information.
\subsection{Components of MLLMs}
Future research in multimodal neuron analysis should focus on modality alignment mechanisms and efficient model editing. While multimodal neurons can perceive concepts across modalities, the mechanisms behind this remain unclear. Further studies should investigate alignment processes through neuron analysis and develop methods for fine-grained, efficient neuron editing. Extending this analysis to circuits could reveal interconnections between units, offering deeper insights into model behavior.
For layer-level interpretability, future work should explore the roles of components and workflows in cross-modal decision-making. This includes understanding how various encoders (e.g., vision, audio, point cloud) and projectors align non-text inputs with LLMs' text space. Additionally, research should clarify how post-projection embeddings are processed, identify layers handling cross-modal inputs, and analyze their impact on LLM inference capabilities.  
\subsection{Model Archtectures}
Future work on architecture-level multimodal interpretability should focus on enhancing the transparency of multimodal models by investigating the specific roles of different architectural components in processing cross-modal information. This includes exploring how various encoders, such as vision, audio, and point cloud encoders, interact with one another and align their outputs within the text space of LLMs. It is essential to understand the flow of information from raw modality inputs to their integrated representations and to uncover how these components contribute to the final decision-making process. Additionally, examining the functionality of post-projection embeddings and identifying which layers are responsible for processing multimodal inputs will be critical in revealing the underlying mechanisms of cross-modal inference. Such insights could pave the way for more interpretable architectures that facilitate trust and understanding while improving the reliability of multimodal models in real-world applications.
\subsection{Training Dynamics and Inferencing}
Future work in multimodal explainability should focus on unified frameworks that integrate interpretability into training and inference. During training, models should prioritize transparency and alignment with human understanding while maintaining scalability. Inference should provide real-time, task-adaptive explanations to enhance trust and clarity. Robust benchmarks for evaluating interpretability at both stages will be essential, enabling the development of transparent, reliable, and high-performing multimodal systems for real-world applications.


\section{Conclusion}




This survey systematically explores the interpretability and explainability of MLLMs, emphasizing the importance of transparency in their decision-making. We categorized interpretability methods into three main areas—data, model, and training and inference—offering a structured framework to organize research and guide future studies.
While significant progress has been made, challenges remain in 
explainability and interpretability methods and ensuring broad applicability. Future efforts should address these gaps to build a unified understanding of MLLMs, fostering innovations that make multimodal systems more reliable and trustworthy.


\section{References Section}

%

\bibliographystyle{IEEEtran}   
\bibliography{references}      

\begin{thebibliography}{100}
\providecommand{\url}[1]{#1}
\csname url@samestyle\endcsname
\providecommand{\newblock}{\relax}
\providecommand{\bibinfo}[2]{#2}
\providecommand{\BIBentrySTDinterwordspacing}{\spaceskip=0pt\relax}
\providecommand{\BIBentryALTinterwordstretchfactor}{4}
\providecommand{\BIBentryALTinterwordspacing}{\spaceskip=\fontdimen2\font plus
\BIBentryALTinterwordstretchfactor\fontdimen3\font minus \fontdimen4\font\relax}
\providecommand{\BIBforeignlanguage}[2]{{%
\expandafter\ifx\csname l@#1\endcsname\relax
\typeout{** WARNING: IEEEtran.bst: No hyphenation pattern has been}%
\typeout{** loaded for the language `#1'. Using the pattern for}%
\typeout{** the default language instead.}%
\else
\language=\csname l@#1\endcsname
\fi
#2}}
\providecommand{\BIBdecl}{\relax}
\BIBdecl

\bibitem{zhao2023survey}
W.~X. Zhao, K.~Zhou, J.~Li, T.~Tang, X.~Wang, Y.~Hou, Y.~Min, B.~Zhang, J.~Zhang, Z.~Dong \emph{et~al.}, ``A survey of large language models,'' \emph{arXiv preprint arXiv:2303.18223}, 2023.

\bibitem{voulodimos2018deep}
A.~Voulodimos, N.~Doulamis, A.~Doulamis, and E.~Protopapadakis, ``Deep learning for computer vision: A brief review,'' \emph{Computational intelligence and neuroscience}, vol. 2018, no.~1, p. 7068349, 2018.

\bibitem{cui2024survey}
C.~Cui, Y.~Ma, X.~Cao, W.~Ye, Y.~Zhou, K.~Liang, J.~Chen, J.~Lu, Z.~Yang, K.-D. Liao \emph{et~al.}, ``A survey on multimodal large language models for autonomous driving,'' in \emph{Proceedings of the IEEE/CVF Winter Conference on Applications of Computer Vision}, 2024, pp. 958--979.

\bibitem{jin2024efficient}
Y.~Jin, J.~Li, Y.~Liu, T.~Gu, K.~Wu, Z.~Jiang, M.~He, B.~Zhao, X.~Tan, Z.~Gan \emph{et~al.}, ``Efficient multimodal large language models: A survey,'' \emph{arXiv preprint arXiv:2405.10739}, 2024.

\bibitem{caffagni2024r}
D.~Caffagni, F.~Cocchi, L.~Barsellotti, N.~Moratelli, S.~Sarto, L.~Baraldi, M.~Cornia, and R.~Cucchiara, ``The (r) evolution of multimodal large language models: A survey,'' \emph{arXiv preprint arXiv:2402.12451}, 2024.

\bibitem{zhang2024mm}
D.~Zhang, Y.~Yu, J.~Dong, C.~Li, D.~Su, C.~Chu, and D.~Yu, ``Mm-llms: Recent advances in multimodal large language models,'' \emph{arXiv preprint arXiv:2401.13601}, 2024.

\bibitem{wu2024visual}
J.~Wu, Z.~Zhang, Y.~Xia, X.~Li, Z.~Xia, A.~Chang, T.~Yu, S.~Kim, R.~A. Rossi, R.~Zhang \emph{et~al.}, ``Visual prompting in multimodal large language models: A survey,'' \emph{arXiv preprint arXiv:2409.15310}, 2024.

\bibitem{xie2024large}
J.~Xie, Z.~Chen, R.~Zhang, X.~Wan, and G.~Li, ``Large multimodal agents: A survey,'' \emph{arXiv preprint arXiv:2402.15116}, 2024.

\bibitem{yan2024errorradar}
Y.~Yan, S.~Wang, J.~Huo, H.~Li, B.~Li, J.~Su, X.~Gao, Y.-F. Zhang, T.~Xu, Z.~Chu \emph{et~al.}, ``Errorradar: Benchmarking complex mathematical reasoning of multimodal large language models via error detection,'' \emph{arXiv preprint arXiv:2410.04509}, 2024.

\bibitem{yan2024urbanclip}
Y.~Yan, H.~Wen, S.~Zhong, W.~Chen, H.~Chen, Q.~Wen, R.~Zimmermann, and Y.~Liang, ``Urbanclip: Learning text-enhanced urban region profiling with contrastive language-image pretraining from the web,'' in \emph{Proceedings of the ACM on Web Conference 2024}, 2024, pp. 4006--4017.

\bibitem{zheng2024reefknot}
K.~Zheng, J.~Chen, Y.~Yan, X.~Zou, and X.~Hu, ``Reefknot: A comprehensive benchmark for relation hallucination evaluation, analysis and mitigation in multimodal large language models,'' \emph{arXiv preprint arXiv:2408.09429}, 2024.

\bibitem{yin2023survey}
S.~Yin, C.~Fu, S.~Zhao, K.~Li, X.~Sun, T.~Xu, and E.~Chen, ``A survey on multimodal large language models,'' \emph{arXiv preprint arXiv:2306.13549}, 2023.

\bibitem{wu2023multimodal}
J.~Wu, W.~Gan, Z.~Chen, S.~Wan, and S.~Y. Philip, ``Multimodal large language models: A survey,'' in \emph{2023 IEEE International Conference on Big Data (BigData)}.\hskip 1em plus 0.5em minus 0.4em\relax IEEE, 2023, pp. 2247--2256.

\bibitem{zou2025deep}
X.~Zou, Y.~Yan, X.~Hao, Y.~Hu, H.~Wen, E.~Liu, J.~Zhang, Y.~Li, T.~Li, Y.~Zheng \emph{et~al.}, ``Deep learning for cross-domain data fusion in urban computing: Taxonomy, advances, and outlook,'' \emph{Information Fusion}, vol. 113, p. 102606, 2025.

\bibitem{song2023bridge}
S.~Song, X.~Li, S.~Li, S.~Zhao, J.~Yu, J.~Ma, X.~Mao, and W.~Zhang, ``How to bridge the gap between modalities: A comprehensive survey on multimodal large language model,'' \emph{arXiv preprint arXiv:2311.07594}, 2023.

\bibitem{zou2024look}
X.~Zou, Y.~Wang, Y.~Yan, S.~Huang, K.~Zheng, J.~Chen, C.~Tang, and X.~Hu, ``Look twice before you answer: Memory-space visual retracing for hallucination mitigation in multimodal large language models,'' \emph{arXiv preprint arXiv:2410.03577}, 2024.

\bibitem{zhou2024mitigating}
G.~Zhou, Y.~Yan, X.~Zou, K.~Wang, A.~Liu, and X.~Hu, ``Mitigating modality prior-induced hallucinations in multimodal large language models via deciphering attention causality,'' \emph{arXiv preprint arXiv:2410.04780}, 2024.

\bibitem{huang2021unifying}
Y.~Huang, H.~Xue, B.~Liu, and Y.~Lu, ``Unifying multimodal transformer for bi-directional image and text generation,'' in \emph{Proceedings of the 29th ACM International Conference on Multimedia}, 2021, pp. 1138--1147.

\bibitem{koh2024generating}
J.~Y. Koh, D.~Fried, and R.~R. Salakhutdinov, ``Generating images with multimodal language models,'' \emph{Advances in Neural Information Processing Systems}, vol.~36, 2024.

\bibitem{hu2024instruct}
H.~Hu, K.~C. Chan, Y.-C. Su, W.~Chen, Y.~Li, K.~Sohn, Y.~Zhao, X.~Ben, B.~Gong, W.~Cohen \emph{et~al.}, ``Instruct-imagen: Image generation with multi-modal instruction,'' in \emph{Proceedings of the IEEE/CVF Conference on Computer Vision and Pattern Recognition}, 2024, pp. 4754--4763.

\bibitem{fu2023mme}
C.~Fu, P.~Chen, Y.~Shen, Y.~Qin, M.~Zhang, X.~Lin, J.~Yang, X.~Zheng, K.~Li, X.~Sun \emph{et~al.}, ``Mme: A comprehensive evaluation benchmark for multimodal large language models,'' \emph{arXiv preprint arXiv:2306.13394}, 2023.

\bibitem{liu2025mmbench}
Y.~Liu, H.~Duan, Y.~Zhang, B.~Li, S.~Zhang, W.~Zhao, Y.~Yuan, J.~Wang, C.~He, Z.~Liu \emph{et~al.}, ``Mmbench: Is your multi-modal model an all-around player?'' in \emph{European Conference on Computer Vision}.\hskip 1em plus 0.5em minus 0.4em\relax Springer, 2025, pp. 216--233.

\bibitem{yue2024mmmu}
X.~Yue, Y.~Ni, K.~Zhang, T.~Zheng, R.~Liu, G.~Zhang, S.~Stevens, D.~Jiang, W.~Ren, Y.~Sun \emph{et~al.}, ``Mmmu: A massive multi-discipline multimodal understanding and reasoning benchmark for expert agi,'' in \emph{Proceedings of the IEEE/CVF Conference on Computer Vision and Pattern Recognition}, 2024, pp. 9556--9567.

\bibitem{lu2023mathvista}
P.~Lu, H.~Bansal, T.~Xia, J.~Liu, C.~Li, H.~Hajishirzi, H.~Cheng, K.-W. Chang, M.~Galley, and J.~Gao, ``Mathvista: Evaluating mathematical reasoning of foundation models in visual contexts,'' \emph{arXiv preprint arXiv:2310.02255}, 2023.

\bibitem{zhang2024unveiling}
Y.~Zhang, F.~Xiao, T.~Huang, C.-K. Fan, H.~Dong, J.~Li, J.~Wang, K.~Cheng, S.~Zhang, and H.~Guo, ``Unveiling the tapestry of consistency in large vision-language models,'' \emph{arXiv preprint arXiv:2405.14156}, 2024.

\bibitem{li2024seed2plus}
B.~Li, Y.~Ge, Y.~Chen, Y.~Ge, R.~Zhang, and Y.~Shan, ``Seed-bench-2-plus: Benchmarking multimodal large language models with text-rich visual comprehension,'' \emph{arXiv preprint arXiv:2404.16790}, 2024.

\bibitem{li2023seed2}
B.~Li, Y.~Ge, Y.~Ge, G.~Wang, R.~Wang, R.~Zhang, and Y.~Shan, ``Seed-bench-2: Benchmarking multimodal large language models,'' \emph{arXiv preprint arXiv:2311.17092}, 2023.

\bibitem{li2023seed}
B.~Li, R.~Wang, G.~Wang, Y.~Ge, Y.~Ge, and Y.~Shan, ``Seed-bench: Benchmarking multimodal llms with generative comprehension,'' \emph{arXiv preprint arXiv:2307.16125}, 2023.

\bibitem{wang2016comprehensive}
K.~Wang, Q.~Yin, W.~Wang, S.~Wu, and L.~Wang, ``A comprehensive survey on cross-modal retrieval,'' \emph{arXiv preprint arXiv:1607.06215}, 2016.

\bibitem{zhen2019deep}
L.~Zhen, P.~Hu, X.~Wang, and D.~Peng, ``Deep supervised cross-modal retrieval,'' in \emph{Proceedings of the IEEE/CVF conference on computer vision and pattern recognition}, 2019, pp. 10\,394--10\,403.

\bibitem{chun2021probabilistic}
S.~Chun, S.~J. Oh, R.~S. De~Rezende, Y.~Kalantidis, and D.~Larlus, ``Probabilistic embeddings for cross-modal retrieval,'' in \emph{Proceedings of the IEEE/CVF Conference on Computer Vision and Pattern Recognition}, 2021, pp. 8415--8424.

\bibitem{abdu2021multimodal}
S.~A. Abdu, A.~H. Yousef, and A.~Salem, ``Multimodal video sentiment analysis using deep learning approaches, a survey,'' \emph{Information Fusion}, vol.~76, pp. 204--226, 2021.

\bibitem{khalid2021fakeavceleb}
H.~Khalid, S.~Tariq, M.~Kim, and S.~S. Woo, ``Fakeavceleb: A novel audio-video multimodal deepfake dataset,'' \emph{arXiv preprint arXiv:2108.05080}, 2021.

\bibitem{seo2022end}
P.~H. Seo, A.~Nagrani, A.~Arnab, and C.~Schmid, ``End-to-end generative pretraining for multimodal video captioning,'' in \emph{Proceedings of the IEEE/CVF Conference on Computer Vision and Pattern Recognition}, 2022, pp. 17\,959--17\,968.

\bibitem{fan2025videoagent}
Y.~Fan, X.~Ma, R.~Wu, Y.~Du, J.~Li, Z.~Gao, and Q.~Li, ``Videoagent: A memory-augmented multimodal agent for video understanding,'' in \emph{European Conference on Computer Vision}.\hskip 1em plus 0.5em minus 0.4em\relax Springer, 2025, pp. 75--92.

\bibitem{dzabraev2021mdmmt}
M.~Dzabraev, M.~Kalashnikov, S.~Komkov, and A.~Petiushko, ``Mdmmt: Multidomain multimodal transformer for video retrieval,'' in \emph{Proceedings of the IEEE/CVF conference on computer vision and pattern recognition}, 2021, pp. 3354--3363.

\bibitem{botach2022end}
A.~Botach, E.~Zheltonozhskii, and C.~Baskin, ``End-to-end referring video object segmentation with multimodal transformers,'' in \emph{Proceedings of the IEEE/CVF Conference on Computer Vision and Pattern Recognition}, 2022, pp. 4985--4995.

\bibitem{luo2020univl}
H.~Luo, L.~Ji, B.~Shi, H.~Huang, N.~Duan, T.~Li, J.~Li, T.~Bharti, and M.~Zhou, ``Univl: A unified video and language pre-training model for multimodal understanding and generation,'' \emph{arXiv preprint arXiv:2002.06353}, 2020.

\bibitem{chen2024evolution}
Z.~Chen, L.~Xu, H.~Zheng, L.~Chen, A.~Tolba, L.~Zhao, K.~Yu, and H.~Feng, ``Evolution and prospects of foundation models: From large language models to large multimodal models.'' \emph{Computers, Materials \& Continua}, vol.~80, no.~2, 2024.

\bibitem{wang2024comprehensive}
J.~Wang, H.~Jiang, Y.~Liu, C.~Ma, X.~Zhang, Y.~Pan, M.~Liu, P.~Gu, S.~Xia, W.~Li \emph{et~al.}, ``A comprehensive review of multimodal large language models: Performance and challenges across different tasks,'' \emph{arXiv preprint arXiv:2408.01319}, 2024.

\bibitem{yan2024georeasoner}
Y.~Yan and J.~Lee, ``Georeasoner: Reasoning on geospatially grounded context for natural language understanding,'' in \emph{Proceedings of the 33rd ACM International Conference on Information and Knowledge Management}, 2024, pp. 4163--4167.

\bibitem{huang2023chatgpt}
H.~Huang, O.~Zheng, D.~Wang, J.~Yin, Z.~Wang, S.~Ding, H.~Yin, C.~Xu, R.~Yang, Q.~Zheng \emph{et~al.}, ``Chatgpt for shaping the future of dentistry: the potential of multi-modal large language model,'' \emph{International Journal of Oral Science}, vol.~15, no.~1, p.~29, 2023.

\bibitem{ye2023mplug}
J.~Ye, A.~Hu, H.~Xu, Q.~Ye, M.~Yan, Y.~Dan, C.~Zhao, G.~Xu, C.~Li, J.~Tian \emph{et~al.}, ``mplug-docowl: Modularized multimodal large language model for document understanding,'' \emph{arXiv preprint arXiv:2307.02499}, 2023.

\bibitem{bayoudh2022survey}
K.~Bayoudh, R.~Knani, F.~Hamdaoui, and A.~Mtibaa, ``A survey on deep multimodal learning for computer vision: advances, trends, applications, and datasets,'' \emph{The Visual Computer}, vol.~38, no.~8, pp. 2939--2970, 2022.

\bibitem{bayoudh2023survey}
K.~Bayoudh, ``A survey of multimodal hybrid deep learning for computer vision: Architectures, applications, trends, and challenges,'' \emph{Information Fusion}, p. 102217, 2023.

\bibitem{xu2023multimodal}
P.~Xu, X.~Zhu, and D.~A. Clifton, ``Multimodal learning with transformers: A survey,'' \emph{IEEE Transactions on Pattern Analysis and Machine Intelligence}, vol.~45, no.~10, pp. 12\,113--12\,132, 2023.

\bibitem{tang2023video}
Y.~Tang, J.~Bi, S.~Xu, L.~Song, S.~Liang, T.~Wang, D.~Zhang, J.~An, J.~Lin, R.~Zhu \emph{et~al.}, ``Video understanding with large language models: A survey,'' \emph{arXiv preprint arXiv:2312.17432}, 2023.

\bibitem{nie2025reason2drive}
M.~Nie, R.~Peng, C.~Wang, X.~Cai, J.~Han, H.~Xu, and L.~Zhang, ``Reason2drive: Towards interpretable and chain-based reasoning for autonomous driving,'' in \emph{European Conference on Computer Vision}.\hskip 1em plus 0.5em minus 0.4em\relax Springer, 2025, pp. 292--308.

\bibitem{li2024cog}
Z.~Li, Y.~Lu, Y.~Mu, and H.~Qiao, ``Cog-ga: A large language models-based generative agent for vision-language navigation in continuous environments,'' \emph{arXiv preprint arXiv:2409.02522}, 2024.

\bibitem{wang2024interpretable}
L.~Wang, H.~Wang, H.~Yang, J.~Mao, Z.~Yang, J.~Shen, and X.~Li, ``Interpretable bilingual multimodal large language model for diverse biomedical tasks,'' \emph{arXiv preprint arXiv:2410.18387}, 2024.

\bibitem{liu2024mcan}
J.~Liu, L.~Han, and J.~Ji, ``Mcan: multimodal causal adversarial networks for dynamic effective connectivity learning from fmri and eeg data,'' \emph{IEEE Transactions on Medical Imaging}, 2024.

\bibitem{xiao2024comprehensive}
H.~Xiao, F.~Zhou, X.~Liu, T.~Liu, Z.~Li, X.~Liu, and X.~Huang, ``A comprehensive survey of large language models and multimodal large language models in medicine,'' \emph{arXiv preprint arXiv:2405.08603}, 2024.

\bibitem{li2024mmro}
J.~Li, Y.~Zhu, Z.~Xu, J.~Gu, M.~Zhu, X.~Liu, N.~Liu, Y.~Peng, F.~Feng, and J.~Tang, ``Mmro: Are multimodal llms eligible as the brain for in-home robotics?'' \emph{arXiv preprint arXiv:2406.19693}, 2024.

\bibitem{li2021toward}
S.~Li, P.~Zheng, J.~Fan, and L.~Wang, ``Toward proactive human--robot collaborative assembly: A multimodal transfer-learning-enabled action prediction approach,'' \emph{IEEE Transactions on Industrial Electronics}, vol.~69, no.~8, pp. 8579--8588, 2021.

\bibitem{xue2020progress}
T.~Xue, W.~Wang, J.~Ma, W.~Liu, Z.~Pan, and M.~Han, ``Progress and prospects of multimodal fusion methods in physical human--robot interaction: A review,'' \emph{IEEE Sensors Journal}, vol.~20, no.~18, pp. 10\,355--10\,370, 2020.

\bibitem{duan2022multimodal}
S.~Duan, Q.~Shi, and J.~Wu, ``Multimodal sensors and ml-based data fusion for advanced robots,'' \emph{Advanced Intelligent Systems}, vol.~4, no.~12, p. 2200213, 2022.

\bibitem{wake2024gpt}
N.~Wake, A.~Kanehira, K.~Sasabuchi, J.~Takamatsu, and K.~Ikeuchi, ``Gpt-4v (ision) for robotics: Multimodal task planning from human demonstration,'' \emph{IEEE Robotics and Automation Letters}, 2024.

\bibitem{sermanet2024robovqa}
P.~Sermanet, T.~Ding, J.~Zhao, F.~Xia, D.~Dwibedi, K.~Gopalakrishnan, C.~Chan, G.~Dulac-Arnold, S.~Maddineni, N.~J. Joshi \emph{et~al.}, ``Robovqa: Multimodal long-horizon reasoning for robotics,'' in \emph{2024 IEEE International Conference on Robotics and Automation (ICRA)}.\hskip 1em plus 0.5em minus 0.4em\relax IEEE, 2024, pp. 645--652.

\bibitem{zwilling2019enhanced}
C.~E. Zwilling, A.~M. Daugherty, C.~H. Hillman, A.~F. Kramer, N.~J. Cohen, and A.~K. Barbey, ``Enhanced decision-making through multimodal training,'' \emph{NPJ science of learning}, vol.~4, no.~1, p.~11, 2019.

\bibitem{zhao2024deep}
F.~Zhao, C.~Zhang, and B.~Geng, ``Deep multimodal data fusion,'' \emph{ACM Computing Surveys}, vol.~56, no.~9, pp. 1--36, 2024.

\bibitem{arrieta2020explainable}
A.~B. Arrieta, N.~D{\'\i}az-Rodr{\'\i}guez, J.~Del~Ser, A.~Bennetot, S.~Tabik, A.~Barbado, S.~Garc{\'\i}a, S.~Gil-L{\'o}pez, D.~Molina, R.~Benjamins \emph{et~al.}, ``Explainable artificial intelligence (xai): Concepts, taxonomies, opportunities and challenges toward responsible ai,'' \emph{Information fusion}, vol.~58, pp. 82--115, 2020.

\bibitem{das2020opportunities}
A.~Das and P.~Rad, ``Opportunities and challenges in explainable artificial intelligence (xai): A survey,'' \emph{arXiv preprint arXiv:2006.11371}, 2020.

\bibitem{adadi2018peeking}
A.~Adadi and M.~Berrada, ``Peeking inside the black-box: a survey on explainable artificial intelligence (xai),'' \emph{IEEE access}, vol.~6, pp. 52\,138--52\,160, 2018.

\bibitem{doshi2017towards}
F.~Doshi-Velez and B.~Kim, ``Towards a rigorous science of interpretable machine learning,'' \emph{arXiv preprint arXiv:1702.08608}, 2017.

\bibitem{du2019techniques}
M.~Du, N.~Liu, and X.~Hu, ``Techniques for interpretable machine learning,'' \emph{Communications of the ACM}, vol.~63, no.~1, pp. 68--77, 2019.

\bibitem{zeiler2014visualizing}
M.~D. Zeiler and R.~Fergus, ``Visualizing and understanding convolutional networks,'' in \emph{Computer Vision--ECCV 2014: 13th European Conference, Zurich, Switzerland, September 6-12, 2014, Proceedings, Part I 13}.\hskip 1em plus 0.5em minus 0.4em\relax Springer, 2014, pp. 818--833.

\bibitem{zhang2018interpretable}
Q.~Zhang, Y.~N. Wu, and S.-C. Zhu, ``Interpretable convolutional neural networks,'' in \emph{Proceedings of the IEEE conference on computer vision and pattern recognition}, 2018, pp. 8827--8836.

\bibitem{chefer2021transformer}
H.~Chefer, S.~Gur, and L.~Wolf, ``Transformer interpretability beyond attention visualization,'' in \emph{Proceedings of the IEEE/CVF conference on computer vision and pattern recognition}, 2021, pp. 782--791.

\bibitem{zhao2024explainability}
H.~Zhao, H.~Chen, F.~Yang, N.~Liu, H.~Deng, H.~Cai, S.~Wang, D.~Yin, and M.~Du, ``Explainability for large language models: A survey,'' \emph{ACM Transactions on Intelligent Systems and Technology}, vol.~15, no.~2, pp. 1--38, 2024.

\bibitem{nauta2023anecdotal}
M.~Nauta, J.~Trienes, S.~Pathak, E.~Nguyen, M.~Peters, Y.~Schmitt, J.~Schl{\"o}tterer, M.~Van~Keulen, and C.~Seifert, ``From anecdotal evidence to quantitative evaluation methods: A systematic review on evaluating explainable ai,'' \emph{ACM Computing Surveys}, vol.~55, no. 13s, pp. 1--42, 2023.

\bibitem{lipton2018mythos}
Z.~C. Lipton, ``The mythos of model interpretability: In machine learning, the concept of interpretability is both important and slippery.'' \emph{Queue}, vol.~16, no.~3, pp. 31--57, 2018.

\bibitem{kaur2020interpreting}
H.~Kaur, H.~Nori, S.~Jenkins, R.~Caruana, H.~Wallach, and J.~Wortman~Vaughan, ``Interpreting interpretability: understanding data scientists' use of interpretability tools for machine learning,'' in \emph{Proceedings of the 2020 CHI conference on human factors in computing systems}, 2020, pp. 1--14.

\bibitem{zhou2008low}
S.-M. Zhou and J.~Q. Gan, ``Low-level interpretability and high-level interpretability: a unified view of data-driven interpretable fuzzy system modelling,'' \emph{Fuzzy sets and systems}, vol. 159, no.~23, pp. 3091--3131, 2008.

\bibitem{guillaume2001designing}
S.~Guillaume, ``Designing fuzzy inference systems from data: An interpretability-oriented review,'' \emph{IEEE Transactions on fuzzy systems}, vol.~9, no.~3, pp. 426--443, 2001.

\bibitem{frye2020shapley}
C.~Frye, D.~de~Mijolla, T.~Begley, L.~Cowton, M.~Stanley, and I.~Feige, ``Shapley explainability on the data manifold,'' \emph{arXiv preprint arXiv:2006.01272}, 2020.

\bibitem{hooker2019benchmark}
S.~Hooker, D.~Erhan, P.-J. Kindermans, and B.~Kim, ``A benchmark for interpretability methods in deep neural networks,'' \emph{Advances in neural information processing systems}, vol.~32, 2019.

\bibitem{ismail2020benchmarking}
A.~A. Ismail, M.~Gunady, H.~Corrada~Bravo, and S.~Feizi, ``Benchmarking deep learning interpretability in time series predictions,'' \emph{Advances in neural information processing systems}, vol.~33, pp. 6441--6452, 2020.

\bibitem{chakraborty2017interpretability}
S.~Chakraborty, R.~Tomsett, R.~Raghavendra, D.~Harborne, M.~Alzantot, F.~Cerutti, M.~Srivastava, A.~Preece, S.~Julier, R.~M. Rao \emph{et~al.}, ``Interpretability of deep learning models: A survey of results,'' in \emph{2017 IEEE smartworld, ubiquitous intelligence \& computing, advanced \& trusted computed, scalable computing \& communications, cloud \& big data computing, Internet of people and smart city innovation (smartworld/SCALCOM/UIC/ATC/CBDcom/IOP/SCI)}.\hskip 1em plus 0.5em minus 0.4em\relax IEEE, 2017, pp. 1--6.

\bibitem{zhang2021survey}
Y.~Zhang, P.~Ti{\v{n}}o, A.~Leonardis, and K.~Tang, ``A survey on neural network interpretability,'' \emph{IEEE Transactions on Emerging Topics in Computational Intelligence}, vol.~5, no.~5, pp. 726--742, 2021.

\bibitem{fan2021interpretability}
F.-L. Fan, J.~Xiong, M.~Li, and G.~Wang, ``On interpretability of artificial neural networks: A survey,'' \emph{IEEE Transactions on Radiation and Plasma Medical Sciences}, vol.~5, no.~6, pp. 741--760, 2021.

\bibitem{linardatos2020explainable}
P.~Linardatos, V.~Papastefanopoulos, and S.~Kotsiantis, ``Explainable ai: A review of machine learning interpretability methods,'' \emph{Entropy}, vol.~23, no.~1, p.~18, 2020.

\bibitem{marcinkevivcs2020interpretability}
R.~Marcinkevi{\v{c}}s and J.~E. Vogt, ``Interpretability and explainability: A machine learning zoo mini-tour,'' \emph{arXiv preprint arXiv:2012.01805}, 2020.

\bibitem{dovsilovic2018explainable}
F.~K. Do{\v{s}}ilovi{\'c}, M.~Br{\v{c}}i{\'c}, and N.~Hlupi{\'c}, ``Explainable artificial intelligence: A survey,'' in \emph{2018 41st International convention on information and communication technology, electronics and microelectronics (MIPRO)}.\hskip 1em plus 0.5em minus 0.4em\relax IEEE, 2018, pp. 0210--0215.

\bibitem{gilpin2018explaining}
L.~H. Gilpin, D.~Bau, B.~Z. Yuan, A.~Bajwa, M.~Specter, and L.~Kagal, ``Explaining explanations: An overview of interpretability of machine learning,'' in \emph{2018 IEEE 5th International Conference on data science and advanced analytics (DSAA)}.\hskip 1em plus 0.5em minus 0.4em\relax IEEE, 2018, pp. 80--89.

\bibitem{madsen2022post}
A.~Madsen, S.~Reddy, and S.~Chandar, ``Post-hoc interpretability for neural nlp: A survey,'' \emph{ACM Computing Surveys}, vol.~55, no.~8, pp. 1--42, 2022.

\bibitem{bibal2016interpretability}
A.~Bibal and B.~Fr{\'e}nay, ``Interpretability of machine learning models and representations: an introduction,'' in \emph{24th european symposium on artificial neural networks, computational intelligence and machine learning}.\hskip 1em plus 0.5em minus 0.4em\relax CIACO, 2016, pp. 77--82.

\bibitem{chen2016infogan}
X.~Chen, Y.~Duan, R.~Houthooft, J.~Schulman, I.~Sutskever, and P.~Abbeel, ``Infogan: Interpretable representation learning by information maximizing generative adversarial nets,'' \emph{Advances in neural information processing systems}, vol.~29, 2016.

\bibitem{zou2310representation}
A.~Zou, L.~Phan, S.~Chen, J.~Campbell, P.~Guo, R.~Ren, A.~Pan, X.~Yin, M.~Mazeika, A.-K. Dombrowski \emph{et~al.}, ``Representation engineering: A top-down approach to ai transparency. corr, abs/2310.01405, 2023. doi: 10.48550,'' \emph{arXiv preprint ARXIV.2310.01405}, 2024.

\bibitem{qian2024towards}
C.~Qian, J.~Zhang, W.~Yao, D.~Liu, Z.~Yin, Y.~Qiao, Y.~Liu, and J.~Shao, ``Towards tracing trustworthiness dynamics: Revisiting pre-training period of large language models,'' \emph{arXiv preprint arXiv:2402.19465}, 2024.

\bibitem{zhang2018visual}
Q.-s. Zhang and S.-C. Zhu, ``Visual interpretability for deep learning: a survey,'' \emph{Frontiers of Information Technology \& Electronic Engineering}, vol.~19, no.~1, pp. 27--39, 2018.

\bibitem{sajjad2022neuron}
H.~Sajjad, N.~Durrani, and F.~Dalvi, ``Neuron-level interpretation of deep nlp models: A survey,'' \emph{Transactions of the Association for Computational Linguistics}, vol.~10, pp. 1285--1303, 2022.

\bibitem{bai2024improving}
H.~Bai and Y.~Ma, ``Improving neuron-level interpretability with white-box language models,'' \emph{arXiv preprint arXiv:2410.16443}, 2024.

\bibitem{qian2024dean}
C.~Qian, D.~Liu, J.~Zhang, Y.~Liu, and J.~Shao, ``Dean: Deactivating the coupled neurons to mitigate fairness-privacy conflicts in large language models,'' \emph{arXiv preprint arXiv:2410.16672}, 2024.

\bibitem{ren2024identifying}
J.~Ren, Q.~Guo, H.~Yan, D.~Liu, Q.~Zhang, X.~Qiu, and D.~Lin, ``Identifying semantic induction heads to understand in-context learning,'' \emph{arXiv preprint arXiv:2402.13055}, 2024.

\bibitem{ribeiro2016model}
M.~T. Ribeiro, S.~Singh, and C.~Guestrin, ``Model-agnostic interpretability of machine learning,'' \emph{arXiv preprint arXiv:1606.05386}, 2016.

\bibitem{danesh2023hybridization}
T.~Danesh, R.~Ouaret, P.~Floquet, and S.~N{\'e}gny, ``Hybridization of model-specific and model-agnostic methods for interpretability of neural network predictions: Application to a power plant,'' \emph{Computers \& Chemical Engineering}, vol. 176, p. 108306, 2023.

\bibitem{rumbelow2024model}
J.~Rumbelow, ``Model agnostic interpretability,'' Ph.D. dissertation, The University of St Andrews, 2024.

\bibitem{early2022model}
J.~Early, C.~Evers, and S.~Ramchurn, ``Model agnostic interpretability for multiple instance learning,'' \emph{arXiv preprint arXiv:2201.11701}, 2022.

\bibitem{li2024survey}
J.~Li, W.~Lu, H.~Fei, M.~Luo, M.~Dai, M.~Xia, Y.~Jin, Z.~Gan, D.~Qi, C.~Fu \emph{et~al.}, ``A survey on benchmarks of multimodal large language models,'' \emph{arXiv preprint arXiv:2408.08632}, 2024.

\bibitem{han2022survey}
K.~Han, Y.~Wang, H.~Chen, X.~Chen, J.~Guo, Z.~Liu, Y.~Tang, A.~Xiao, C.~Xu, Y.~Xu \emph{et~al.}, ``A survey on vision transformer,'' \emph{IEEE transactions on pattern analysis and machine intelligence}, vol.~45, no.~1, pp. 87--110, 2022.

\bibitem{kashefi2023explainability}
R.~Kashefi, L.~Barekatain, M.~Sabokrou, and F.~Aghaeipoor, ``Explainability of vision transformers: A comprehensive review and new perspectives,'' \emph{arXiv preprint arXiv:2311.06786}, 2023.

\bibitem{park2018multimodal}
D.~H. Park, L.~A. Hendricks, Z.~Akata, A.~Rohrbach, B.~Schiele, T.~Darrell, and M.~Rohrbach, ``Multimodal explanations: Justifying decisions and pointing to the evidence,'' in \emph{Proceedings of the IEEE conference on computer vision and pattern recognition}, 2018, pp. 8779--8788.

\bibitem{leng2024mitigating}
S.~Leng, H.~Zhang, G.~Chen, X.~Li, S.~Lu, C.~Miao, and L.~Bing, ``Mitigating object hallucinations in large vision-language models through visual contrastive decoding,'' in \emph{Proceedings of the IEEE/CVF Conference on Computer Vision and Pattern Recognition}, 2024, pp. 13\,872--13\,882.

\bibitem{walker2023causal}
E.~Walker, J.~A. Actor, C.~Martinez, and N.~Trask, ``Causal disentanglement of multimodal data,'' \emph{arXiv preprint arXiv:2310.18471}, 2023.

\bibitem{klaassen2024doublemldeep}
S.~Klaassen, J.~Teichert-Kluge, P.~Bach, V.~Chernozhukov, M.~Spindler, and S.~Vijaykumar, ``Doublemldeep: Estimation of causal effects with multimodal data,'' \emph{arXiv preprint arXiv:2402.01785}, 2024.

\bibitem{liang2022high}
P.~P. Liang, Y.~Lyu, X.~Fan, J.~Tsaw, Y.~Liu, S.~Mo, D.~Yogatama, L.-P. Morency, and R.~Salakhutdinov, ``High-modality multimodal transformer: Quantifying modality \& interaction heterogeneity for high-modality representation learning,'' \emph{arXiv preprint arXiv:2203.01311}, 2022.

\bibitem{huang2024survey}
J.~Huang and J.~Zhang, ``A survey on evaluation of multimodal large language models,'' \emph{arXiv preprint arXiv:2408.15769}, 2024.

\bibitem{tasdizen2024vista}
T.~Tasdizen \emph{et~al.}, ``Vista: A visual and textual attention dataset for interpreting multimodal models,'' \emph{arXiv preprint arXiv:2410.04609}, 2024.

\bibitem{cai2023benchlmm}
R.~Cai, Z.~Song, D.~Guan, Z.~Chen, X.~Luo, C.~Yi, and A.~Kot, ``Benchlmm: Benchmarking cross-style visual capability of large multimodal models,'' \emph{arXiv preprint arXiv:2312.02896}, 2023.

\bibitem{dang2024exploring}
Y.~Dang, M.~Gao, Y.~Yan, X.~Zou, Y.~Gu, A.~Liu, and X.~Hu, ``Exploring response uncertainty in mllms: An empirical evaluation under misleading scenarios,'' \emph{arXiv preprint arXiv:2411.02708}, 2024.

\bibitem{zhang2024benchmarking}
Y.~Zhang, Y.~Huang, Y.~Sun, C.~Liu, Z.~Zhao, Z.~Fang, Y.~Wang, H.~Chen, X.~Yang, X.~Wei \emph{et~al.}, ``Benchmarking trustworthiness of multimodal large language models: A comprehensive study,'' \emph{arXiv preprint arXiv:2406.07057}, 2024.

\bibitem{hu2023tifa}
Y.~Hu, B.~Liu, J.~Kasai, Y.~Wang, M.~Ostendorf, R.~Krishna, and N.~A. Smith, ``Tifa: Accurate and interpretable text-to-image faithfulness evaluation with question answering,'' in \emph{Proceedings of the IEEE/CVF International Conference on Computer Vision}, 2023, pp. 20\,406--20\,417.

\bibitem{verma2024cross}
G.~Verma, M.~Choi, K.~Sharma, J.~Watson-Daniels, S.~Oh, and S.~Kumar, ``Cross-modal projection in multimodal llms doesn’t really project visual attributes to textual space,'' in \emph{Proceedings of the 62nd Annual Meeting of the Association for Computational Linguistics (Volume 2: Short Papers)}, 2024, pp. 657--664.

\bibitem{tiong2024we}
A.~M.~H. Tiong, J.~Zhao, B.~Li, J.~Li, S.~C. Hoi, and C.~Xiong, ``What are we measuring when we evaluate large vision-language models? an analysis of latent factors and biases,'' \emph{arXiv preprint arXiv:2404.02415}, 2024.

\bibitem{alipour2020study}
K.~Alipour, J.~P. Schulze, Y.~Yao, A.~Ziskind, and G.~Burachas, ``A study on multimodal and interactive explanations for visual question answering,'' \emph{arXiv preprint arXiv:2003.00431}, 2020.

\bibitem{liu2020deep}
Y.~Liu and T.~Tuytelaars, ``A deep multi-modal explanation model for zero-shot learning,'' \emph{IEEE Transactions on Image Processing}, vol.~29, pp. 4788--4803, 2020.

\bibitem{lu2022learn}
P.~Lu, S.~Mishra, T.~Xia, L.~Qiu, K.-W. Chang, S.-C. Zhu, O.~Tafjord, P.~Clark, and A.~Kalyan, ``Learn to explain: Multimodal reasoning via thought chains for science question answering,'' in \emph{The 36th Conference on Neural Information Processing Systems (NeurIPS)}, 2022.

\bibitem{shaham2024multimodal}
T.~R. Shaham, S.~Schwettmann, F.~Wang, A.~Rajaram, E.~Hernandez, J.~Andreas, and A.~Torralba, ``A multimodal automated interpretability agent,'' in \emph{Forty-first International Conference on Machine Learning}, 2024.

\bibitem{cuadra2024digital}
A.~Cuadra, J.~Breuch, S.~Estrada, D.~Ihim, I.~Hung, D.~Askaryar, M.~Hassanien, K.~L. Fessele, and J.~A. Landay, ``Digital forms for all: A holistic multimodal large language model agent for health data entry,'' \emph{Proceedings of the ACM on Interactive, Mobile, Wearable and Ubiquitous Technologies}, vol.~8, no.~2, pp. 1--39, 2024.

\bibitem{liu2022make}
Y.~Liu, Z.~Yuan, H.~Mao, Z.~Liang, W.~Yang, Y.~Qiu, T.~Cheng, X.~Li, H.~Xu, and K.~Gao, ``Make acoustic and visual cues matter: Ch-sims v2. 0 dataset and av-mixup consistent module,'' in \emph{Proceedings of the 2022 international conference on multimodal interaction}, 2022, pp. 247--258.

\bibitem{kanehira2019multimodal}
A.~Kanehira, K.~Takemoto, S.~Inayoshi, and T.~Harada, ``Multimodal explanations by predicting counterfactuality in videos,'' in \emph{Proceedings of the IEEE/CVF Conference on Computer Vision and Pattern Recognition}, 2019, pp. 8594--8602.

\bibitem{zang2023discovering}
C.~Zang, H.~Wang, M.~Pei, and W.~Liang, ``Discovering the real association: Multimodal causal reasoning in video question answering,'' in \emph{Proceedings of the IEEE/CVF Conference on Computer Vision and Pattern Recognition}, 2023, pp. 19\,027--19\,036.

\bibitem{ko2023large}
D.~Ko, J.~S. Lee, W.~Kang, B.~Roh, and H.~J. Kim, ``Large language models are temporal and causal reasoners for video question answering,'' \emph{arXiv preprint arXiv:2310.15747}, 2023.

\bibitem{zhang2024holmes}
H.~Zhang, X.~Xu, X.~Wang, J.~Zuo, C.~Han, X.~Huang, C.~Gao, Y.~Wang, and N.~Sang, ``Holmes-vad: Towards unbiased and explainable video anomaly detection via multi-modal llm,'' \emph{arXiv preprint arXiv:2406.12235}, 2024.

\bibitem{sanders2024tv}
K.~Sanders, N.~Weir, and B.~Van~Durme, ``Tv-trees: Multimodal entailment trees for neuro-symbolic video reasoning,'' \emph{arXiv preprint arXiv:2402.19467}, 2024.

\bibitem{hu2019multi}
Y.~Hu, W.~Zhan, L.~Sun, and M.~Tomizuka, ``Multi-modal probabilistic prediction of interactive behavior via an interpretable model,'' in \emph{2019 IEEE Intelligent Vehicles Symposium (IV)}.\hskip 1em plus 0.5em minus 0.4em\relax IEEE, 2019, pp. 557--563.

\bibitem{xu2024drivegpt4}
Z.~Xu, Y.~Zhang, E.~Xie, Z.~Zhao, Y.~Guo, K.-Y.~K. Wong, Z.~Li, and H.~Zhao, ``Drivegpt4: Interpretable end-to-end autonomous driving via large language model,'' \emph{IEEE Robotics and Automation Letters}, 2024.

\bibitem{liu2022group}
P.~Liu, K.~Li, and H.~Meng, ``Group gated fusion on attention-based bidirectional alignment for multimodal emotion recognition,'' \emph{arXiv preprint arXiv:2201.06309}, 2022.

\bibitem{zadeh2018multimodal}
A.~B. Zadeh, P.~P. Liang, S.~Poria, E.~Cambria, and L.-P. Morency, ``Multimodal language analysis in the wild: Cmu-mosei dataset and interpretable dynamic fusion graph,'' in \emph{Proceedings of the 56th Annual Meeting of the Association for Computational Linguistics (Volume 1: Long Papers)}, 2018, pp. 2236--2246.

\bibitem{sotirou2024musiclime}
T.~Sotirou, V.~Lyberatos, O.~M. Mastromichalakis, and G.~Stamou, ``Musiclime: Explainable multimodal music understanding,'' \emph{arXiv preprint arXiv:2409.10496}, 2024.

\bibitem{amara2024enhancing}
J.~Amara, B.~K{\"o}nig-Ries, and S.~Samuel, ``Enhancing explainability in multimodal large language models using ontological context,'' \emph{arXiv preprint arXiv:2409.18753}, 2024.

\bibitem{gandelsman2023interpreting}
Y.~Gandelsman, A.~A. Efros, and J.~Steinhardt, ``Interpreting clip's image representation via text-based decomposition,'' \emph{arXiv preprint arXiv:2310.05916}, 2023.

\bibitem{neo2024towards}
C.~Neo, L.~Ong, P.~Torr, M.~Geva, D.~Krueger, and F.~Barez, ``Towards interpreting visual information processing in vision-language models,'' \emph{arXiv preprint arXiv:2410.07149}, 2024.

\bibitem{zhang2024redundancy}
X.~Zhang, C.~Shen, X.~Yuan, S.~Yan, L.~Xie, W.~Wang, C.~Gu, H.~Tang, and J.~Ye, ``From redundancy to relevance: Enhancing explainability in multimodal large language models,'' \emph{arXiv preprint arXiv:2406.06579}, 2024.

\bibitem{chen2024image}
L.~Chen, H.~Zhao, T.~Liu, S.~Bai, J.~Lin, C.~Zhou, and B.~Chang, ``An image is worth 1/2 tokens after layer 2: Plug-and-play inference acceleration for large vision-language models,'' \emph{arXiv preprint arXiv:2403.06764}, 2024.

\bibitem{wang2023visual}
Y.~Wang, T.~G. Rudner, and A.~G. Wilson, ``Visual explanations of image-text representations via multi-modal information bottleneck attribution,'' \emph{Advances in Neural Information Processing Systems}, vol.~36, pp. 16\,009--16\,027, 2023.

\bibitem{yao2024deco}
L.~Yao, L.~Li, S.~Ren, L.~Wang, Y.~Liu, X.~Sun, and L.~Hou, ``Deco: Decoupling token compression from semantic abstraction in multimodal large language models,'' \emph{arXiv preprint arXiv:2405.20985}, 2024.

\bibitem{bi2023vl}
J.~Bi, D.~Cheng, P.~Yao, B.~Pang, Y.~Zhan, C.~Yang, Y.~Wang, H.~Sun, W.~Deng, and Q.~Zhang, ``Vl-match: Enhancing vision-language pretraining with token-level and instance-level matching,'' in \emph{Proceedings of the IEEE/CVF International Conference on Computer Vision}, 2023, pp. 2584--2593.

\bibitem{zhao2024first}
Q.~Zhao, M.~Xu, K.~Gupta, A.~Asthana, L.~Zheng, and S.~Gould, ``The first to know: How token distributions reveal hidden knowledge in large vision-language models?'' \emph{arXiv preprint arXiv:2403.09037}, 2024.

\bibitem{li2024unified}
Y.~Li, Y.~Wang, Y.~Fu, D.~Ru, Z.~Zhang, and T.~He, ``Unified lexical representation for interpretable visual-language alignment,'' \emph{arXiv preprint arXiv:2407.17827}, 2024.

\bibitem{dai2022plausible_hjh3}
W.~Dai, Z.~Liu, Z.~Ji, D.~Su, and P.~Fung, ``Plausible may not be faithful: Probing object hallucination in vision-language pre-training,'' \emph{arXiv preprint arXiv:2210.07688}, 2022.

\bibitem{huang2024opera}
Q.~Huang, X.~Dong, P.~Zhang, B.~Wang, C.~He, J.~Wang, D.~Lin, W.~Zhang, and N.~Yu, ``Opera: Alleviating hallucination in multi-modal large language models via over-trust penalty and retrospection-allocation,'' in \emph{Proceedings of the IEEE/CVF Conference on Computer Vision and Pattern Recognition}, 2024, pp. 13\,418--13\,427.

\bibitem{shi2024eagle}
M.~Shi, F.~Liu, S.~Wang, S.~Liao, S.~Radhakrishnan, D.-A. Huang, H.~Yin, K.~Sapra, Y.~Yacoob, H.~Shi \emph{et~al.}, ``Eagle: Exploring the design space for multimodal llms with mixture of encoders,'' \emph{arXiv preprint arXiv:2408.15998}, 2024.

\bibitem{hendricks2021probing}
L.~A. Hendricks and A.~Nematzadeh, ``Probing image-language transformers for verb understanding,'' \emph{arXiv preprint arXiv:2106.09141}, 2021.

\bibitem{moayeri2023text}
M.~Moayeri, K.~Rezaei, M.~Sanjabi, and S.~Feizi, ``Text-to-concept (and back) via cross-model alignment,'' in \emph{International Conference on Machine Learning}.\hskip 1em plus 0.5em minus 0.4em\relax PMLR, 2023, pp. 25\,037--25\,060.

\bibitem{derby2018using}
S.~Derby, P.~Miller, B.~Murphy, and B.~Devereux, ``Using sparse semantic embeddings learned from multimodal text and image data to model human conceptual knowledge,'' \emph{arXiv preprint arXiv:1809.02534}, 2018.

\bibitem{chen2023stair}
C.~Chen, B.~Zhang, L.~Cao, J.~Shen, T.~Gunter, A.~M. Jose, A.~Toshev, J.~Shlens, R.~Pang, and Y.~Yang, ``Stair: learning sparse text and image representation in grounded tokens,'' \emph{arXiv preprint arXiv:2301.13081}, 2023.

\bibitem{dominici2023sharcs}
G.~Dominici, P.~Barbiero, L.~C. Magister, P.~Li{\`o}, and N.~Simidjievski, ``Sharcs: Shared concept space for explainable multimodal learning,'' \emph{arXiv preprint arXiv:2307.00316}, 2023.

\bibitem{bhalla2024interpreting}
U.~Bhalla, A.~Oesterling, S.~Srinivas, F.~P. Calmon, and H.~Lakkaraju, ``Interpreting clip with sparse linear concept embeddings (splice),'' \emph{arXiv preprint arXiv:2402.10376}, 2024.

\bibitem{frank2021vision}
S.~Frank, E.~Bugliarello, and D.~Elliott, ``Vision-and-language or vision-for-language? on cross-modal influence in multimodal transformers,'' \emph{arXiv preprint arXiv:2109.04448}, 2021.

\bibitem{wangfreebind2024}
Z.~Wang, Z.~Zhang, X.~Cheng, R.~Huang, L.~Liu, Z.~Ye, H.~Huang, Y.~Zhao, T.~Jin, P.~Gao \emph{et~al.}, ``Freebind: Free lunch in unified multimodal space via knowledge fusion,'' in \emph{Forty-first International Conference on Machine Learning}, 2024.

\bibitem{parekh2024concept}
J.~Parekh, P.~Khayatan, M.~Shukor, A.~Newson, and M.~Cord, ``A concept-based explainability framework for large multimodal models,'' \emph{arXiv preprint arXiv:2406.08074}, 2024.

\bibitem{evirgen2024text}
N.~Evirgen, R.~Wang, and X.~Chen, ``From text to pixels: Enhancing user understanding through text-to-image model explanations,'' in \emph{Proceedings of the 29th International Conference on Intelligent User Interfaces}, 2024, pp. 74--87.

\bibitem{salin2022vision}
E.~Salin, B.~Farah, S.~Ayache, and B.~Favre, ``Are vision-language transformers learning multimodal representations? a probing perspective,'' in \emph{Proceedings of the AAAI Conference on Artificial Intelligence}, vol.~36, no.~10, 2022, pp. 11\,248--11\,257.

\bibitem{ramesh2022investigation}
K.~Ramesh and Y.~S. Koh, ``Investigation of explainability techniques for multimodal transformers,'' in \emph{Australasian Conference on Data Mining}.\hskip 1em plus 0.5em minus 0.4em\relax Springer, 2022, pp. 90--98.

\bibitem{goh2021multimodal}
G.~Goh, N.~Cammarata, C.~Voss, S.~Carter, M.~Petrov, L.~Schubert, A.~Radford, and C.~Olah, ``Multimodal neurons in artificial neural networks,'' \emph{Distill}, vol.~6, no.~3, p. e30, 2021.

\bibitem{gandelsman2024interpreting}
Y.~Gandelsman, A.~A. Efros, and J.~Steinhardt, ``Interpreting the second-order effects of neurons in clip,'' \emph{arXiv preprint arXiv:2406.04341}, 2024.

\bibitem{pan2023finding}
H.~Pan, Y.~Cao, X.~Wang, and X.~Yang, ``Finding and editing multi-modal neurons in pre-trained transformer,'' \emph{arXiv preprint arXiv:2311.07470}, 2023.

\bibitem{huo2024mmneuron}
J.~Huo, Y.~Yan, B.~Hu, Y.~Yue, and X.~Hu, ``Mmneuron: Discovering neuron-level domain-specific interpretation in multimodal large language model,'' \emph{arXiv preprint arXiv:2406.11193}, 2024.

\bibitem{Huang2024MINERMT}
K.~Huang, J.~Huo, Y.~Yan, K.~Wang, Y.~Yue, and X.~Hu, ``Miner: Mining the underlying pattern of modality-specific neurons in multimodal large language models,'' \emph{arXiv preprint arXiv:2410.04819}, 2024.

\bibitem{cao2020behind}
J.~Cao, Z.~Gan, Y.~Cheng, L.~Yu, Y.-C. Chen, and J.~Liu, ``Behind the scene: Revealing the secrets of pre-trained vision-and-language models,'' in \emph{Computer Vision--ECCV 2020: 16th European Conference, Glasgow, UK, August 23--28, 2020, Proceedings, Part VI 16}.\hskip 1em plus 0.5em minus 0.4em\relax Springer, 2020, pp. 565--580.

\bibitem{quantmeyer2024and}
V.~Quantmeyer, P.~Mosteiro, and A.~Gatt, ``How and where does clip process negation?'' \emph{arXiv preprint arXiv:2407.10488}, 2024.

\bibitem{xu2023bridging}
Z.~Xu, Z.~Chen, Y.~Zhang, Y.~Song, X.~Wan, and G.~Li, ``Bridging vision and language encoders: Parameter-efficient tuning for referring image segmentation,'' in \emph{Proceedings of the IEEE/CVF International Conference on Computer Vision}, 2023, pp. 17\,503--17\,512.

\bibitem{tao2024probing}
M.~Tao, Q.~Huang, K.~Xu, L.~Chen, Y.~Feng, and D.~Zhao, ``Probing multimodal large language models for global and local semantic representation,'' \emph{arXiv preprint arXiv:2402.17304}, 2024.

\bibitem{nguyen2019multi}
D.-K. Nguyen and T.~Okatani, ``Multi-task learning of hierarchical vision-language representation,'' in \emph{Proceedings of the IEEE/CVF Conference on Computer Vision and Pattern Recognition}, 2019, pp. 10\,492--10\,501.

\bibitem{lyu2022dime}
Y.~Lyu, P.~P. Liang, Z.~Deng, R.~Salakhutdinov, and L.-P. Morency, ``Dime: Fine-grained interpretations of multimodal models via disentangled local explanations,'' in \emph{Proceedings of the 2022 AAAI/ACM Conference on AI, Ethics, and Society}, 2022, pp. 455--467.

\bibitem{goyal2016towards}
Y.~Goyal, A.~Mohapatra, D.~Parikh, and D.~Batra, ``Towards transparent ai systems: Interpreting visual question answering models,'' \emph{arXiv preprint arXiv:1608.08974}, 2016.

\bibitem{wu2018faithful}
J.~Wu and R.~J. Mooney, ``Faithful multimodal explanation for visual question answering,'' \emph{arXiv preprint arXiv:1809.02805}, 2018.

\bibitem{natarajan2024vale}
P.~Natarajan and A.~Nambiar, ``Vale: A multimodal visual and language explanation framework for image classifiers using explainable ai and language models,'' \emph{arXiv preprint arXiv:2408.12808}, 2024.

\bibitem{stan2024lvlm}
G.~B.~M. Stan, R.~Y. Rohekar, Y.~Gurwicz, M.~L. Olson, A.~Bhiwandiwalla, E.~Aflalo, C.~Wu, N.~Duan, S.-Y. Tseng, and V.~Lal, ``Lvlm-intrepret: An interpretability tool for large vision-language models,'' \emph{arXiv preprint arXiv:2404.03118}, 2024.

\bibitem{lee2023diffusion}
S.~Lee, B.~Hoover, H.~Strobelt, Z.~J. Wang, S.~Peng, A.~Wright, K.~Li, H.~Park, H.~Yang, and D.~H. Chau, ``Diffusion explainer: Visual explanation for text-to-image stable diffusion,'' \emph{arXiv preprint arXiv:2305.03509}, 2023.

\bibitem{yang2024law}
S.~Yang, B.~Zhai, Q.~You, J.~Yuan, H.~Yang, and C.~Xu, ``Law of vision representation in mllms,'' \emph{arXiv preprint arXiv:2408.16357}, 2024.

\bibitem{wan2020nbdt}
A.~Wan, L.~Dunlap, D.~Ho, J.~Yin, S.~Lee, H.~Jin, S.~Petryk, S.~A. Bargal, and J.~E. Gonzalez, ``Nbdt: Neural-backed decision trees,'' \emph{arXiv preprint arXiv:2004.00221}, 2020.

\bibitem{yang2023language}
Y.~Yang, A.~Panagopoulou, S.~Zhou, D.~Jin, C.~Callison-Burch, and M.~Yatskar, ``Language in a bottle: Language model guided concept bottlenecks for interpretable image classification,'' in \emph{Proceedings of the IEEE/CVF Conference on Computer Vision and Pattern Recognition}, 2023, pp. 19\,187--19\,197.

\bibitem{guo2024trace}
Y.~Guo, J.~Liu, M.~Li, X.~Tang, Q.~Liu, and X.~Chen, ``Trace: Temporal grounding video llm via causal event modeling,'' \emph{arXiv preprint arXiv:2410.05643}, 2024.

\bibitem{li2024multimodal}
S.~Li, F.~Xue, K.~Liu, D.~Guo, and R.~Hong, ``Multimodal graph causal embedding for multimedia-based recommendation,'' \emph{IEEE Transactions on Knowledge and Data Engineering}, 2024.

\bibitem{swamy2024multimodn}
V.~Swamy, M.~Satayeva, J.~Frej, T.~Bossy, T.~Vogels, M.~Jaggi, T.~K{\"a}ser, and M.-A. Hartley, ``Multimodn—multimodal, multi-task, interpretable modular networks,'' \emph{Advances in Neural Information Processing Systems}, vol.~36, 2024.

\bibitem{shen2023scaling}
S.~Shen, Z.~Yao, C.~Li, T.~Darrell, K.~Keutzer, and Y.~He, ``Scaling vision-language models with sparse mixture of experts,'' \emph{arXiv preprint arXiv:2303.07226}, 2023.

\bibitem{wen2023imkga}
Y.~Wen, B.~Luo, and Y.~Zhao, ``Imkga-sm: Interpretable multimodal knowledge graph answer prediction via sequence modeling,'' \emph{arXiv preprint arXiv:2301.02445}, 2023.

\bibitem{zang2024pre}
Y.~Zang, T.~Yun, H.~Tan, T.~Bui, and C.~Sun, ``Pre-trained vision-language models learn discoverable visual concepts,'' \emph{arXiv preprint arXiv:2404.12652}, 2024.

\bibitem{zhang2024disttrain}
Z.~Zhang, Y.~Zhong, R.~Ming, H.~Hu, J.~Sun, Z.~Ge, Y.~Zhu, and X.~Jin, ``Disttrain: Addressing model and data heterogeneity with disaggregated training for multimodal large language models,'' \emph{arXiv e-prints}, pp. arXiv--2408, 2024.

\bibitem{sun2023aligning}
Z.~Sun, S.~Shen, S.~Cao, H.~Liu, C.~Li, Y.~Shen, C.~Gan, L.-Y. Gui, Y.-X. Wang, Y.~Yang \emph{et~al.}, ``Aligning large multimodal models with factually augmented rlhf,'' \emph{arXiv preprint arXiv:2309.14525}, 2023.

\bibitem{yan2024vigor}
S.~Yan, M.~Bai, W.~Chen, X.~Zhou, Q.~Huang, and L.~E. Li, ``Vigor: Improving visual grounding of large vision language models with fine-grained reward modeling,'' \emph{arXiv preprint arXiv:2402.06118}, 2024.

\bibitem{yu2024rlhf}
T.~Yu, Y.~Yao, H.~Zhang, T.~He, Y.~Han, G.~Cui, J.~Hu, Z.~Liu, H.-T. Zheng, M.~Sun \emph{et~al.}, ``Rlhf-v: Towards trustworthy mllms via behavior alignment from fine-grained correctional human feedback,'' in \emph{Proceedings of the IEEE/CVF Conference on Computer Vision and Pattern Recognition}, 2024, pp. 13\,807--13\,816.

\bibitem{tsai2020multimodal}
Y.-H.~H. Tsai, M.~Q. Ma, M.~Yang, R.~Salakhutdinov, and L.-P. Morency, ``Multimodal routing: Improving local and global interpretability of multimodal language analysis,'' in \emph{Proceedings of the Conference on Empirical Methods in Natural Language Processing. Conference on Empirical Methods in Natural Language Processing}, vol. 2020.\hskip 1em plus 0.5em minus 0.4em\relax NIH Public Access, 2020, p. 1823.

\bibitem{mallick2024ifi}
R.~Mallick, J.~Benois-Pineau, and A.~Zemmari, ``Ifi: Interpreting for improving: A multimodal transformer with an interpretability technique for recognition of risk events,'' in \emph{International Conference on Multimedia Modeling}.\hskip 1em plus 0.5em minus 0.4em\relax Springer, 2024, pp. 117--131.

\bibitem{zhao2023beyond}
Z.~Zhao, B.~Wang, L.~Ouyang, X.~Dong, J.~Wang, and C.~He, ``Beyond hallucinations: Enhancing lvlms through hallucination-aware direct preference optimization,'' \emph{arXiv preprint arXiv:2311.16839}, 2023.

\bibitem{yu2024rlaif}
T.~Yu, H.~Zhang, Y.~Yao, Y.~Dang, D.~Chen, X.~Lu, G.~Cui, T.~He, Z.~Liu, T.-S. Chua \emph{et~al.}, ``Rlaif-v: Aligning mllms through open-source ai feedback for super gpt-4v trustworthiness,'' \emph{arXiv preprint arXiv:2405.17220}, 2024.

\bibitem{zhou2024aligning}
Y.~Zhou, C.~Cui, R.~Rafailov, C.~Finn, and H.~Yao, ``Aligning modalities in vision large language models via preference fine-tuning,'' \emph{arXiv preprint arXiv:2402.11411}, 2024.

\bibitem{liu2024survey}
H.~Liu, W.~Xue, Y.~Chen, D.~Chen, X.~Zhao, K.~Wang, L.~Hou, R.~Li, and W.~Peng, ``A survey on hallucination in large vision-language models,'' \emph{arXiv preprint arXiv:2402.00253}, 2024.

\bibitem{bai2024hallucination}
Z.~Bai, P.~Wang, T.~Xiao, T.~He, Z.~Han, Z.~Zhang, and M.~Z. Shou, ``Hallucination of multimodal large language models: A survey,'' \emph{arXiv preprint arXiv:2404.18930}, 2024.

\bibitem{zhang2023multimodal}
Z.~Zhang, A.~Zhang, M.~Li, H.~Zhao, G.~Karypis, and A.~Smola, ``Multimodal chain-of-thought reasoning in language models,'' \emph{arXiv preprint arXiv:2302.00923}, 2023.

\bibitem{ge2023chain}
J.~Ge, H.~Luo, S.~Qian, Y.~Gan, J.~Fu, and S.~Zhang, ``Chain of thought prompt tuning in vision language models,'' \emph{arXiv preprint arXiv:2304.07919}, 2023.

\bibitem{yao2023thinking}
F.~Yao, C.~Tian, J.~Liu, Z.~Zhang, Q.~Liu, L.~Jin, S.~Li, X.~Li, and X.~Sun, ``Thinking like an expert: Multimodal hypergraph-of-thought (hot) reasoning to boost foundation modals,'' \emph{arXiv preprint arXiv:2308.06207}, 2023.

\bibitem{zintgraf2017visualizing}
L.~M. Zintgraf, T.~S. Cohen, T.~Adel, and M.~Welling, ``Visualizing deep neural network decisions: Prediction difference analysis,'' \emph{arXiv preprint arXiv:1702.04595}, 2017.

\bibitem{szegedy2013intriguing}
C.~Szegedy, ``Intriguing properties of neural networks,'' \emph{arXiv preprint arXiv:1312.6199}, 2013.

\bibitem{petsiuk2018rise}
V.~Petsiuk, ``Rise: Randomized input sampling for explanation of black-box models,'' \emph{arXiv preprint arXiv:1806.07421}, 2018.

\bibitem{fong2019understanding}
R.~Fong, M.~Patrick, and A.~Vedaldi, ``Understanding deep networks via extremal perturbations and smooth masks,'' in \emph{Proceedings of the IEEE/CVF international conference on computer vision}, 2019, pp. 2950--2958.

\bibitem{kanehira2019learning}
A.~Kanehira and T.~Harada, ``Learning to explain with complemental examples,'' in \emph{Proceedings of the IEEE/CVF conference on computer vision and pattern recognition}, 2019, pp. 8603--8611.

\bibitem{fel2024holistic}
T.~Fel, V.~Boutin, L.~B{\'e}thune, R.~Cad{\`e}ne, M.~Moayeri, L.~And{\'e}ol, M.~Chalvidal, and T.~Serre, ``A holistic approach to unifying automatic concept extraction and concept importance estimation,'' \emph{Advances in Neural Information Processing Systems}, vol.~36, 2024.

\bibitem{morioka2023connectivity}
H.~Morioka and A.~Hyvarinen, ``Connectivity-contrastive learning: Combining causal discovery and representation learning for multimodal data,'' in \emph{International conference on artificial intelligence and statistics}.\hskip 1em plus 0.5em minus 0.4em\relax PMLR, 2023, pp. 3399--3426.

\bibitem{yuan2024diffusion}
X.~Yuan and Y.~Qiao, ``Diffusion-ts: Interpretable diffusion for general time series generation,'' \emph{arXiv preprint arXiv:2403.01742}, 2024.

\bibitem{luo2022understanding}
C.~Luo, ``Understanding diffusion models: A unified perspective,'' \emph{arXiv preprint arXiv:2208.11970}, 2022.

\bibitem{yang2023diffusion}
L.~Yang, Z.~Zhang, Y.~Song, S.~Hong, R.~Xu, Y.~Zhao, W.~Zhang, B.~Cui, and M.-H. Yang, ``Diffusion models: A comprehensive survey of methods and applications,'' \emph{ACM Computing Surveys}, vol.~56, no.~4, pp. 1--39, 2023.

\bibitem{jeanneret2022diffusion}
G.~Jeanneret, L.~Simon, and F.~Jurie, ``Diffusion models for counterfactual explanations,'' in \emph{Proceedings of the Asian Conference on Computer Vision}, 2022, pp. 858--876.

\bibitem{tang2022daam}
R.~Tang, L.~Liu, A.~Pandey, Z.~Jiang, G.~Yang, K.~Kumar, P.~Stenetorp, J.~Lin, and F.~Ture, ``What the daam: Interpreting stable diffusion using cross attention,'' \emph{arXiv preprint arXiv:2210.04885}, 2022.

\bibitem{kong2023interpretable}
X.~Kong, O.~Liu, H.~Li, D.~Yogatama, and G.~V. Steeg, ``Interpretable diffusion via information decomposition,'' \emph{arXiv preprint arXiv:2310.07972}, 2023.

\bibitem{mao2023coco}
X.~Mao, Y.~Chen, Y.~Zhu, D.~Chen, H.~Su, R.~Zhang, and H.~Xue, ``Coco-o: A benchmark for object detectors under natural distribution shifts,'' in \emph{Proceedings of the IEEE/CVF International Conference on Computer Vision}, 2023, pp. 6339--6350.

\bibitem{madhyastha2018defoiling}
P.~Madhyastha, J.~Wang, and L.~Specia, ``Defoiling foiled image captions,'' \emph{arXiv preprint arXiv:1805.06549}, 2018.

\bibitem{hewitt2019designing}
J.~Hewitt and P.~Liang, ``Designing and interpreting probes with control tasks,'' \emph{arXiv preprint arXiv:1909.03368}, 2019.

\bibitem{schwettmann2024find}
S.~Schwettmann, T.~Shaham, J.~Materzynska, N.~Chowdhury, S.~Li, J.~Andreas, D.~Bau, and A.~Torralba, ``Find: A function description benchmark for evaluating interpretability methods,'' \emph{Advances in Neural Information Processing Systems}, vol.~36, 2024.

\bibitem{tang2023explainable}
X.~Tang, J.~Zhang, Y.~He, X.~Zhang, Z.~Lin, S.~Partarrieu, E.~B. Hanna, Z.~Ren, H.~Shen, Y.~Yang \emph{et~al.}, ``Explainable multi-task learning for multi-modality biological data analysis,'' \emph{Nature communications}, vol.~14, no.~1, p. 2546, 2023.

\bibitem{rawls2021integrated}
E.~Rawls, E.~Kummerfeld, and A.~Zilverstand, ``An integrated multimodal model of alcohol use disorder generated by data-driven causal discovery analysis,'' \emph{Communications biology}, vol.~4, no.~1, p. 435, 2021.

\bibitem{chu2023qwen}
Y.~Chu, J.~Xu, X.~Zhou, Q.~Yang, S.~Zhang, Z.~Yan, C.~Zhou, and J.~Zhou, ``Qwen-audio: Advancing universal audio understanding via unified large-scale audio-language models,'' \emph{arXiv preprint arXiv:2311.07919}, 2023.

\bibitem{sundar2024multimodal}
A.~S. Sundar, C.-H.~H. Yang, D.~M. Chan, S.~Ghosh, V.~Ravichandran, and P.~S. Nidadavolu, ``Multimodal attention merging for improved speech recognition and audio event classification,'' in \emph{2024 IEEE International Conference on Acoustics, Speech, and Signal Processing Workshops (ICASSPW)}.\hskip 1em plus 0.5em minus 0.4em\relax IEEE, 2024, pp. 655--659.

\bibitem{jalal2020empirical}
M.~A. Jalal, R.~Milner, and T.~Hain, ``Empirical interpretation of speech emotion perception with attention based model for speech emotion recognition,'' in \emph{Proceedings of Interspeech 2020}.\hskip 1em plus 0.5em minus 0.4em\relax International Speech Communication Association (ISCA), 2020, pp. 4113--4117.

\bibitem{won2019toward}
M.~Won, S.~Chun, and X.~Serra, ``Toward interpretable music tagging with self-attention,'' \emph{arXiv preprint arXiv:1906.04972}, 2019.

\bibitem{lyberatos2024perceptual}
V.~Lyberatos, S.~Kantarelis, E.~Dervakos, and G.~Stamou, ``Perceptual musical features for interpretable audio tagging,'' in \emph{2024 IEEE International Conference on Acoustics, Speech, and Signal Processing Workshops (ICASSPW)}.\hskip 1em plus 0.5em minus 0.4em\relax IEEE, 2024, pp. 878--882.

\bibitem{alonso2024leveraging}
P.~Alonso-Jim{\'e}nez, L.~Pepino, R.~Batlle-Roca, P.~Zinemanas, D.~Bogdanov, X.~Serra, and M.~Rocamora, ``Leveraging pre-trained autoencoders for interpretable prototype learning of music audio,'' \emph{arXiv preprint arXiv:2402.09318}, 2024.

\bibitem{foscarin2022concept}
F.~Foscarin, K.~Hoedt, V.~Praher, A.~Flexer, and G.~Widmer, ``Concept-based techniques for" musicologist-friendly" explanations in a deep music classifier,'' \emph{arXiv preprint arXiv:2208.12485}, 2022.

\bibitem{wei2024dopra}
J.~Wei and X.~Zhang, ``Dopra: Decoding over-accumulation penalization and re-allocation in specific weighting layer,'' \emph{arXiv preprint arXiv:2407.15130}, 2024.

\bibitem{park2024explaining}
J.-H. Park, Y.-J. Ju, and S.-W. Lee, ``Explaining generative diffusion models via visual analysis for interpretable decision-making process,'' \emph{Expert Systems with Applications}, vol. 248, p. 123231, 2024.

\bibitem{prasad2024tree}
V.~Prasad, H.~van Gorp, C.~Humer, A.~Vilanova, and N.~Pezzotti, ``The tree of diffusion life: Evolutionary embeddings to understand the generation process of diffusion models,'' \emph{arXiv preprint arXiv:2406.17462}, 2024.

\bibitem{wolfe2022contrastive}
R.~Wolfe and A.~Caliskan, ``Contrastive visual semantic pretraining magnifies the semantics of natural language representations,'' \emph{arXiv preprint arXiv:2203.07511}, 2022.

\bibitem{lindstrom2021probing}
A.~D. Lindstr{\"o}m, S.~Bensch, J.~Bj{\"o}rklund, and F.~Drewes, ``Probing multimodal embeddings for linguistic properties: the visual-semantic case,'' \emph{arXiv preprint arXiv:2102.11115}, 2021.

\bibitem{crabbe2023robust}
J.~Crabb{\'e}, P.~Rodr{\'\i}guez, V.~Shankar, L.~Zappella, and A.~Blaas, ``Robust multimodal models have outlier features and encode more concepts,'' \emph{arXiv preprint arXiv:2310.13040}, 2023.

\bibitem{kwon2022diffusion}
M.~Kwon, J.~Jeong, and Y.~Uh, ``Diffusion models already have a semantic latent space,'' \emph{arXiv preprint arXiv:2210.10960}, 2022.

\bibitem{zhou2018interpretable}
B.~Zhou, Y.~Sun, D.~Bau, and A.~Torralba, ``Interpretable basis decomposition for visual explanation,'' in \emph{Proceedings of the European Conference on Computer Vision (ECCV)}, 2018, pp. 119--134.

\bibitem{ma2023visualizing}
J.~Ma, Y.~Bai, B.~Zhong, W.~Zhang, T.~Yao, and T.~Mei, ``Visualizing and understanding patch interactions in vision transformer,'' \emph{IEEE Transactions on Neural Networks and Learning Systems}, 2023.

\bibitem{kim2022vit}
S.~Kim, J.~Nam, and B.~C. Ko, ``Vit-net: Interpretable vision transformers with neural tree decoder,'' in \emph{International conference on machine learning}.\hskip 1em plus 0.5em minus 0.4em\relax PMLR, 2022, pp. 11\,162--11\,172.

\bibitem{qiang2023interpretability}
Y.~Qiang, C.~Li, P.~Khanduri, and D.~Zhu, ``Interpretability-aware vision transformer,'' \emph{arXiv preprint arXiv:2309.08035}, 2023.

\bibitem{rao2021dynamicvit}
Y.~Rao, W.~Zhao, B.~Liu, J.~Lu, J.~Zhou, and C.-J. Hsieh, ``Dynamicvit: Efficient vision transformers with dynamic token sparsification,'' \emph{Advances in neural information processing systems}, vol.~34, pp. 13\,937--13\,949, 2021.

\bibitem{alain2016understanding}
G.~Alain, ``Understanding intermediate layers using linear classifier probes,'' \emph{arXiv preprint arXiv:1610.01644}, 2016.

\bibitem{kim2018interpretability}
B.~Kim, M.~Wattenberg, J.~Gilmer, C.~Cai, J.~Wexler, F.~Viegas \emph{et~al.}, ``Interpretability beyond feature attribution: Quantitative testing with concept activation vectors (tcav),'' in \emph{International conference on machine learning}.\hskip 1em plus 0.5em minus 0.4em\relax PMLR, 2018, pp. 2668--2677.

\bibitem{zhang2024better}
J.~Zhang, D.~Liu, C.~Qian, Z.~Gan, Y.~Liu, Y.~Qiao, and J.~Shao, ``The better angels of machine personality: How personality relates to llm safety,'' \emph{arXiv preprint arXiv:2407.12344}, 2024.

\bibitem{zhang2024reef}
J.~Zhang, D.~Liu, C.~Qian, L.~Zhang, Y.~Liu, Y.~Qiao, and J.~Shao, ``Reef: Representation encoding fingerprints for large language models,'' \emph{arXiv preprint arXiv:2410.14273}, 2024.

\bibitem{lee2024mechanistic}
A.~Lee, X.~Bai, I.~Pres, M.~Wattenberg, J.~K. Kummerfeld, and R.~Mihalcea, ``A mechanistic understanding of alignment algorithms: A case study on dpo and toxicity,'' \emph{arXiv preprint arXiv:2401.01967}, 2024.

\bibitem{sundararajan2017axiomatic}
M.~Sundararajan, A.~Taly, and Q.~Yan, ``Axiomatic attribution for deep networks,'' in \emph{International conference on machine learning}.\hskip 1em plus 0.5em minus 0.4em\relax PMLR, 2017, pp. 3319--3328.

\bibitem{bau2017network}
D.~Bau, B.~Zhou, A.~Khosla, A.~Oliva, and A.~Torralba, ``Network dissection: Quantifying interpretability of deep visual representations,'' in \emph{Proceedings of the IEEE conference on computer vision and pattern recognition}, 2017, pp. 6541--6549.

\bibitem{zhou2018interpreting}
B.~Zhou, D.~Bau, A.~Oliva, and A.~Torralba, ``Interpreting deep visual representations via network dissection,'' \emph{IEEE transactions on pattern analysis and machine intelligence}, vol.~41, no.~9, pp. 2131--2145, 2018.

\bibitem{zhu2021and}
X.~Zhu, C.~Xu, and D.~Tao, ``Where and what? examining interpretable disentangled representations,'' in \emph{Proceedings of the IEEE/CVF Conference on Computer Vision and Pattern Recognition}, 2021, pp. 5861--5870.

\bibitem{hennigen2020intrinsic}
L.~T. Hennigen, A.~Williams, and R.~Cotterell, ``Intrinsic probing through dimension selection,'' \emph{arXiv preprint arXiv:2010.02812}, 2020.

\bibitem{noutahi2019towards}
E.~Noutahi, D.~Beaini, J.~Horwood, S.~Gigu{\`e}re, and P.~Tossou, ``Towards interpretable sparse graph representation learning with laplacian pooling,'' \emph{arXiv preprint arXiv:1905.11577}, 2019.

\bibitem{bau2018gan}
D.~Bau, J.-Y. Zhu, H.~Strobelt, B.~Zhou, J.~B. Tenenbaum, W.~T. Freeman, and A.~Torralba, ``Gan dissection: Visualizing and understanding generative adversarial networks,'' \emph{arXiv preprint arXiv:1811.10597}, 2018.

\bibitem{bau2020understanding}
D.~Bau, J.-Y. Zhu, H.~Strobelt, A.~Lapedriza, B.~Zhou, and A.~Torralba, ``Understanding the role of individual units in a deep neural network,'' \emph{Proceedings of the National Academy of Sciences}, vol. 117, no.~48, pp. 30\,071--30\,078, 2020.

\bibitem{hernandez2021natural}
E.~Hernandez, S.~Schwettmann, D.~Bau, T.~Bagashvili, A.~Torralba, and J.~Andreas, ``Natural language descriptions of deep visual features,'' in \emph{International Conference on Learning Representations}, 2021.

\bibitem{dai2021knowledge}
D.~Dai, L.~Dong, Y.~Hao, Z.~Sui, B.~Chang, and F.~Wei, ``Knowledge neurons in pretrained transformers,'' \emph{arXiv preprint arXiv:2104.08696}, 2021.

\bibitem{oikarinen2022clip}
T.~Oikarinen and T.-W. Weng, ``Clip-dissect: Automatic description of neuron representations in deep vision networks,'' \emph{arXiv preprint arXiv:2204.10965}, 2022.

\bibitem{meng2022locating}
K.~Meng, D.~Bau, A.~Andonian, and Y.~Belinkov, ``Locating and editing factual associations in gpt,'' \emph{Advances in Neural Information Processing Systems}, vol.~35, pp. 17\,359--17\,372, 2022.

\bibitem{wang2022hint}
A.~Wang, W.-N. Lee, and X.~Qi, ``Hint: Hierarchical neuron concept explainer,'' in \emph{Proceedings of the IEEE/CVF Conference on Computer Vision and Pattern Recognition}, 2022, pp. 10\,254--10\,264.

\bibitem{dravid2023rosetta}
A.~Dravid, Y.~Gandelsman, A.~A. Efros, and A.~Shocher, ``Rosetta neurons: Mining the common units in a model zoo,'' in \emph{Proceedings of the IEEE/CVF International Conference on Computer Vision}, 2023, pp. 1934--1943.

\bibitem{bills2023language}
S.~Bills, N.~Cammarata, D.~Mossing, H.~Tillman, L.~Gao, G.~Goh, I.~Sutskever, J.~Leike, J.~Wu, and W.~Saunders, ``Language models can explain neurons in language models,'' \emph{URL https://openaipublic. blob. core. windows. net/neuron-explainer/paper/index. html.(Date accessed: 14.05. 2023)}, vol.~2, 2023.

\bibitem{Liu2023ConesCN}
\BIBentryALTinterwordspacing
Z.~Liu, R.~Feng, K.~Zhu, Y.~Zhang, K.~Zheng, Y.~Liu, D.~Zhao, J.~Zhou, and Y.~Cao, ``Cones: Concept neurons in diffusion models for customized generation,'' in \emph{International Conference on Machine Learning}, 2023. [Online]. Available: \url{https://api.semanticscholar.org/CorpusID:257427549}
\BIBentrySTDinterwordspacing

\bibitem{chen2024journey}
Y.~Chen, P.~Cao, Y.~Chen, K.~Liu, and J.~Zhao, ``Journey to the center of the knowledge neurons: Discoveries of language-independent knowledge neurons and degenerate knowledge neurons,'' in \emph{Proceedings of the AAAI Conference on Artificial Intelligence}, vol.~38, no.~16, 2024, pp. 17\,817--17\,825.

\bibitem{gao2024scaling}
L.~Gao, T.~D. la~Tour, H.~Tillman, G.~Goh, R.~Troll, A.~Radford, I.~Sutskever, J.~Leike, and J.~Wu, ``Scaling and evaluating sparse autoencoders,'' \emph{arXiv preprint arXiv:2406.04093}, 2024.

\bibitem{Hintersdorf2024FindingNL}
\BIBentryALTinterwordspacing
D.~Hintersdorf, L.~Struppek, K.~Kersting, A.~Dziedzic, and F.~Boenisch, ``Finding nemo: Localizing neurons responsible for memorization in diffusion models,'' \emph{ArXiv}, vol. abs/2406.02366, 2024. [Online]. Available: \url{https://api.semanticscholar.org/CorpusID:270226911}
\BIBentrySTDinterwordspacing

\bibitem{schubert2021high}
L.~Schubert, C.~Voss, N.~Cammarata, G.~Goh, and C.~Olah, ``High-low frequency detectors,'' \emph{Distill}, vol.~6, no.~1, pp. e00\,024--005, 2021.

\bibitem{cammarata2020curve}
N.~Cammarata, G.~Goh, S.~Carter, L.~Schubert, M.~Petrov, and C.~Olah, ``Curve detectors,'' \emph{Distill}, vol.~5, no.~6, pp. e00\,024--003, 2020.

\bibitem{olah2020zoom}
C.~Olah, N.~Cammarata, L.~Schubert, G.~Goh, M.~Petrov, and S.~Carter, ``Zoom in: An introduction to circuits,'' \emph{Distill}, vol.~5, no.~3, pp. e00\,024--001, 2020.

\bibitem{Mueller2022CausalAO}
\BIBentryALTinterwordspacing
A.~Mueller, Y.~Xia, and T.~Linzen, ``Causal analysis of syntactic agreement neurons in multilingual language models,'' in \emph{Conference on Computational Natural Language Learning}, 2022. [Online]. Available: \url{https://api.semanticscholar.org/CorpusID:253116568}
\BIBentrySTDinterwordspacing

\bibitem{schwettmann2023multimodal}
S.~Schwettmann, N.~Chowdhury, S.~Klein, D.~Bau, and A.~Torralba, ``Multimodal neurons in pretrained text-only transformers,'' in \emph{Proceedings of the IEEE/CVF International Conference on Computer Vision}, 2023, pp. 2862--2867.

\bibitem{Tang2024LanguageSpecificNT}
\BIBentryALTinterwordspacing
T.~Tang, W.~Luo, H.~Huang, D.~Zhang, X.~Wang, X.~Zhao, F.~Wei, and J.-R. Wen, ``Language-specific neurons: The key to multilingual capabilities in large language models,'' in \emph{Annual Meeting of the Association for Computational Linguistics}, 2024. [Online]. Available: \url{https://api.semanticscholar.org/CorpusID:268032136}
\BIBentrySTDinterwordspacing

\bibitem{Kojima2024OnTM}
\BIBentryALTinterwordspacing
T.~Kojima, I.~Okimura, Y.~Iwasawa, H.~Yanaka, and Y.~Matsuo, ``On the multilingual ability of decoder-based pre-trained language models: Finding and controlling language-specific neurons,'' \emph{ArXiv}, vol. abs/2404.02431, 2024. [Online]. Available: \url{https://api.semanticscholar.org/CorpusID:268875934}
\BIBentrySTDinterwordspacing

\bibitem{gurnee2024universal}
W.~Gurnee, T.~Horsley, Z.~C. Guo, T.~R. Kheirkhah, Q.~Sun, W.~Hathaway, N.~Nanda, and D.~Bertsimas, ``Universal neurons in gpt2 language models,'' \emph{arXiv preprint arXiv:2401.12181}, 2024.

\bibitem{cohen2011measuring}
M.~R. Cohen and A.~Kohn, ``Measuring and interpreting neuronal correlations,'' \emph{Nature neuroscience}, vol.~14, no.~7, pp. 811--819, 2011.

\bibitem{raffel2020exploring}
C.~Raffel, N.~Shazeer, A.~Roberts, K.~Lee, S.~Narang, M.~Matena, Y.~Zhou, W.~Li, and P.~J. Liu, ``Exploring the limits of transfer learning with a unified text-to-text transformer,'' \emph{Journal of machine learning research}, vol.~21, no. 140, pp. 1--67, 2020.

\bibitem{mahendran2015understanding}
A.~Mahendran and A.~Vedaldi, ``Understanding deep image representations by inverting them,'' in \emph{Proceedings of the IEEE conference on computer vision and pattern recognition}, 2015, pp. 5188--5196.

\bibitem{dosovitskiy2015inverting}
A.~Dosovitskiy, T.~Brox \emph{et~al.}, ``Inverting convolutional networks with convolutional networks,'' \emph{arXiv preprint arXiv:1506.02753}, vol.~4, no.~2, p.~3, 2015.

\bibitem{cordonnier2019relationship}
J.-B. Cordonnier, A.~Loukas, and M.~Jaggi, ``On the relationship between self-attention and convolutional layers,'' \emph{arXiv preprint arXiv:1911.03584}, 2019.

\bibitem{sukhbaatar2019augmenting}
S.~Sukhbaatar, E.~Grave, G.~Lample, H.~Jegou, and A.~Joulin, ``Augmenting self-attention with persistent memory,'' \emph{arXiv preprint arXiv:1907.01470}, 2019.

\bibitem{geva2020transformer}
M.~Geva, R.~Schuster, J.~Berant, and O.~Levy, ``Transformer feed-forward layers are key-value memories,'' \emph{arXiv preprint arXiv:2012.14913}, 2020.

\bibitem{michel2019sixteen}
P.~Michel, O.~Levy, and G.~Neubig, ``Are sixteen heads really better than one?'' \emph{Advances in neural information processing systems}, vol.~32, 2019.

\bibitem{voita2019analyzing}
E.~Voita, D.~Talbot, F.~Moiseev, R.~Sennrich, and I.~Titov, ``Analyzing multi-head self-attention: Specialized heads do the heavy lifting, the rest can be pruned,'' \emph{arXiv preprint arXiv:1905.09418}, 2019.

\bibitem{clark2019does}
K.~Clark, ``What does bert look at? an analysis of bert’s attention,'' \emph{arXiv preprint arXiv:1906.04341}, 2019.

\bibitem{kovaleva2019revealing}
O.~Kovaleva, A.~Romanov, A.~Rogers, and A.~Rumshisky, ``Revealing the dark secrets of bert,'' \emph{arXiv preprint arXiv:1908.08593}, 2019.

\bibitem{htut2019attention}
P.~M. Htut, J.~Phang, S.~Bordia, and S.~R. Bowman, ``Do attention heads in bert track syntactic dependencies?'' \emph{arXiv preprint arXiv:1911.12246}, 2019.

\bibitem{si2024freeu}
C.~Si, Z.~Huang, Y.~Jiang, and Z.~Liu, ``Freeu: Free lunch in diffusion u-net,'' in \emph{Proceedings of the IEEE/CVF Conference on Computer Vision and Pattern Recognition}, 2024, pp. 4733--4743.

\bibitem{kowal2024visual}
M.~Kowal, R.~P. Wildes, and K.~G. Derpanis, ``Visual concept connectome (vcc): Open world concept discovery and their interlayer connections in deep models,'' in \emph{Proceedings of the IEEE/CVF Conference on Computer Vision and Pattern Recognition}, 2024, pp. 10\,895--10\,905.

\bibitem{van2019does}
B.~Van~Aken, B.~Winter, A.~L{\"o}ser, and F.~A. Gers, ``How does bert answer questions? a layer-wise analysis of transformer representations,'' in \emph{Proceedings of the 28th ACM international conference on information and knowledge management}, 2019, pp. 1823--1832.

\bibitem{tenney2019bert}
I.~Tenney, ``Bert rediscovers the classical nlp pipeline,'' \emph{arXiv preprint arXiv:1905.05950}, 2019.

\bibitem{palit2023towards}
V.~Palit, R.~Pandey, A.~Arora, and P.~P. Liang, ``Towards vision-language mechanistic interpretability: A causal tracing tool for blip,'' in \emph{Proceedings of the IEEE/CVF International Conference on Computer Vision}, 2023, pp. 2856--2861.

\bibitem{prasad2023unraveling}
V.~Prasad, C.~Zhu-Tian, A.~Vilanova, H.~Pfister, N.~Pezzotti, and H.~Strobelt, ``Unraveling the temporal dynamics of the unet in diffusion models,'' \emph{arXiv preprint arXiv:2312.14965}, 2023.

\bibitem{ribeiro2016should}
M.~T. Ribeiro, S.~Singh, and C.~Guestrin, ``" why should i trust you?" explaining the predictions of any classifier,'' in \emph{Proceedings of the 22nd ACM SIGKDD international conference on knowledge discovery and data mining}, 2016, pp. 1135--1144.

\bibitem{shrikumar2017learning}
A.~Shrikumar, P.~Greenside, and A.~Kundaje, ``Learning important features through propagating activation differences,'' in \emph{International conference on machine learning}.\hskip 1em plus 0.5em minus 0.4em\relax PMlR, 2017, pp. 3145--3153.

\bibitem{lundberg2017unified}
S.~Lundberg, ``A unified approach to interpreting model predictions,'' \emph{arXiv preprint arXiv:1705.07874}, 2017.

\bibitem{zhou2016learning}
B.~Zhou, A.~Khosla, A.~Lapedriza, A.~Oliva, and A.~Torralba, ``Learning deep features for discriminative localization,'' in \emph{Proceedings of the IEEE conference on computer vision and pattern recognition}, 2016, pp. 2921--2929.

\bibitem{selvaraju2017grad}
R.~R. Selvaraju, M.~Cogswell, A.~Das, R.~Vedantam, D.~Parikh, and D.~Batra, ``Grad-cam: Visual explanations from deep networks via gradient-based localization,'' in \emph{Proceedings of the IEEE international conference on computer vision}, 2017, pp. 618--626.

\bibitem{patro2019u}
B.~N. Patro, M.~Lunayach, S.~Patel, and V.~P. Namboodiri, ``U-cam: Visual explanation using uncertainty based class activation maps,'' in \emph{Proceedings of the IEEE/CVF International Conference on Computer Vision}, 2019, pp. 7444--7453.

\bibitem{wang2020score}
H.~Wang, Z.~Wang, M.~Du, F.~Yang, Z.~Zhang, S.~Ding, P.~Mardziel, and X.~Hu, ``Score-cam: Score-weighted visual explanations for convolutional neural networks,'' in \emph{Proceedings of the IEEE/CVF conference on computer vision and pattern recognition workshops}, 2020, pp. 24--25.

\bibitem{chen2022gscorecam}
P.~Chen, Q.~Li, S.~Biaz, T.~Bui, and A.~Nguyen, ``gscorecam: What objects is clip looking at?'' in \emph{Proceedings of the Asian Conference on Computer Vision}, 2022, pp. 1959--1975.

\bibitem{simonyan2013deep}
K.~Simonyan, ``Deep inside convolutional networks: Visualising image classification models and saliency maps,'' \emph{arXiv preprint arXiv:1312.6034}, 2013.

\bibitem{binder2016layer}
A.~Binder, G.~Montavon, S.~Lapuschkin, K.-R. M{\"u}ller, and W.~Samek, ``Layer-wise relevance propagation for neural networks with local renormalization layers,'' in \emph{Artificial Neural Networks and Machine Learning--ICANN 2016: 25th International Conference on Artificial Neural Networks, Barcelona, Spain, September 6-9, 2016, Proceedings, Part II 25}.\hskip 1em plus 0.5em minus 0.4em\relax Springer, 2016, pp. 63--71.

\bibitem{montavon2017explaining}
G.~Montavon, S.~Lapuschkin, A.~Binder, W.~Samek, and K.-R. M{\"u}ller, ``Explaining nonlinear classification decisions with deep taylor decomposition,'' \emph{Pattern recognition}, vol.~65, pp. 211--222, 2017.

\bibitem{qi2019visualizing}
Z.~Qi, S.~Khorram, and F.~Li, ``Visualizing deep networks by optimizing with integrated gradients.'' in \emph{CVPR workshops}, vol.~2, 2019, pp. 1--4.

\bibitem{chefer2021generic}
H.~Chefer, S.~Gur, and L.~Wolf, ``Generic attention-model explainability for interpreting bi-modal and encoder-decoder transformers,'' in \emph{Proceedings of the IEEE/CVF International Conference on Computer Vision}, 2021, pp. 397--406.

\bibitem{olah2018building}
C.~Olah, A.~Satyanarayan, I.~Johnson, S.~Carter, L.~Schubert, K.~Ye, and A.~Mordvintsev, ``The building blocks of interpretability,'' \emph{Distill}, vol.~3, no.~3, p. e10, 2018.

\bibitem{bach2015pixel}
S.~Bach, A.~Binder, G.~Montavon, F.~Klauschen, K.-R. M{\"u}ller, and W.~Samek, ``On pixel-wise explanations for non-linear classifier decisions by layer-wise relevance propagation,'' \emph{PloS one}, vol.~10, no.~7, p. e0130140, 2015.

\bibitem{sundararajan2020shapley}
M.~Sundararajan, K.~Dhamdhere, and A.~Agarwal, ``The shapley taylor interaction index,'' in \emph{International conference on machine learning}.\hskip 1em plus 0.5em minus 0.4em\relax PMLR, 2020, pp. 9259--9268.

\bibitem{janizek2021explaining}
J.~D. Janizek, P.~Sturmfels, and S.-I. Lee, ``Explaining explanations: Axiomatic feature interactions for deep networks,'' \emph{Journal of Machine Learning Research}, vol.~22, no. 104, pp. 1--54, 2021.

\bibitem{tsai2023faith}
C.-P. Tsai, C.-K. Yeh, and P.~Ravikumar, ``Faith-shap: The faithful shapley interaction index,'' \emph{Journal of Machine Learning Research}, vol.~24, no.~94, pp. 1--42, 2023.

\bibitem{ren2023defining}
J.~Ren, M.~Li, Q.~Chen, H.~Deng, and Q.~Zhang, ``Defining and quantifying the emergence of sparse concepts in dnns,'' in \emph{Proceedings of the IEEE/CVF conference on computer vision and pattern recognition}, 2023, pp. 20\,280--20\,289.

\bibitem{li2023does}
M.~Li and Q.~Zhang, ``Does a neural network really encode symbolic concepts?'' in \emph{International conference on machine learning}.\hskip 1em plus 0.5em minus 0.4em\relax PMLR, 2023, pp. 20\,452--20\,469.

\bibitem{renwe}
Q.~Ren, J.~Gao, W.~Shen, and Q.~Zhang, ``Where we have arrived in proving the emergence of sparse interaction primitives in dnns,'' in \emph{The Twelfth International Conference on Learning Representations}, 2024.

\bibitem{ren2021can}
J.~Ren, Z.~Zhou, Q.~Chen, and Q.~Zhang, ``Can we faithfully represent masked states to compute shapley values on a dnn?'' \emph{arXiv preprint arXiv:2105.10719}, 2021.

\bibitem{chen2024defining}
L.~Chen, S.~Lou, B.~Huang, and Q.~Zhang, ``Defining and extracting generalizable interaction primitives from dnns,'' \emph{arXiv preprint arXiv:2401.16318}, 2024.

\bibitem{cheng2024layerwise}
X.~Cheng, L.~Cheng, Z.~Peng, Y.~Xu, T.~Han, and Q.~Zhang, ``Layerwise change of knowledge in neural networks,'' \emph{arXiv preprint arXiv:2409.08712}, 2024.

\bibitem{harkonen2020ganspace}
E.~H{\"a}rk{\"o}nen, A.~Hertzmann, J.~Lehtinen, and S.~Paris, ``Ganspace: Discovering interpretable gan controls,'' \emph{Advances in neural information processing systems}, vol.~33, pp. 9841--9850, 2020.

\bibitem{ren2023we}
Q.~Ren, J.~Gao, W.~Shen, and Q.~Zhang, ``Where we have arrived in proving the emergence of sparse symbolic concepts in ai models,'' \emph{arXiv preprint arXiv:2305.01939}, 2023.

\bibitem{wang2020unified}
X.~Wang, J.~Ren, S.~Lin, X.~Zhu, Y.~Wang, and Q.~Zhang, ``A unified approach to interpreting and boosting adversarial transferability,'' \emph{arXiv preprint arXiv:2010.04055}, 2020.

\bibitem{ren2021towards}
J.~Ren, D.~Zhang, Y.~Wang, L.~Chen, Z.~Zhou, Y.~Chen, X.~Cheng, X.~Wang, M.~Zhou, J.~Shi \emph{et~al.}, ``Towards a unified game-theoretic view of adversarial perturbations and robustness,'' \emph{Advances in Neural Information Processing Systems}, vol.~34, pp. 3797--3810, 2021.

\bibitem{zhou2024explaining}
H.~Zhou, H.~Zhang, H.~Deng, D.~Liu, W.~Shen, S.-H. Chan, and Q.~Zhang, ``Explaining generalization power of a dnn using interactive concepts,'' in \emph{Proceedings of the AAAI Conference on Artificial Intelligence}, vol.~38, no.~15, 2024, pp. 17\,105--17\,113.

\bibitem{liu2024towards}
D.~Liu, H.~Deng, X.~Cheng, Q.~Ren, K.~Wang, and Q.~Zhang, ``Towards the difficulty for a deep neural network to learn concepts of different complexities,'' \emph{Advances in Neural Information Processing Systems}, vol.~36, 2024.

\bibitem{deng2021discovering}
H.~Deng, Q.~Ren, H.~Zhang, and Q.~Zhang, ``Discovering and explaining the representation bottleneck of dnns,'' \emph{arXiv preprint arXiv:2111.06236}, 2021.

\bibitem{ren2023bayesian}
Q.~Ren, H.~Deng, Y.~Chen, S.~Lou, and Q.~Zhang, ``Bayesian neural networks avoid encoding complex and perturbation-sensitive concepts,'' in \emph{International Conference on Machine Learning}.\hskip 1em plus 0.5em minus 0.4em\relax PMLR, 2023, pp. 28\,889--28\,913.

\bibitem{ren2024towards}
Q.~Ren, Y.~Xu, J.~Zhang, Y.~Xin, D.~Liu, and Q.~Zhang, ``Towards the dynamics of a dnn learning symbolic interactions,'' \emph{arXiv preprint arXiv:2407.19198}, 2024.

\bibitem{zhang2024two}
J.~Zhang, Q.~Li, L.~Lin, and Q.~Zhang, ``Two-phase dynamics of interactions explains the starting point of a dnn learning over-fitted features,'' \emph{arXiv preprint arXiv:2405.10262}, 2024.

\bibitem{deng2024unifying}
H.~Deng, N.~Zou, M.~Du, W.~Chen, G.~Feng, Z.~Yang, Z.~Li, and Q.~Zhang, ``Unifying fourteen post-hoc attribution methods with taylor interactions,'' \emph{IEEE Transactions on Pattern Analysis and Machine Intelligence}, 2024.

\bibitem{tran2018importance}
K.~Tran, A.~Bisazza, and C.~Monz, ``The importance of being recurrent for modeling hierarchical structure,'' \emph{arXiv preprint arXiv:1803.03585}, 2018.

\bibitem{tang2021interpretable}
X.~Tang, W.~Zhang, Y.~Yu, K.~Turner, T.~Derr, M.~Wang, and E.~Ntoutsi, ``Interpretable visual understanding with cognitive attention network,'' in \emph{Artificial Neural Networks and Machine Learning--ICANN 2021: 30th International Conference on Artificial Neural Networks, Bratislava, Slovakia, September 14--17, 2021, Proceedings, Part I 30}.\hskip 1em plus 0.5em minus 0.4em\relax Springer, 2021, pp. 555--568.

\bibitem{liu2018improving}
X.~Liu, X.~Wang, and S.~Matwin, ``Improving the interpretability of deep neural networks with knowledge distillation,'' in \emph{2018 IEEE International Conference on Data Mining Workshops (ICDMW)}.\hskip 1em plus 0.5em minus 0.4em\relax IEEE, 2018, pp. 905--912.

\bibitem{wong2021leveraging}
E.~Wong, S.~Santurkar, and A.~Madry, ``Leveraging sparse linear layers for debuggable deep networks,'' in \emph{International Conference on Machine Learning}.\hskip 1em plus 0.5em minus 0.4em\relax PMLR, 2021, pp. 11\,205--11\,216.

\bibitem{koh2020concept}
P.~W. Koh, T.~Nguyen, Y.~S. Tang, S.~Mussmann, E.~Pierson, B.~Kim, and P.~Liang, ``Concept bottleneck models,'' in \emph{International conference on machine learning}.\hskip 1em plus 0.5em minus 0.4em\relax PMLR, 2020, pp. 5338--5348.

\bibitem{yuksekgonul2022post}
M.~Yuksekgonul, M.~Wang, and J.~Zou, ``Post-hoc concept bottleneck models,'' \emph{arXiv preprint arXiv:2205.15480}, 2022.

\bibitem{chenllcp}
G.~Chen, Y.~Li, X.~Liu, Z.~Li, E.~Al~Suradi, D.~Wei, and K.~Zhang, ``Llcp: Learning latent causal processes for reasoning-based video question answer,'' in \emph{ICLR}, 2024.

\bibitem{li2024steering}
J.~Li, Z.~Tang, X.~Liu, P.~Spirtes, K.~Zhang, L.~Leqi, and Y.~Liu, ``Steering llms towards unbiased responses: A causality-guided debiasing framework,'' \emph{arXiv preprint arXiv:2403.08743}, 2024.

\bibitem{liu2024revealing}
Y.~Liu, Z.~Zhang, D.~Gong, B.~Huang, M.~Gong, A.~v.~d. Hengel, K.~Zhang, and J.~Q. Shi, ``Revealing multimodal contrastive representation learning through latent partial causal models,'' \emph{arXiv preprint arXiv:2402.06223}, 2024.

\bibitem{yun2022vision}
T.~Yun, U.~Bhalla, E.~Pavlick, and C.~Sun, ``Do vision-language pretrained models learn composable primitive concepts?'' \emph{arXiv preprint arXiv:2203.17271}, 2022.

\bibitem{zhou2024lima}
C.~Zhou, P.~Liu, P.~Xu, S.~Iyer, J.~Sun, Y.~Mao, X.~Ma, A.~Efrat, P.~Yu, L.~Yu \emph{et~al.}, ``Lima: Less is more for alignment,'' \emph{Advances in Neural Information Processing Systems}, vol.~36, 2024.

\bibitem{li2023silkie}
L.~Li, Z.~Xie, M.~Li, S.~Chen, P.~Wang, L.~Chen, Y.~Yang, B.~Wang, and L.~Kong, ``Silkie: Preference distillation for large visual language models,'' \emph{arXiv preprint arXiv:2312.10665}, 2023.

\bibitem{smilkov2017smoothgrad}
D.~Smilkov, N.~Thorat, B.~Kim, F.~Vi{\'e}gas, and M.~Wattenberg, ``Smoothgrad: removing noise by adding noise,'' \emph{arXiv preprint arXiv:1706.03825}, 2017.

\bibitem{huang2023survey}
L.~Huang, W.~Yu, W.~Ma, W.~Zhong, Z.~Feng, H.~Wang, Q.~Chen, W.~Peng, X.~Feng, B.~Qin \emph{et~al.}, ``A survey on hallucination in large language models: Principles, taxonomy, challenges, and open questions,'' \emph{arXiv preprint arXiv:2311.05232}, 2023.

\bibitem{rawte2023survey}
V.~Rawte, A.~Sheth, and A.~Das, ``A survey of hallucination in large foundation models,'' \emph{arXiv preprint arXiv:2309.05922}, 2023.

\bibitem{zheng2023ddcot}
G.~Zheng, B.~Yang, J.~Tang, H.-Y. Zhou, and S.~Yang, ``Ddcot: Duty-distinct chain-of-thought prompting for multimodal reasoning in language models,'' \emph{Advances in Neural Information Processing Systems}, vol.~36, pp. 5168--5191, 2023.

\bibitem{shao2024visual}
H.~Shao, S.~Qian, H.~Xiao, G.~Song, Z.~Zong, L.~Wang, Y.~Liu, and H.~Li, ``Visual cot: Unleashing chain-of-thought reasoning in multi-modal language models,'' \emph{arXiv preprint arXiv:2403.16999}, 2024.

\bibitem{mondal2024kam}
D.~Mondal, S.~Modi, S.~Panda, R.~Singh, and G.~S. Rao, ``Kam-cot: Knowledge augmented multimodal chain-of-thoughts reasoning,'' in \emph{Proceedings of the AAAI Conference on Artificial Intelligence}, vol.~38, no.~17, 2024, pp. 18\,798--18\,806.

\bibitem{chen2024m}
Q.~Chen, L.~Qin, J.~Zhang, Z.~Chen, X.~Xu, and W.~Che, ``M3cot: A novel benchmark for multi-domain multi-step multi-modal chain-of-thought,'' \emph{arXiv preprint arXiv:2405.16473}, 2024.

\bibitem{rose2023visual}
D.~Rose, V.~Himakunthala, A.~Ouyang, R.~He, A.~Mei, Y.~Lu, M.~Saxon, C.~Sonar, D.~Mirza, and W.~Y. Wang, ``Visual chain of thought: bridging logical gaps with multimodal infillings,'' \emph{arXiv preprint arXiv:2305.02317}, 2023.

\bibitem{yellinek20233vl}
N.~Yellinek, L.~Karlinsky, and R.~Giryes, ``3vl: using trees to teach vision \& language models compositional concepts,'' \emph{arXiv preprint arXiv:2312.17345}, 2023.

\bibitem{dong2022survey}
Q.~Dong, L.~Li, D.~Dai, C.~Zheng, J.~Ma, R.~Li, H.~Xia, J.~Xu, Z.~Wu, T.~Liu \emph{et~al.}, ``A survey on in-context learning,'' \emph{arXiv preprint arXiv:2301.00234}, 2022.

\bibitem{bansal2022rethinking}
H.~Bansal, K.~Gopalakrishnan, S.~Dingliwal, S.~Bodapati, K.~Kirchhoff, and D.~Roth, ``Rethinking the role of scale for in-context learning: An interpretability-based case study at 66 billion scale,'' \emph{arXiv preprint arXiv:2212.09095}, 2022.

\bibitem{miyanishi2024multimodal}
Y.~Miyanishi and M.~L. Nguyen, ``Multimodal contrastive in-context learning,'' \emph{arXiv preprint arXiv:2408.12959}, 2024.

\end{thebibliography}









\newpage


\vfill

\end{document}